\documentclass[letterpaper]{article} 
\usepackage{aaai25}  
\usepackage{times}  
\usepackage{helvet}  
\usepackage{courier}  
\usepackage[hyphens]{url}  
\usepackage{graphicx} 
\usepackage{natbib}  
\usepackage{caption} 
\usepackage{algorithm}
\usepackage{algorithmic}

\usepackage{amsfonts}
\usepackage{amsmath}
\usepackage{multirow}
\usepackage{subcaption}

\usepackage{newfloat}
\usepackage{listings}
\DeclareCaptionStyle{ruled}{labelfont=normalfont,labelsep=colon,strut=off} 
\floatstyle{ruled}
\newfloat{listing}{tb}{lst}{}
\floatname{listing}{Listing}
\title{xPatch: Dual-Stream Time Series Forecasting with \\
Exponential Seasonal-Trend Decomposition}
\author {
    Artyom Stitsyuk \textsuperscript{\rm 1}, 
    Jaesik Choi \textsuperscript{\rm 1,2}
}
\affiliations {
    \textsuperscript{\rm 1}Korea Advanced Institute of Science and Technology (KAIST), South Korea\\
    \textsuperscript{\rm 2}INEEJI, South Korea\\
    \{stitsyuk, jaesik.choi\}@kaist.ac.kr
}

\begin{document}

\maketitle

\begin{abstract}
In recent years, the application of transformer-based models in time-series forecasting has received significant attention.
While often demonstrating promising results, the transformer architecture encounters challenges in fully exploiting the temporal relations within time series data due to its attention mechanism.
In this work, we design e\textbf{X}ponential \textbf{Patch} (xPatch for short), a novel dual-stream architecture that utilizes exponential decomposition.
Inspired by the classical exponential smoothing approaches, xPatch introduces the innovative seasonal-trend exponential decomposition module.
Additionally, we propose a dual-flow architecture that consists of an MLP-based linear stream and a CNN-based non-linear stream.
This model investigates the benefits of employing patching and channel-independence techniques within a non-transformer model.
Finally, we develop a robust arctangent loss function and a sigmoid learning rate adjustment scheme, which prevent overfitting and boost forecasting performance.
The code is available at the following repository: \url{https://github.com/stitsyuk/xPatch}.
\end{abstract}

\section{Introduction}
\label{introduction}
Long-term time series forecasting (LTSF) is one of the fundamental tasks in time series analysis.
The task is focused on predicting future values over an extended period, based on historical data.
With the advent of deep learning models, they have recently demonstrated superior performance in LTSF compared to traditional approaches such as ARIMA \cite{box2015time} and LSTM \cite{bahdanau2014neural}.

Transformer-based models \cite{vaswani2017attention} have revolutionized the LTSF task, enabling powerful AI systems to achieve state-of-the-art performance.
The transformer architecture is considered highly successful in capturing semantic correlations among elements in long sequences.
Recent research efforts have been primarily focused on adapting transformers to the LTSF task and addressing such limitations of the vanilla transformer as quadratic time and memory complexity \cite{li2019enhancing, zhou2021informer, wen2022transformers}.

The self-attention mechanism employed in transformers is permutation-invariant.
Although techniques like positional encoding can partially retain ordering information, preserving temporal information remains a challenge for transformer-based models.
This limitation can adversely affect the performance of the LTSF task dealing with a continuous set of points.
As a result, the effectiveness of transformers in the LTSF task has been challenged by a simple linear approach utilizing a Multi-Layer Perceptron (MLP) network \cite{zeng2023transformers}.
Surprisingly, a simple linear model named DLinear has surpassed the state-of-the-art forecasting performance of all previous transformer-based models, raising a fundamental question: “Are Transformers effective for long-term time series forecasting?”.

Due to the non-stationary nature of real-world systems, time series data usually contain complex temporal patterns.
To handle this complexity and non-stationarity \cite{liu2022non}, many recent LTSF models have adopted a paradigm of decomposing inputs.
They use a seasonal-trend decomposition to capture linear trend features and non-linear seasonal variations.
For handling time series trend features, certain transformer-based models, including Autoformer \cite{wu2021autoformer} and FEDformer \cite{zhou2022fedformer}, incorporate seasonal-trend data decomposition.
By partitioning the signal into two components, each with distinct function behavior, it becomes more feasible to capture semantic features from each component and make separate predictions.

Both Autoformer and FEDformer focus on refining the transformer architecture by introducing an auto-correlation mechanism and a frequency-enhanced method while decomposing the signal using a simple average pooling method.
This technique requires padding at both ends, essentially repeating the last and first values.
Consequently, we argue that this approach introduces a bias towards the initial and final values, potentially altering the behavior of trend values.

We propose a simple yet effective decomposition technique based on a generally applicable time series smoothing method named Exponential Moving Average (EMA) \cite{gardner1985exponential}.
The proposed strategy assigns exponentially decreasing weights over time, facilitating more efficient feature learning from the decomposed data.
The resulting exponentially smoothed sequence represents the trend, while the residual difference encapsulates the seasonality.

Currently, the state-of-the-art models for the LTSF task are transformer-based architectures CARD \cite{wang2024card} and PatchTST \cite{yuqietal2023patch}.
These models rely on channel-independence and segmentation of time series into patches, which are used as input tokens for the transformer.
However, we assume that the permutation-invariance of the attention mechanism in transformers may impede the model from attaining the optimal forecasting performance.
Therefore, we are aiming to explore channel-independence and patching approaches within a non-transformer architecture, proposing the xPatch model.

In this study, we introduce the utilization of the exponential seasonal-trend decomposition technique.
Furthermore, we propose a robust arctangent loss with weight decay and a novel learning rate adjustment strategy that improves training adaptability.
Additionally, we present the xPatch, a novel dual-flow network architecture that integrates Convolutional Neural Networks (CNNs), Multi-Layer Perceptrons (MLPs), patching, channel-independence, exponential seasonal-trend decomposition, and dual stream prediction.

We summarized our main contributions as follows:
\begin{itemize}
    \item We propose a novel method for seasonal-trend decomposition that utilizes an Exponential Moving Average (EMA).
    \item We introduce the dual-flow network and investigate the patching and channel-independence approaches within the CNN-based backbone.
    \item We develop a robust arctangent loss and a novel sigmoid learning rate adjustment scheme with a warm-up that results in smoother training.
\end{itemize}

\section{Related Work}
Informer \cite{zhou2021informer} is the first well-known transformer-based model designed for the LTSF task.
It employs ProbSparse self-attention and a generative style decoder for addressing quadratic time and memory complexity.
Notably, this work also contributes to the field by curating data and introducing the Electricity Transformer Temperature (ETT) benchmark dataset that is now commonly used for LTSF experiments by most of the models.

TimesNet \cite{wu2023timesnet} utilizes Fourier Transform to decompose time series into multiple components with varying period lengths, enhancing its focus on temporal variation modeling.
The official repository provides a forecasting protocol with standardized hyperparameter settings and fairly implemented baselines.

To address the issue of non-stationarity in time series data, several models employ series decomposition to better capture complex temporal patterns.
Autoformer \cite{wu2021autoformer} and FEDformer \cite{zhou2022fedformer} are two recent transformer-based solutions for the LTSF task, leveraging auto-correlation mechanism and frequency-enhanced structure, respectively.
Both models incorporate seasonal-trend decomposition within each neural block to enhance the predictability of time-series data.
Specifically, they apply a moving average kernel to the input sequence with padding at both ends, extracting the trend component.
The difference between the original time series and the extracted trend component is identified as the seasonal component.

DLinear \cite{zeng2023transformers} is a recent one-layer linear model that uses seasonal-trend decomposition as a pre-processing step.
Initially, the model decomposes the raw data into trend and seasonal components using a moving average technique.
Two linear layers are then applied independently to each of these components.
The resulting features are subsequently aggregated to generate the final prediction.

MICN \cite{wang2023micn} is a recent CNN-based solution that employs multi-scale hybrid seasonal-trend decomposition.
After decomposing the input series into seasonal and trend components, the model integrates both global and local contexts to enhance forecasting accuracy.

TimeMixer \cite{wang2024timemixer} is an MLP-based approach that employs a decomposable multiscale-mixing method.
The model uses the same series decomposition block from Autoformer \cite{wu2021autoformer} to break down multiscale time series into multiple seasonal and trend components.
By leveraging the multiscale past information obtained after seasonal and trend mixing, the model predicts future values.

ETSformer \cite{woo2022etsformer} and CARD \cite{wang2024card} are two transformer-based architectures that incorporate the exponential smoothing approach.
ETSformer introduces Exponential Smoothing Attention (ESA), while CARD applies exponential smoothing to the query and key tokens before the token blending module within one prediction head of the attention mechanism.
In contrast to these models, the proposed xPatch architecture employs Exponential Moving Average (EMA) decomposition to separate the time series into trend and seasonal components, which are then processed separately.

Crossformer \cite{zhang2022crossformer} and PatchTST \cite{yuqietal2023patch} are transformer-based models that introduce a segmentation technique to LTSF.
PatchTST divides time series data into subseries-level patches that serve as input tokens for the transformer.
This approach is motivated by the vision transformer \cite{dosovitskiy2020vit} and designed for LTSF with channel-independence.
Currently, PatchTST is recognized as the state-of-the-art solution for multivariate long-term forecasting.
In our proposed xPatch model, we also incorporate patching and channel-independence approaches.
Given that xPatch is a CNN-based approach, we investigate whether the superior performance of PatchTST can be attributed to its patching and channel-independence modules rather than its transformer architecture.
To explore this, we examine if a CNN-based model can achieve improved results by leveraging these techniques. 

MobileNet \cite{howard2017mobilenets} and ConvMixer \cite{trockman2022patches} are notable models designed for Computer Vision (CV) tasks that demonstrate the advantages of depthwise separable convolutions.
In the proposed xPatch approach, we incorporate depthwise separable convolution as the non-linear stream of the dual-flow network.


\section{Proposed Method}
In multivariate time series forecasting, given the observation of the historical $L$ values $x = (x_1, x_2, ... , x_L)$, the task is to predict the future $T$ timesteps $\hat{x} = (x_{L+1}, x_{L+2}, ... , x_{L+T})$.
Each $x_t$ value at timestep $t$ is multivariate, representing a vector of $M$ variables.
Therefore, the multivariate lookback series is denoted as $x \in \mathbb{R}^{M \times L}$ and the multivariate prediction is represented by $\hat{x} \in \mathbb{R}^{M \times T}$.




\subsection{Seasonal-Trend Decomposition}
Seasonal-trend decomposition facilitates the learning of complex temporal patterns by separating the time series signal into trend and seasonal components.
Trend features generally represent the long-term direction of the data, which can be linear or smoothly varying.
In contrast, seasonal components capture repeating patterns or cycles that occur at regular intervals and are often non-linear due to the complexities and variations in periodic behavior.
The model first learns the features of these components individually and then combines them to generate the final forecast.

\textbf{Simple Moving Average (SMA)} is the decomposition approach utilized in Autoformer \cite{wu2021autoformer}, FEDformer \cite{zhou2022fedformer}, DLinear \cite{zeng2023transformers}, MICN \cite{wang2023micn}, and TimeMixer \cite{wang2024timemixer} models.
SMA is defined as the unweighted mean of the previous $k$ data points.

\begin{figure}[ht]
\centering
\includegraphics[width=1\columnwidth]{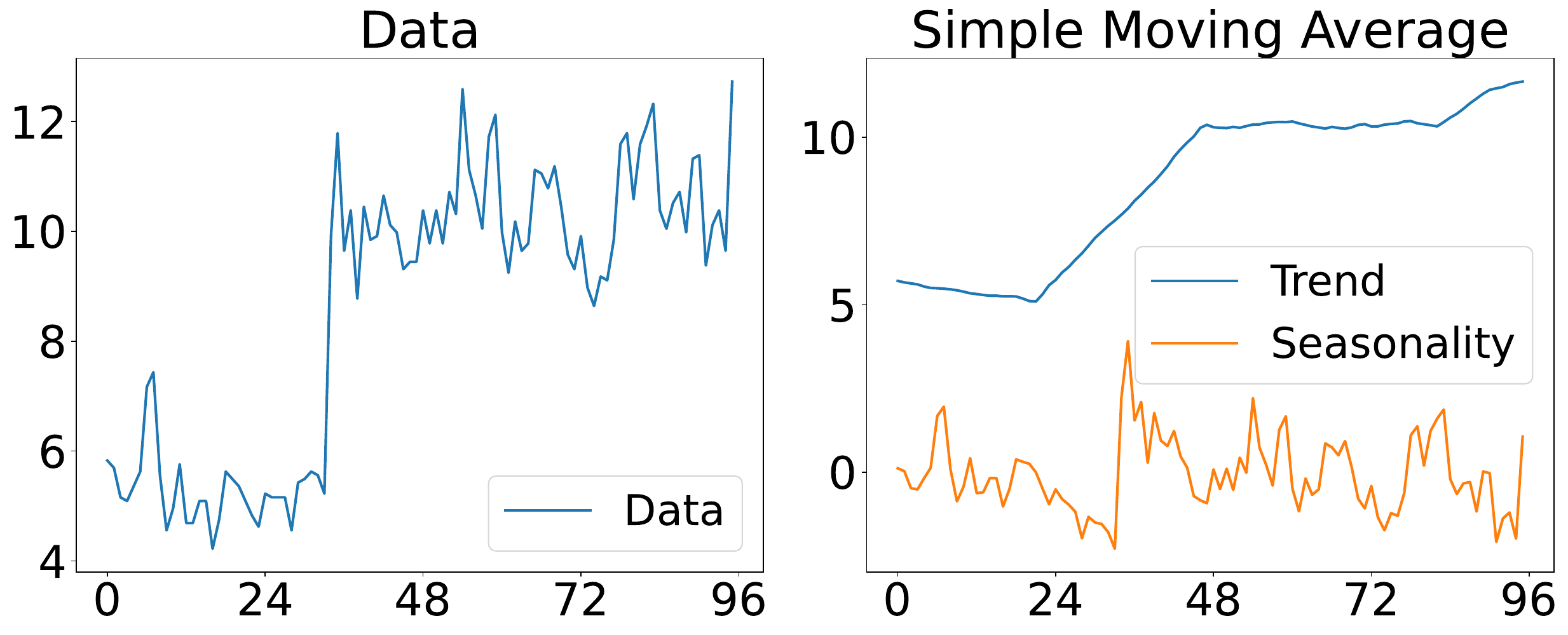}
\caption{Example of SMA decomposition with kernel k = 25 on a 96-length sample from the ETTh1 dataset.}
\label{fig:ma}
\end{figure}

Moving average mean point $s_t$ of the $k$ entries with $t$ being moving step, $n$ being dataset length, and $X = x_1, x_2, ..., x_n$ being data points is calculated as:
\begin{equation}
\begin{aligned}
\label{equ:sma}
s_t &= \frac{x_{t}+x_{t+1}+...+x_{t+k-1}}{k} = \frac{1}{k}\sum^{t+k-1}_{i=t}x_i \\
X_T &= \text{AvgPool}(\text{Padding}(X)) \\
X_S &= X - X_T
\end{aligned}
\end{equation}
where $\text{AvgPool}(\cdot)$ denotes moving average with the padding operation, while $X_T$ and $X_S$ correspond to trend and seasonality components.
Padding is employed to maintain the length of the time series unchanged after performing average pooling.
Figure \ref{fig:ma} illustrates an example of SMA decomposition.

Firstly, we argue that the average pooling operation results in the loss of significant trend features (see Appendix \ref{app:sma}).
Additionally, alignment requires padding on both ends of the series, which can distort the sequence at the head and tail.

Secondly, the primary goal of decomposition is to enhance the interpretability of both decomposed signals.
This entails improving the clarity of the trend and seasonality components while enriching them with more distinct features for learning.
However, SMA produces an overly simplistic trend signal with limited diverse features and a complex seasonality pattern.
As a result, we investigate an alternative decomposition method to address this issue.

\textbf{Exponential Moving Average (EMA)} \cite{gardner1985exponential} is an exponential smoothing method that assigns greater weight to more recent data points while smoothing out older data.
This exponential weighting scheme allows EMA to respond more promptly to changes in the underlying trends of the time series, without the need for padding repeated values.

\begin{figure}[ht]
\centering
\includegraphics[width=1\columnwidth]{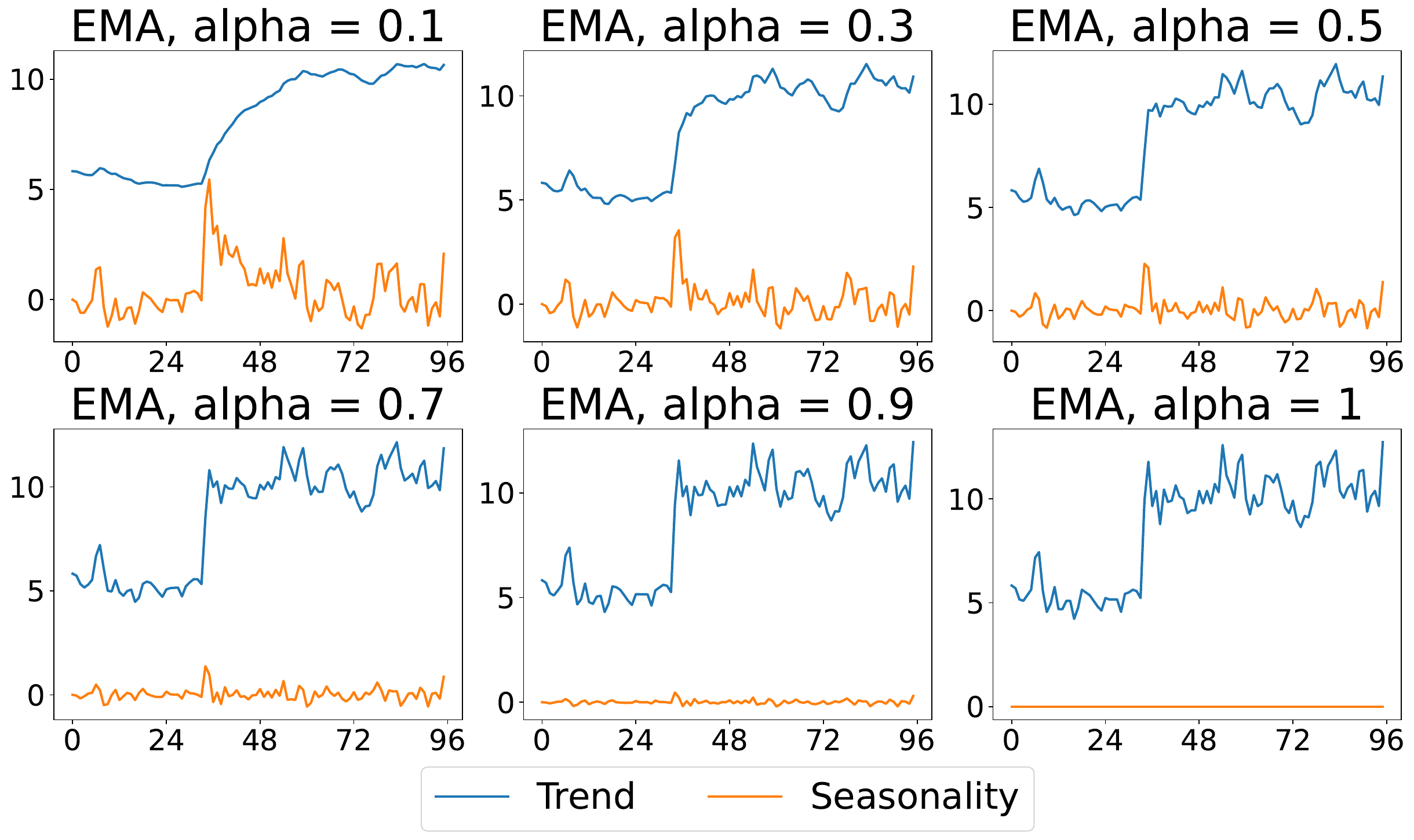}
\caption{Example of EMA decomposition with $\alpha = \{ 0.1, 0.3, 0.5, 0.7, 0.9, 1 \}$ on a 96-length sample from the ETTh1 dataset.}
\label{fig:ema}
\end{figure}

EMA point ${s_t}$ of data ${x_t}$ beginning at time $t=0$ is represented by:
\begin{equation}
\begin{aligned}
\label{equ:ema}
s_0 &= x_0 \\
s_t &= \alpha x_t+(1-\alpha)s_{t-1},\ t>0 \\
X_T &= \text{EMA}(X) \\
X_S &= X - X_T
\end{aligned}
\end{equation}
where $\alpha$ is the smoothing factor, $0<\alpha<1$, $\text{EMA}(\cdot)$ denotes exponential moving average, while $X_T$ and $X_S$ correspond to trend and seasonality components.
Figure \ref{fig:ema} shows an example of EMA decomposition.


The exponential method offers greater control over the behavior of both trend and seasonality components.
Given that data can exhibit diverse patterns, including stationary and non-stationary characteristics with varying periods and behaviors, the adaptability of exponential decomposition provides advantages in feature extraction (see Appendix \ref{app:sma}).
Compared to SMA, EMA presents a more flexible approach to decomposition, as it adjusts its weighting scheme based on the exponential decay of data points.
This adaptability allows EMA to capture changing trends more effectively, making it particularly suitable for time series with dynamic and evolving patterns (see Appendix \ref{app:ema-stat}).

\begin{figure*}[ht]
\centering
\includegraphics[width=2\columnwidth]{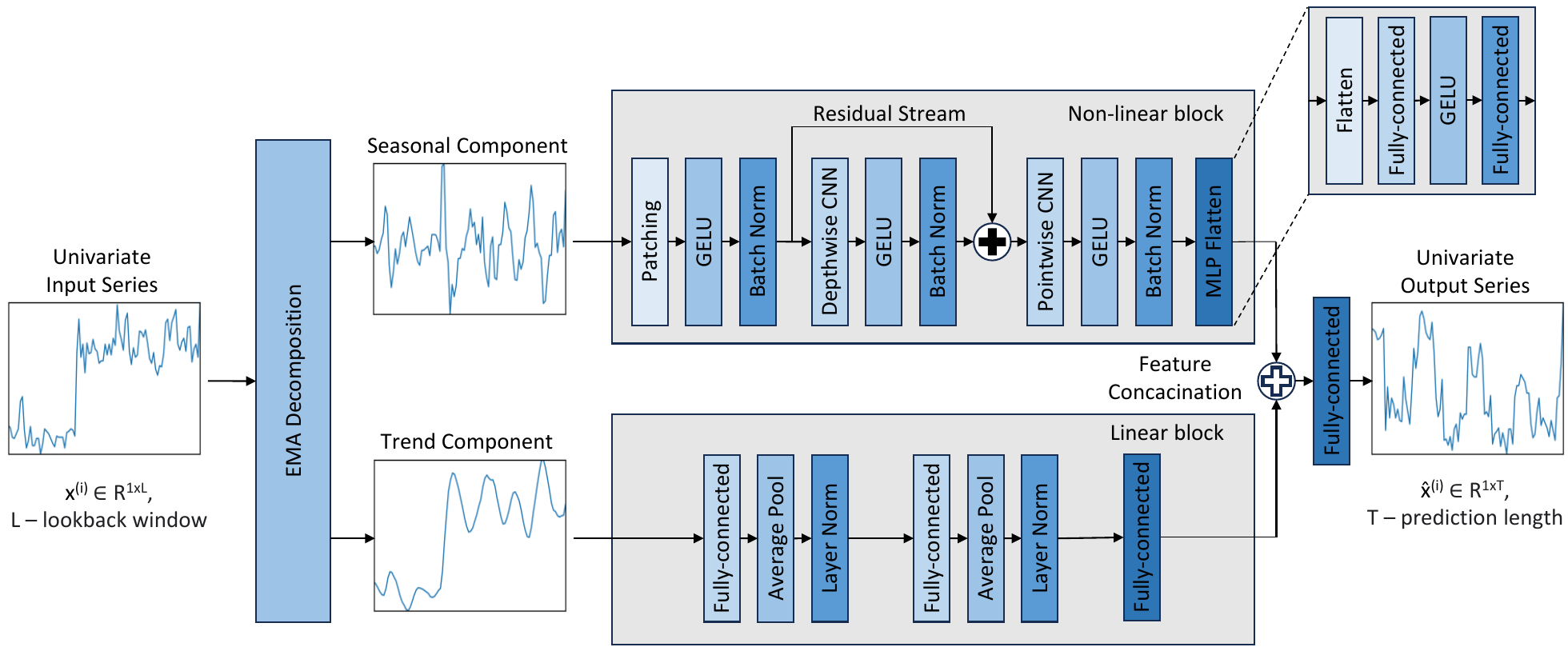}
\caption{xPatch Model Overview.
Every univariate series is passed through exponential decomposition.
Consequently, the trend and seasonal components are processed through the dual flow network.
}
\label{fig:overview}
\end{figure*}

\subsection{Model Architecture}
\textbf{Channel-Independence}.
The multivariate time series $x = (x_1, x_2, ... , x_L)$ is divided into $M$ univariate sequences $x^{(i)} = (x_1^{(i)}, x_2^{(i)}, ... , x_L^{(i)})$, where $x^{(i)} \in \mathbb{R}^L$ and $L$ is lookback of recent historical data points.
Each of these univariate series is then individually fed into the backbone model, which consequently generates a prediction sequence $\hat{x}^{(i)} = (\hat{x}_{L+1}^{(i)}, \hat{x}_{L+2}^{(i)}, ... , \hat{x}_{L+T}^{(i)})$, where $\hat{x}^{(i)} \in \mathbb{R}^T$ and $T$ is future steps observations.
This partitioning approach has proven to work well in both linear models and transformers \cite{zeng2023transformers, yuqietal2023patch, han2023capacity}.


\textbf{Exponential Decomposition}.
Using the EMA method, we decompose each univariate series into trend and seasonality components, which are then processed separately by the dual-flow architecture.
After processing, the learned trend and seasonal features are aggregated and passed to the final output layer to comprise the final prediction as illustrated in Figure \ref{fig:overview}.
Details on optimization and ablation studies of EMA are available in Appendix \ref{app:optimization}, \ref{app:ema-extended}.

\textbf{Dual Flow Net}.
As the main backbone, we employ two distinct flows to analyze trend and seasonality: linear and non-linear streams.
The trend component is processed through the linear MLP-based stream, while the seasonal component is handled by the non-linear CNN-based block.

Seasonality represents periodic fluctuations around a constant level, meaning that the statistical properties of these fluctuations, such as mean and variance, remain stable over time, meaning that the seasonal component is stationary.
In contrast, the trend reflects long-term progression with either increasing or decreasing behavior and a changing mean, which makes the trend component non-stationary.

To summarize, in most cases, the seasonal component is non-linear and stationary, while the trend component is linear and non-stationary.
However, some datasets might exhibit unusual behavior, such as a stationary trend.
Therefore, the dual-stream architecture is designed to enhance the model's adaptability to both stationary and non-stationary data.
For the exploration of the dual-flow architecture, see Appendix \ref{app:dual-flow}.


\textbf{Linear Stream.}
The linear stream is an MLP-based network that includes average pooling and layer normalization, intentionally omitting activation functions to emphasize linear features.

The decomposed data $x^{(i)}$ is processed through two linear blocks, each consisting of a fully connected layer followed by average pooling with a kernel $k=2$ for feature smoothing and layer normalization for training stability.
Each linear layer and average pooling operation contribute to dimensionality reduction, encouraging the network to compress feature representations to fit the available space effectively.
This reduction in the number of features, combined with the absence of activation functions and a bottleneck architecture, aims to retain only the most significant linear features of the smoothed trend.
\begin{flalign}
x^{(i)} = \text{LayerNorm}(\text{AvgPool}(\text{Linear}(x^{(i)}), k=2))
\end{flalign}
The final expansion layer takes the bottleneck representation and upscales it to the prediction length.
\begin{flalign}
\label{equ:linear}
\hat{x}^{(i)}_{lin} = \text{Linear}(x^{(i)})
\end{flalign}

\textbf{Patching.}
Patching is a technique inspired by the vision transformer \cite{dosovitskiy2020vit} and was first introduced in the context of LTSF by PatchTST \cite{yuqietal2023patch}.
This method unfolds each univariate time series using a sliding window.
We incorporate patching into the non-linear block to emphasize repetitive seasonal features.
By using patching, the model can better focus on these repetitive patterns, effectively capturing their inter-pattern dependencies more effectively.

The patch length is denoted as $P$, and the non-overlapping region between two consecutive patches is referred to as stride $S$.
We apply patching in the non-linear stream to each normalized univariate decomposed sequence $x^{(i)} \in \mathbb{R}^L$, which generates a sequence of $N$ 2D patches $x_p^{(i)} \in \mathbb{R} ^ {N \times P}$.
The number of patches is calculated as $N = \lfloor \frac{L-P}{S} \rfloor + 2$.
In our implementation, for a fair comparison with PatchTST and CARD, we adopt their setup for patch embedding, setting $P = 16$ and $S = 8$.

\textbf{Non-linear Stream.}
The non-linear stream is a CNN-based network that introduces non-linearity through activation functions.
By applying convolutions on top of patching, the CNN-based stream captures spatio-temporal patterns and inter-patch correlations, focusing on the non-linear features of the seasonal signal.

First, the patched data $x_p^{(i)} \in \mathbb{R} ^ {N \times P}$ is embedded for increasing the number of features with activation function $\sigma$ and batch normalization \cite{ioffe2015batch}.
Since the seasonal variations have many zero values, we employ GELU \cite{hendrycks2016gaussian} as an activation function for its smooth transition around zero and non-linearity.
The resulting embedded shape is denoted as $x^{N \times P^2}_p$.
\begin{flalign}
x^{N \times P^2}_p = \text{BatchNorm}(\sigma (\text{Embed}(x_p^{(i)})))
\end{flalign}
Following embedding, the data is processed through depthwise separable convolution.
This method splits the computation into two steps: depthwise convolution applies a single convolutional filter per input channel, and pointwise convolution creates a linear combination of the output of the depthwise convolution, with an additional residual stream between them.

Given that the xPatch architecture leverages channel-independence, it was determined to employ patching to increment the number of dimensions, enabling patches to function as channels in the data $x^{N \times P^2}_p$.
Consequently, rather than relying on inter-channel feature representations, we utilize channel-independent inter-patch representations.
This approach aims to capture comprehensive semantic information that may not be available at the point level and allows to focus on non-linear features.

For depthwise convolution, we employ grouped convolution with the number of groups $g$ equal to the number of patches $N$, a large kernel size $k$ equal to the patch length $P$, and a convolution stride $s$ equal to the patch length $P$.
\begin{flalign}
x^{N \times P}_p &= \text{Conv}_{N \rightarrow N}(x^{N \times P^2}_p,k=P,s=P,g=N) \\
x^{N \times P}_p &= \text{BatchNorm}(\sigma ( x^{N \times P}_p))
\end{flalign}
Depthwise convolution applies a single convolutional filter per input channel, generating $N$ feature maps, each corresponding to a specific patch.
This approach enables the model to capture temporal features with group convolution that is consistent for periodic patches.

Subsequently, the data is updated with a linear residual connection spanning the depthwise convolution.
Although depthwise convolution captures temporal relations between periodic patterns, it may not effectively capture inter-patch feature correlations.
Therefore, the sequence is further processed through the pointwise convolution layer with the number of groups $g=1$, a small kernel size $k=1$, and a convolution stride $s=1$.
\begin{flalign}
x^{N \times P}_p &= \text{DepthwiseConv}(x^{N \times P^2}_p) + x^{N \times P^2}_p \\
x^{N \times P}_p &= \text{Conv}_{N \rightarrow N}(x^{N \times P}_p,k=1,s=1,g=1) \\
x^{N \times P}_p &= \text{BatchNorm}(\sigma ( x^{N \times P}_p))
\end{flalign}
Pointwise convolution creates a linear combination of the output and aggregates features across different patches without skipping elements.

These predictions are then processed through the MLP flatten layer.
This layer is designed in a similar style to PatchTST: the first linear layer doubles the hidden dimension, while the second linear layer projects it back with a GELU activation function between them.
\begin{flalign}
\label{equ:non-lin}
\hat{x}^{(i)}_{nonlin} = \text{Linear}(\sigma (\text{Linear}(\text{Flatten}(x^{N \times P}_p))))
\end{flalign}
Finally, linear features (\ref{equ:linear}) and non-linear features (\ref{equ:non-lin}) are concatenated and fed into the final linear layer, which merges linear and non-linear features for the output prediction.
\begin{flalign}
\hat{x}^{(i)} = \text{Linear}(\text{concat}( \hat{x}^{(i)}_{lin}, \hat{x}^{(i)}_{nonlin} ))
\end{flalign}
We concatenate the linear and non-linear features from the two flows, representing learned representations from the MLP and CNN streams.
This mechanism enables the model to dynamically weigh the significance of both linear and non-linear features in the final prediction, providing adaptability to diverse patterns in time series data.

\begin{table*}[th]
  \scriptsize
  \centering
  \setlength\tabcolsep{3pt}
  \begin{tabular}{c|cc|cc|cc|cc|cc|cc|cc|cc|cc|cc}
    \hline
    \multirow{2}{*}{Models}                                                     &
    \multicolumn{2}{|c}{\textbf{xPatch}}    & \multicolumn{2}{|c}{CARD}         &
    \multicolumn{2}{|c}{TimeMixer}          & \multicolumn{2}{|c}{iTransformer} & 
    \multicolumn{2}{|c}{RLinear}            & \multicolumn{2}{|c}{PatchTST}     & 
    \multicolumn{2}{|c}{MICN}               & \multicolumn{2}{|c}{DLinear}      &
    \multicolumn{2}{|c}{TimesNet}           & \multicolumn{2}{|c}{ETSformer}    \\
    \multirow{2}{*}{}                                                           &
    \multicolumn{2}{|c}{\textbf{(ours)}}    & \multicolumn{2}{|c}{(2024)}       & 
    \multicolumn{2}{|c}{(2024)}             & \multicolumn{2}{|c}{(2024)}       & 
    \multicolumn{2}{|c}{(2023)}             & \multicolumn{2}{|c}{(2023)}       & 
    \multicolumn{2}{|c}{(2023)}             & \multicolumn{2}{|c}{(2023)}       &
    \multicolumn{2}{|c}{(2023)}             & \multicolumn{2}{|c}{(2022)}       \\
    \hline
    Metric
    & MSE & MAE & MSE & MAE & MSE & MAE & MSE & MAE & MSE & MAE & MSE & MAE & MSE & MAE & MSE & MAE & MSE & MAE & MSE & MAE \\
    \hline
    ETTh1
    & \textbf{0.428} &	\textbf{0.419} &	0.442 &	0.429 &	0.447 &	0.44 &	0.454 &	0.448 &	\underline{0.438} &	\underline{0.427} &	0.45 &	0.441 &	0.559 &	0.535 &	0.456 &	0.452 &	0.458 &	0.450 &	0.542 &	0.510 \\
    \hline
    ETTh2
    & \textbf{0.319} &	\textbf{0.361} &	0.368 &	\underline{0.390} &	0.365 &	0.395 &	0.383 &	0.407 &	\underline{0.362} &	0.394 &	0.365 &	0.394 &	0.588 &	0.525 &	0.559 &	0.515 &	0.414 &	0.427 &	0.439 &	0.452 \\
    \hline
    ETTm1
    & \textbf{0.377} &	\underline{0.384} &	0.382 &	\textbf{0.383} &	\underline{0.381} &	0.396 &	0.407 &	0.410 &	0.409 &	0.401 &	0.383 &	0.394 &	0.392 &	0.414 &	0.403 &	0.407 &	0.400 &	0.406 &	0.429 &	0.425 \\
    \hline
    ETTm2
    & \textbf{0.267} &	\textbf{0.313} &	\underline{0.272} &	\underline{0.317} &	0.275 &	0.323 &	0.288 &	0.332 &	0.286 &	0.328 &	0.284 &	0.327 &	0.328 &	0.382 &	0.350 &	0.401 &	0.291 &	0.333 &	0.293 &	0.342 \\
    \hline
    Weather
    & \textbf{0.232} &	\textbf{0.261} &	\underline{0.239} &	\underline{0.265} &	0.240 &	0.272 &	0.258 &	0.278 &	0.269 &	0.289 &	0.257 &	0.280 &	0.243 &	0.299 &	0.265 &	0.317 &	0.259 &	0.287 &	0.271 &	0.334 \\
    \hline
    Traffic
    & 0.499 &	\textbf{0.279} &	\underline{0.453} &	\underline{0.282} &	0.485 &	0.298 &	\textbf{0.428} &	\underline{0.282} &	0.623 &	0.372 &	0.467 &	0.292 &	0.542 &	0.316 &	0.625 &	0.383 &	0.620 &	0.336 &	0.621 &	0.396 \\
    \hline
    Electricity
    & 0.179 &	\underline{0.264} &	\textbf{0.168} &	\textbf{0.258} &	0.182 &	0.273 &	\underline{0.178} &	0.270 &	0.214 &	0.291 &	0.190 &	0.275 &	0.187 &	0.295 &	0.212 &	0.300 &	0.193 &	0.295 &	0.208 &	0.323 \\
    \hline
    Exchange
    & 0.375 &	0.408 &	0.360 &	\underline{0.402} &	0.408 &	0.422 &	0.360 &	0.403 &	0.380 &	0.410 &	0.364 &	\textbf{0.400} &	\textbf{0.315} &	0.404 &	\underline{0.354} &	0.414 &	0.416 &	0.443 &	0.410 &	0.427 \\
    \hline
    Solar
    & 0.239 &	\textbf{0.236} &	0.237 &	\underline{0.239} &	\textbf{0.216} &	0.280 &	\underline{0.233} &	0.262 &	0.369 &	0.357 &	0.254 &	0.289 &	0.283 &	0.358 &	0.327 &	0.398 &	0.301 &	0.319 &	0.603 &	0.615 \\
    \hline
    ILI
    & \textbf{1.442} &	\textbf{0.725} &	1.916 &	0.842 &	1.708 &	0.820 &	2.918 &	1.154 &	2.452 &	0.978 &	\underline{1.626} &	\underline{0.804} &	2.664 &	1.086 &	2.616 &	1.090 &	2.139 &	0.931 &	2.497 &	1.004 \\
    \hline
    \end{tabular}
    \caption{Averaged long-term forecasting results with unified lookback window $L = 36$ for the ILI dataset, and $L = 96$ for all other datasets.
  All results are averaged from 4 different prediction lengths: $T = \{24,36,48,60\}$ for the ILI dataset, and $ T = \{96,192,336,720\}$ for all other datasets, respectively.
  The best model is \textbf{boldface} and the second best is \underline{underlined}.
  See Table \ref{tab:full-experiments} in Appendix \ref{app:full-experiments} for the full results.}
    \label{tab:experiments}
\end{table*}

\subsection{Loss Function}
Mean Squared Error (MSE) loss is a training loss scheme commonly used by LTSF models.
The MSE loss $\mathcal{L}_{\text{MSE}}$ between the predicted univariate sequence $\hat{x}_{1:T}^{(i)}$ and the ground truth observations $x_{1:T}^{(i)}$, where $T$ is future prediction length, is denoted as:
\begin{flalign}
\mathcal{L}_{\text{MSE}}=\frac{1}{T}\sum_{i=1}^T ||\hat{x}_{1:T}^{(i)} - x_{1:T}^{(i)}||_2^2
\end{flalign}
The recent transformer-based model CARD \cite{wang2024card} introduced a novel signal decay-based loss function, where they scale down the far-future Mean Absolute Error (MAE) loss to address the high variance.
MAE was chosen since it is more resilient to outliers than MSE.
\begin{flalign}
\label{equ:card_loss}
\mathcal{L}_{\text{CARD}} = \frac{1}{T} \sum _{i=1}^T i^{-\frac{1}{2}} || \hat{x}_{1:T}^{(i)} - x_{1:T}^{(i)} ||
\end{flalign}
where $i$ corresponds to the prediction point in the future.
This training scheme was proven by CARD to be efficient and to increase the performance of existing models.

To identify a more effective scaling loss coefficient, we extend Equation (\ref{equ:card_loss}) to a universally applicable MAE scalable loss function:
\begin{flalign}
\label{equ:scalable_loss}
\mathcal{L} = \frac{1}{T} \sum _{i=1}^T \rho(i) || \hat{x}_{1:T}^{(i)} - x_{1:T}^{(i)} ||
\end{flalign}
where $\rho(i)$ represents the scaling coefficient.
Thus, the $\mathcal{L}_{\text{CARD}}$ loss defined in Equation (\ref{equ:card_loss}) emerges as a specific instance of the scalable loss function delineated in Equation (\ref{equ:scalable_loss}), with $\rho(i) = i^{-\frac{1}{2}}$.

We find that the scaling coefficient $\rho_{CARD}(i) = i^{-\frac{1}{2}}$ exhibits a too rapid decrease rate for our task.
Therefore, we propose a novel arctangent loss $\mathcal{L}_{arctan}$, which features a slower increase rate compared to the exponential functions analyzed in CARD \cite{wang2024card}:
\begin{flalign}
    \mathcal{L}_{arctan} = \frac{1}{T} \sum _{i=1}^T \rho_{arctan}(i) || \hat{x}_{1:T}^{(i)} - x_{1:T}^{(i)} || \\
    \rho_{arctan}(i) = -\arctan(i)+\frac{\pi}{4}+1
\end{flalign}

Mathematical proofs, ablation studies on state-of-the-art models employing the arctangent loss, and the arctangent function's scaling analysis can be found in Appendix \ref{app:arctan_loss}.

\subsection{Learning Rate Adjustment Scheme}
Most recent LTSF models \cite{zhou2021informer, wu2021autoformer, zhou2022fedformer, woo2022etsformer, wu2023timesnet, zeng2023transformers, li2023revisiting, liu2024itransformer} adapt standard learning rate adjustment technique.
Learning rate $\alpha_t$ at epoch $t$ with initial learning rate $\alpha_0$ is calculated as:
\begin{flalign}
\label{equ:standard}
    \alpha_t = \alpha_{t-1} * 0.5^{t-1}, \text{ for } t\ge1
\end{flalign}
This strategy results in a decreasing learning rate with each successive epoch.
Such a rapidly decreasing scheme was effective since the models were trained with a small number of epochs, usually limited to 10.

PatchTST \cite{yuqietal2023patch} introduced a long training approach with an upper limit of 100 epochs and a new learning rate adjustment schedule:
\begin{flalign}
    \alpha_t &= \alpha_0, \text{ for } t<3, \\
\label{equ:patchtst}
    \alpha_t &= \alpha_{t-1} * 0.9^{t-3}, \text{ for } t\ge3
\end{flalign}

Consequently, CARD \cite{wang2024card} developed a new linear warm-up of the model with subsequent cosine learning rate decay.
Learning rate $\alpha_t$ at epoch $t$ with initial learning rate $\alpha_0$, number of warmup epochs $w$, and upper limit of 100 epochs is calculated as:
\begin{flalign}
    \alpha_t &= \alpha_{t-1} * \frac{t}{w}, \text{ for } t<w,\\
\label{equ:cosine}
    \alpha_t &= 0.5\alpha (1+cos(\pi * \frac{(t - w)}{100 - w})), \text{ for } t\ge w
\end{flalign}

We introduce a novel sigmoid learning rate adjustment scheme.
The learning rate $\alpha_t$ at epoch $t$, with an initial learning rate $\alpha_0$, logistic growth rate $k$, decreasing curve smoothing rate $s$, and warm-up coefficient $w$, is calculated as follows:
\begin{flalign}
    \label{equ:sigmoid}
    \alpha_t = \frac{\alpha_0}{1+e^{-k(t-w)}} - \frac{\alpha_0}{1+e^{-\frac{k}{s}(t-sw)}}
\end{flalign}

Mathematical proofs, ablation studies on state-of-the-art models using the sigmoid learning rate adjustment approach, and hyperparameters selection are available in Appendix \ref{app:sigmoid}.

\begin{table*}[th]
  \scriptsize
  \centering
  \setlength\tabcolsep{3pt}
  \begin{tabular}{c|cc|cc|cc|cc|cc|cc|cc|cc|cc|cc}
    \hline
    \multirow{2}{*}{Models}                                                     &
    \multicolumn{2}{|c}{\textbf{xPatch}}    & \multicolumn{2}{|c}{CARD}         &
    \multicolumn{2}{|c}{TimeMixer}          & \multicolumn{2}{|c}{iTransformer} & 
    \multicolumn{2}{|c}{RLinear}            & \multicolumn{2}{|c}{PatchTST}     & 
    \multicolumn{2}{|c}{MICN}               & \multicolumn{2}{|c}{DLinear}      &
    \multicolumn{2}{|c}{TimesNet}           & \multicolumn{2}{|c}{ETSformer}    \\
    \multirow{2}{*}{}                                                           &
    \multicolumn{2}{|c}{\textbf{(ours)}}    & \multicolumn{2}{|c}{(2024)}       & 
    \multicolumn{2}{|c}{(2024)}             & \multicolumn{2}{|c}{(2024)}       & 
    \multicolumn{2}{|c}{(2023)}             & \multicolumn{2}{|c}{(2023)}       & 
    \multicolumn{2}{|c}{(2023)}             & \multicolumn{2}{|c}{(2023)}       &
    \multicolumn{2}{|c}{(2023)}             & \multicolumn{2}{|c}{(2022)}       \\
    \hline
    Metric
    & MSE & MAE & MSE & MAE & MSE & MAE & MSE & MAE & MSE & MAE & MSE & MAE & MSE & MAE & MSE & MAE & MSE & MAE & MSE & MAE \\
    \hline
    ETTh1
    & \textbf{0.391} &	\textbf{0.412} &	\underline{0.401} & 	\underline{0.422} &	0.411 &	0.423 &	0.501 &	0.492 &	0.413 &	0.427 &	0.413 &	0.434 &	0.440 &	0.462 &	0.423 &	0.437 &	0.458 &	0.450 &	0.542 &	0.510 \\
    \hline
    ETTh2
    & \textbf{0.299} &	\textbf{0.351} &	0.321 & 	\underline{0.373} &	\underline{0.316} &	0.384 &	0.385 &	0.417 &	0.328 &	0.382 &	0.331 &	0.381 &	0.403 &	0.437 &	0.431 &	0.447 &	0.414 &	0.427 &	0.439 &	0.452 \\
    \hline
    ETTm1
    & \textbf{0.341} &	\textbf{0.368} &	0.350 & 	\textbf{0.368} &	\underline{0.348} &	\underline{0.376} &	0.373 &	0.404 &	0.359 &	0.378 &	0.353 &	0.382 &	0.387 &	0.411 &	0.357 &	0.379 &	0.400 &	0.406 &	0.429 &	0.425 \\
    \hline
    ETTm2
    & \textbf{0.242} &	\textbf{0.300} &	0.255 & 	\underline{0.310} &	0.256 &	0.316 &	0.274 &	0.335 &	\underline{0.253} &	0.313 &	0.256 &	0.317 &	0.284 &	0.340 &	0.267 &	0.332 &	0.291 &	0.333 &	0.293 &	0.342 \\
    \hline
    Weather
    & \textbf{0.211} &	\textbf{0.247} &	\underline{0.220} & 	\underline{0.248} &	0.222 &	0.262 &	0.271 &	0.297 &	0.242 &	0.278 &	0.226 &	0.264 &	0.243 &	0.299 &	0.246 &	0.300 &	0.259 &	0.287 &	0.271 &	0.334 \\
    \hline
    Traffic
    & 0.392 &	\textbf{0.248} &	\underline{0.381} & 	\underline{0.251} &	0.388 &	0.263 &	\textbf{0.378} &	0.270 &	0.417 &	0.283 &	0.391 &	0.264 &	0.542 &	0.316 &	0.434 &	0.295 &	0.620 &	0.336 &	0.621 &	0.396 \\
    \hline
    Electricity
    & \textbf{0.153} &	\textbf{0.245} &	0.157 & 	0.251 &	\underline{0.156} &	\underline{0.247} &	0.161 &	0.257 &	0.164 &	0.257 &	0.159 &	0.253 &	0.187 &	0.295 &	0.166 &	0.264 &	0.193 &	0.295 &	0.208 &	0.323 \\
    \hline
    Exchange
    & 0.366 &	0.404 &	0.360 & 	\underline{0.402} &	0.471 &	0.452 &	0.458 &	0.469 &	0.423 &	0.427 &	0.405 &	0.426 &	\underline{0.315} &	0.404 &	\textbf{0.297} &	\textbf{0.378} &	0.416 &	0.443 &	0.410 &	0.427 \\
    \hline
    Solar
    & \underline{0.194} &	\textbf{0.214} &	0.198 & 	\underline{0.225} &	\textbf{0.192} &	0.244 &	0.197 &	0.262 &	0.235 &	0.266 &	0.256 &	0.298 &	0.213 &	0.266 &	0.329 &	0.400 &	0.244 &	0.334 &	0.603 &	0.615 \\
    \hline
    ILI
    & \textbf{1.281} &	\textbf{0.688} &	1.916 & 	0.842 &	1.971 &	0.924 &	2.947 &	1.193 &	1.803 &	0.874 &	\underline{1.480} &	\underline{0.807} &	2.567 &	1.056 &	2.169 &	1.041 &	2.139 &	0.931 &	2.497 &	1.004 \\
    \hline
    \end{tabular}
    \caption{Averaged long-term forecasting results under hyperparameter searching.
  All results are averaged from 4 different prediction lengths: $ T = \{24,36,48,60\}$ for the ILI dataset, and $ T = \{96,192,336,720\}$ for all other datasets, respectively.
  The best model is \textbf{boldface} and the second best is \underline{underlined}.
  See Table \ref{tab:full-experiments-search} in Appendix \ref{app:full-experiments} for the full results.}
    \label{tab:experiments2}
\end{table*}

\section{Experiments}
\textbf{Datasets.} We conduct extensive experiments on nine real-world multivariate time series datasets, including Electricity Transform Temperature (ETTh1, ETTh2, ETTm1, ETTm2) \cite{zhou2021informer}, Weather, Traffic, Electricity, Exchange-rate, ILI \cite{wu2021autoformer}, and Solar-energy \cite{lai2018modeling}.

\textbf{Evaluation Metrics.} Following previous works, we use Mean Squared Error (MSE) and Mean Absolute Error (MAE) metrics to assess the performance.

\textbf{Implementation Details.}
All the experiments are implemented in PyTorch \cite{paszke2019pytorch}, and conducted on a single Quadro RTX 6000 GPU.

\textbf{Baselines.} We choose the last state-of-the-art LTSF models, including Autoformer (2021) \cite{wu2021autoformer}, FEDformer (2022) \cite{zhou2022fedformer}, ETSformer (2022) \cite{woo2022etsformer}, TimesNet (2023) \cite{wu2023timesnet}, DLinear (2023) \cite{zeng2023transformers}, RLinear (2023) \cite{li2023revisiting}, MICN (2023) \cite{wang2023micn}, PatchTST (2023) \cite{yuqietal2023patch}, iTransformer (2024) \cite{liu2024itransformer}, TimeMixer (2024) \cite{wang2024timemixer}, and CARD (2024) \cite{wang2024card} as baselines for our experiments.

\textbf{Unified Experimental Settings.} To ensure a fair comparison, we conduct 2 types of experiments.
The first experiment uses unified settings based on the forecasting protocol proposed by TimesNet \cite{wu2023timesnet}: a lookback length $L=36$, prediction lengths $T = \{ 24, 36, 48, 60 \}$ for the ILI dataset, and $L=96$, $T = \{ 96, 192, 336, 720 \}$ for all other datasets.
The averaged results are reported in Table \ref{tab:experiments}.

To handle data heterogeneity and distribution shift, we apply reversible instance normalization \cite{kim2021reversible}.
In Appendix \ref{app:revin}, we examine the impact of instance normalization on the forecasting results of xPatch and other state-of-the-art models, comparing their performance with and without the RevIN module.


\textbf{Hyperparameter Search.} In the second experiment, we aim to determine the upper bounds of the compared models and conduct a hyperparameter search.
We evaluate all models to see if they benefit from longer historical data to identify the optimal lookback length for each, as detailed in Appendix \ref{app:lookback}.
For the models that benefit from a longer input length, namely xPatch, CARD, TimeMixer, iTransformer, RLinear, PatchTST, and DLinear, we perform a hyperparameter search similar to TimeMixer \cite{wang2024timemixer}.
The averaged results are reported in Table \ref{tab:experiments2}.


All implementations are derived from the models' official repository code, maintaining the same configurations.
It is also important to note that we strictly adhere to the settings specified in the official implementations, including the number of epochs (100 for CARD and PatchTST, 15 for RLinear) and the learning rate adjustment strategy.

\textbf{Results.} In the unified experimental settings, xPatch achieves the best averaged performance on 60\% of the datasets using the MSE metric and 70\% of the datasets using the MAE metric.
Compared to CARD, xPatch surpasses it by 2.46\% in MSE and 2.34\% in MAE.
Compared to TimeMixer, xPatch surpasses it by 3.34\% in MSE and 6.34\% in MAE.
Compared to PatchTST, xPatch surpasses it by 4.76\% in MSE and 6.20\% in MAE.

In the hyperparameter search settings, xPatch achieves the best averaged performance on 70\% of the datasets using the MSE metric and 90\% of the datasets using the MAE metric.
Compared to CARD, xPatch surpasses it by 5.29\% in MSE and 3.81\% in MAE.
Compared to TimeMixer, xPatch surpasses it by 7.45\% in MSE and 7.85\% in MAE.
Compared to PatchTST, xPatch surpasses it by 7.87\% in MSE and 8.59\% in MAE.

\textbf{Computational Cost.} While it is true that the proposed dual-flow architecture incurs higher computational costs compared to single-stream CNN and MLP models, it is important to note that convolution and linear operations are initially not as computationally expensive as transformer-based solutions.
The overall increase in computational costs remains relatively small, as shown in Table \ref{tab:comp}.
Moreover, the enhanced performance of the introduced dual-stream architecture outweighs these additional computational costs.

\begin{table}[th]
  \centering
  \begin{tabular}{c|c|c}
    \hline
    Method      & Training time & Inference time \\
    \hline
    MLP-stream  & 0.948 msec    & 0.540 msec \\
    \hline
    CNN-stream  & 1.811 msec    & 0.963 msec \\
    \hline
    xPatch      & 3.099 msec    & 1.303 msec \\
    \hline
    CARD        & 14.877 msec   & 7.162 msec \\
    \hline
    TimeMixer   & 13.174 msec   & 8.848 msec \\
    \hline
    iTransformer& 6.290 msec    & 2.743 msec \\
    \hline
    PatchTST    & 6.618 msec    & 2.917 msec \\
    \hline
    DLinear     & 0.420 msec    & 0.310 msec \\
    \hline
    \end{tabular}
  \caption{The average per step running and inference time maintaining the same settings for all benchmarks.}
    \label{tab:comp}
\end{table}

\section{Conclusion}
This study introduces xPatch, a novel dual-flow architecture for long-term time series forecasting (LTSF).
xPatch combines the strengths of both Convolutional Neural Networks (CNNs) and Multi-Layer Perceptrons (MLPs) to achieve superior performance.
Our findings demonstrate that the integration of an Exponential Moving Average (EMA) seasonal-trend decomposition module effectively captures underlying trends and enhances forecasting accuracy.
The dual-stream network further enhances xPatch's adaptability by dynamically weighing the importance of linear and non-linear features for diverse time series patterns.
Additionally, this study introduces a robust arctangent loss function and a novel sigmoid learning rate adjustment approach, both of which consistently improve the performance of existing models.
By investigating patching and channel-independence within a CNN-based backbone, xPatch offers a compelling alternative to transformer-based architectures, achieving superior performance while maintaining computational efficiency.

\section*{Acknowledgements}
We would like to thank Enver Menadjiev, Kyowoon Lee, Jihyeon Seong, Jiyeon Han and the anonymous reviewers for their valuable comments.

This work was supported by NAVER, Institute of Information \& Communications Technology Planning \& Evaluation (IITP), and the Korean Ministry of Science and ICT (MSIT) under the grant agreement No. RS-2019-II190075, Artificial Intelligence Graduate School Program (KAIST); No. RS-2022-II220984, Development of Artificial Intelligence Technology for Personalized Plug-and-Play Explanation and Verification of Explanation; No.RS-2022-II220184, Development and Study of AI Technologies to Inexpensively Conform to Evolving Policy on Ethics.

\bibliography{xPatch}

\newpage
\clearpage
\appendix
\section*{Appendix}
\section{Datasets}
We conduct experiments on nine real-world multivariate time series datasets to evaluate the performance of the proposed xPatch model:
\begin{itemize}
    \item Electricity Transformer Temperature (ETT)\footnote{\url{https://github.com/zhouhaoyi/ETDataset}}: has four subsets, where ETTh1 and ETTh2 are recorded every hour, while ETTm1 and ETTm2 are recorded every 15 minutes.
    Data is collected from two different electric transformers.
    \item Weather\footnote{\url{https://www.bgc-jena.mpg.de/wetter}}: collects 21 meteorological indicators in Germany, such as humidity and air temperature, collected every 10 minutes from the Weather Station of the Max Planck Biogeochemistry Institute in 2020.
    \item Traffic\footnote{\url{https://pems.dot.ca.gov}}: records hourly road occupancy rates measured by 862 sensors on San Francisco Bay area freeways from January 2015 to December 2016.
    \item Electricity\footnote{\url{https://archive.ics.uci.edu/dataset/321/electricityloaddiagrams20112014}}: describes the hourly electricity consumption data of 321 clients from 2012 to 2014.
    \item Exchange-rate\footnote{\url{https://github.com/laiguokun/multivariate-time-series-data}}: collects the panel data of daily exchange rates from 8 countries from 1990 to 2016.
    \item Solar-energy\footnote{\url{https://www.nrel.gov/grid/solar-power-data.html}}: contains the solar power production records from 137 PV plants in 2006.
    \item ILI\footnote{\url{https://gis.cdc.gov/grasp/fluview/fluportaldashboard.html}}: records the weekly number of patients and influenza-like illness ratio in the USA between 2002 and 2021.
\end{itemize}

Due to the size, the prediction length of the ILI dataset is \{24, 36, 48, 60\}, while for all other datasets prediction length is set to \{96, 192, 336, 720\}.
Table \ref{tab:datasets} summarizes details of statistics of datasets.
\begin{table}[th]
  \centering
    \setlength\tabcolsep{3pt}
  \begin{tabular}{c|c|c|c}
    \hline
    Dataset & Dim & Dataset Size & Frequency \\
    \hline
    ETTh1,ETTh2 & 7 & (8545,2881,2881) & Hourly \\
    \hline
    ETTm1,ETTm2 & 7 & (34465,11521,11521) & 15 min \\
    \hline
    Weather & 21 & (36792,5271,10540) & 10 min \\
    \hline
    Traffic & 862 & (12185,1757,3509) & Hourly \\
    \hline
    Electricity & 321 & (18317,2633,5261) & Hourly \\
    \hline
    Exchange-rate & 8 & (5120,665,1422) & Daily \\
    \hline
    Solar-energy & 137 & (36792,5271,10540) & 10 min \\
    \hline
    ILI & 7 & (617,74,170) & Weekly \\
    \hline
    \end{tabular}
  \caption{Detailed dataset descriptions.
  Dim denotes dimension, which is the variate number of each dataset.
  Dataset size denotes the total number of time points in (Train, Validation, Test) split.
  Frequency denotes the sampling interval of time points.
    }
    \label{tab:datasets}
\end{table}

\section{Seasonal-Trend Decomposition}
\label{app:sma}
\begin{figure}[ht]
     \centering
     \begin{subfigure}[b]{0.22\textwidth}
         \centering
         \includegraphics[width=1\columnwidth]{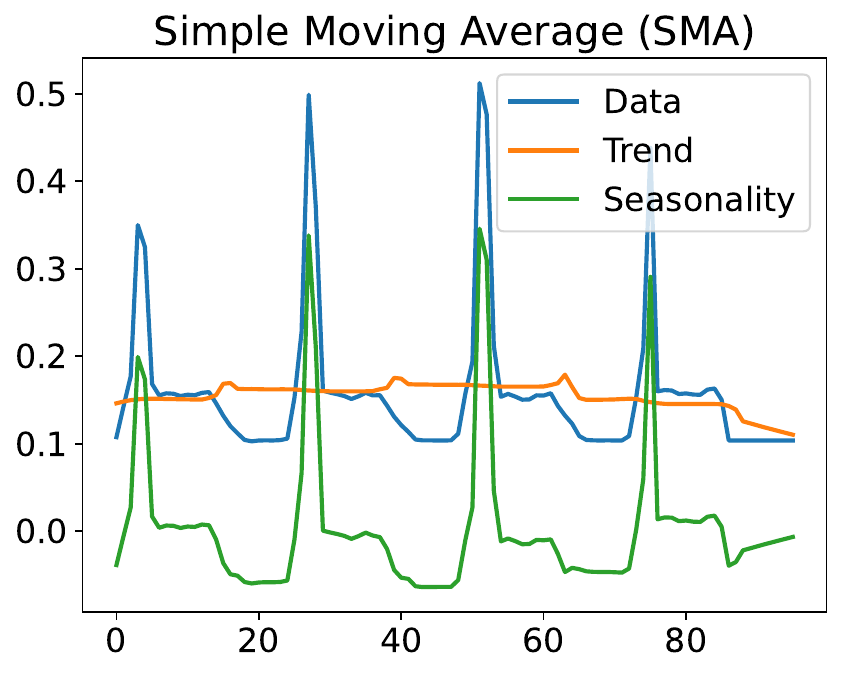}
         \caption{SMA decomposition.}
         \label{ex2}
     \end{subfigure}
     \begin{subfigure}[b]{0.23\textwidth}
         \centering
         \includegraphics[width=1\columnwidth]{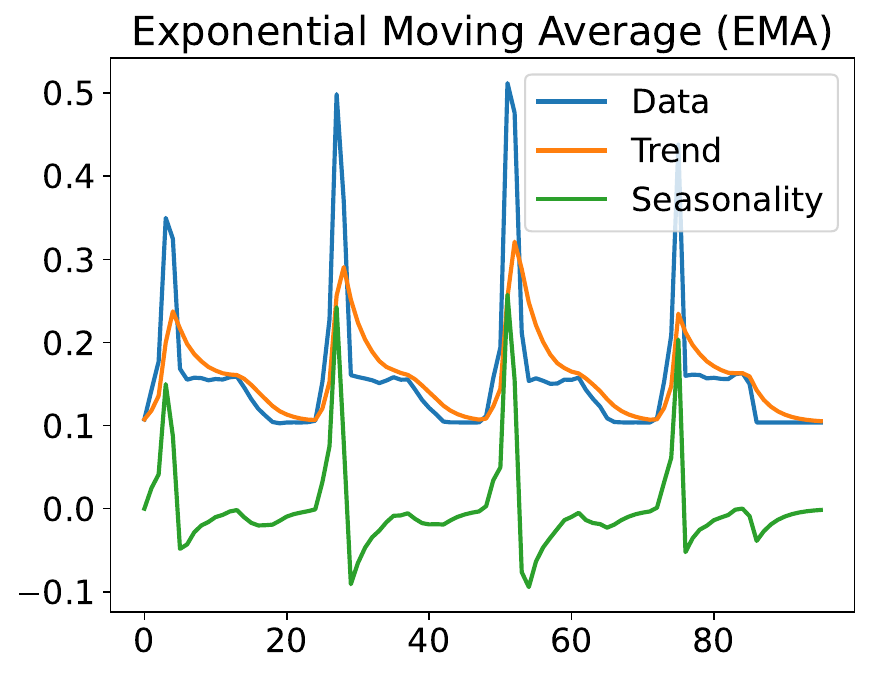}
         \caption{EMA decomposition.}
         \label{ex3}
     \end{subfigure}
        \caption{SMA and EMA smoothing and decomposition on a 96-length sample from the Traffic dataset.}
        \label{ex1}
\end{figure}

For a qualitative comparison between the Simple Moving Average (SMA) introduced in Equation (\ref{equ:sma}) and the Exponential Moving Average (EMA) introduced in Equation (\ref{equ:ema}), we provide a sample from the Traffic dataset in Figure \ref{ex1}.
It is evident that when data exhibits spikes and waving patterns, SMA struggles to extract significant trend features.
Due to its use of average pooling, SMA is unable to capture spikes since the average of a sliding window fails to account for sudden peaks.
The increasing curves are observed after the actual spike, leaving the extremely changing features to represent seasonality.
Conversely, EMA effectively smooths the data, highlighting appropriate trend features.

Figure \ref{ex2} presents an example of SMA decomposition that fails to produce interpretable trend and seasonality patterns.
The seasonal pattern shows minimal change from the initial data, primarily exhibiting a vertical shift.

In contrast, Figure \ref{ex3} shows EMA decomposition on the same sample.
EMA effectively decomposes the sample into smoothed trend features and discernible seasonal variations.

\section{Stationarity of EMA}
\label{app:ema-stat}
The goal of seasonal-trend decomposition is to split data into simpler components with distinct features: a stationary seasonality and a non-stationary trend.
To analyze the impact of decomposition on the stationarity of specific segments, we conduct experiments using the Augmented Dickey-Fuller (ADF) \cite{dickey1979distribution} stationarity test.

We select a window size $L=720$, as it is the longest lookback window analyzed in the experimental settings.
We divide the dataset into chunks of length L, consistent with how forecasting is performed, and apply the proposed EMA decomposition to each chunk with $\alpha = 0.3$.
Table \ref{tab:adf} illustrates the effect of decomposition on data stationarity:

\begin{table}[th]
  \scriptsize
  \centering
  \begin{tabular}{c|c|c|c|c|c|c}
    \hline
    Dataset & ADF & Stat & Trend ADF & Stat & Seasonal ADF & Stat \\
    \hline
    ETTh1 & 0.131 & NS & 0.157 & NS & $0.369 * 10^{-6}$ & S \\
    ETTh2 & 0.370 & NS & 0.198 & NS & $0.447 * 10^{-6}$ & S \\
    Electricity & 0.165 & NS & 0.134 & NS & $0.567 * 10^{-9}$ & S \\
    \hline
    \end{tabular}
  \caption{Stationarity of initial data and components decomposed by EMA.
  ADF for mean ADF p-value, NS for non-stationary, and S for stationary.}
    \label{tab:adf}
\end{table}


The primary objective of decomposition is to isolate a non-stationary trend component, making the remaining seasonality component a stationary sub-series.
To compare the EMA and SMA decompositions, we use the Traffic dataset, which is the longest and the most complex among the benchmark datasets.
The complexity of the Traffic dataset stems from its highly fluctuating data, where the seasonality pattern is much stronger than the trend.
For this reason, the entire dataset is initially classified as stationary by the ADF test.

Table \ref{tab:adfema} compares the stationarity of the whole dataset with the trend-only components decomposed by SMA and EMA with varying $\alpha$ values:

\begin{table}[th]
  \scriptsize
  \centering
    \setlength\tabcolsep{5pt}
  \begin{tabular}{c|c|c|c|c|c}
    \hline
    Chunks & Dataset & SMA & EMA (0.1) & EMA (0.3) & EMA (0.5) \\
    \hline
    ADF (L=720) $\uparrow$ & 0.029 & 0.034 & 0.064 & \textbf{0.219} & 0.162 \\
    S (max 25) $\downarrow$ & 20 & 22 & 10 & \textbf{3} & 5 \\
    \hline
    \end{tabular}
  \caption{Mean ADF p-values for trend components filtered by the SMA and EMA with $\alpha=\{0.1, 0.3, 0.5\}$ decompositions.
  The dataset is Traffic.
  S for the number of stationary chunks of length $L = 720$.
  ADF p-value $< 0.05$ indicates stationarity.}
    \label{tab:adfema}
\end{table}

The mean ADF p-value for the entire dataset is $0.029$, with 20 out of 25 regions classified as stationary.
This suggests that the Traffic dataset is stationary according to the ADF test.
However, the dataset also contains non-stationary trend features, which are weaker than the seasonality component.
Therefore, the objective of decomposition is to extract even weak non-stationary characteristics into the trend component to enhance forecastability.

The results reveal that SMA decomposition fails to capture meaningful trend patterns, as most trend segments remain stationary.
In contrast, EMA effectively isolates the weak non-stationary trend from stationary seasonality.
Additionally, this experiment demonstrates that $\alpha = 0.3$ captures trend features optimally.

\section{EMA Optimization}
\label{app:optimization}
The straightforward implementation of EMA introduced in Equation (\ref{equ:ema}) requires a for loop, which is $O(n)$ time complexity.
Therefore, we are aiming to optimize the EMA decomposition module to constant time complexity.
The equation can be derived as the following sequence:

\begin{equation}
\begin{aligned}
s_t &= \alpha x_t+(1-\alpha)s_{t-1}& \\
&    = \alpha x_t+(1-\alpha) (\alpha x_{t-1}+(1-\alpha)s_{t-2})& \\
&    = \alpha x_t+(1-\alpha)\alpha x_{t-1} + (1-\alpha)^2 s_{t-2}& \\
&    = ... + (1-\alpha)^2 (\alpha x_{t-2}+(1-\alpha)s_{t-3})& \\
&    = ... + (1-\alpha)^2 \alpha x_{t-2} + (1-\alpha)^3 s_{t-3}& \\
\label{equ:6}
&    = \alpha x_t +...+ (1-\alpha)^{t-1}\alpha x_1+(1-\alpha)^t x_0&
\end{aligned}
\end{equation}

To match the order of the data sequence $x = [ x_0, x_1, ... , x_{t-1}, x_t ]$, we rewrite the Equation (\ref{equ:6}) backwards:
\begin{equation}
\begin{aligned}
\label{equ:7}
&s_t = (1-\alpha)^t x_0 + (1-\alpha)^{t-1}\alpha x_1 +...& \\
&...+(1-\alpha)^2 \alpha x_{t-2} + (1-\alpha) \alpha x_{t-1} + \alpha x_t&
\end{aligned}
\end{equation}

Consequently, we store the weights $w$ geometric sequence of length $t$ with the first term $w_0=1$, and common ratio being equal to $(1-\alpha)$, written in reverse order:
\begin{equation}
\begin{aligned}
\label{equ:8}
&w = [ (1-\alpha)^t, (1-\alpha)^{t-1}, ... , (1-\alpha), 1 ]&
\end{aligned}
\end{equation}

All entries of the Equation (\ref{equ:8}), except the first one, are multiplied by $\alpha$.
The first entry does not have $\alpha$ weight by the definition of EMA smoothing since the first value $s_0$ is equal to the first data item $x_0$:
\begin{equation}
\begin{aligned}
\label{equ:9}
&\hat{w} = [ (1-\alpha)^t, (1-\alpha)^{t-1}\alpha, ... , (1-\alpha)\alpha, \alpha ]&
\end{aligned}
\end{equation}

Finally, we apply dot product of data slice $x = [ x_0, x_1, ... , x_{t-1}, x_t ]$ and $\hat{w}$ from the Equation (\ref{equ:9}):
\begin{equation}
\begin{aligned}
\label{equ:10}
&\hat{w} \cdot x = (1-\alpha)^t x_0 + (1-\alpha)^{t-1} \alpha x_1 +...& \\
&...+ (1-\alpha)^2 \alpha x_{t-2} + (1-\alpha) \alpha x_{t-1} + \alpha x_t&
\end{aligned}
\end{equation}



The resulting Equation (\ref{equ:10}) is equal to the Equation (\ref{equ:7}), which means that we optimized the EMA decomposition module to $O(1)$ time complexity.

\section{Ablation Study on EMA}
\label{app:ema-extended}
Initially, we conducted experiments to determine the suitable $\alpha$ parameter for each model.
We tested five variations of fixed $\alpha = \{ 0.1, 0.3, 0.5, 0.7, 0.9 \}$, where $\alpha = 0.1$ represents the heaviest smoothing and $\alpha = 0.9$ indicates slight smoothing.
Since our main goal is to make both trend and seasonality streams more interpretable, we assume that smaller $\alpha$ values are more appropriate for this task.
Larger values lead to the problem that the trend is not smooth enough, resulting in a complicated trend and easy seasonality.
For this experiment, we utilized large datasets: Weather, Traffic, and Electricity since they have longer time series data.

It is important to note that decomposition is used in xPatch, DLinear, and PatchTST differently in comparison to Autoformer and FEDformer.
In xPatch, DLinear, and PatchTST, data is decomposed into trend and seasonality, and both trend and seasonality are separately predicted by the model.
Since the decomposed data is the final shape that should be predicted, both trend and seasonality should be close in terms of complexity and interpretability.

On the other hand, Autoformer and FEDformer employ the inner series decomposition blocks.
The transformer encoder first eliminates the long-term trend part by series decomposition blocks and focuses on seasonal pattern modeling.
The decoder accumulates the trend part extracted from hidden variables progressively.
The past seasonal information from the encoder is utilized by the encoder-decoder Auto-Correlation.
In both Autoformer and FEDformer, the encoder contains two series decomposition blocks, while the decoder contains three series decomposition blocks, where they gradually smooth the data to focus on trend, rather than decompose the data into two streams.

\begin{figure}[ht]
\centering
\includegraphics[width=1\columnwidth]{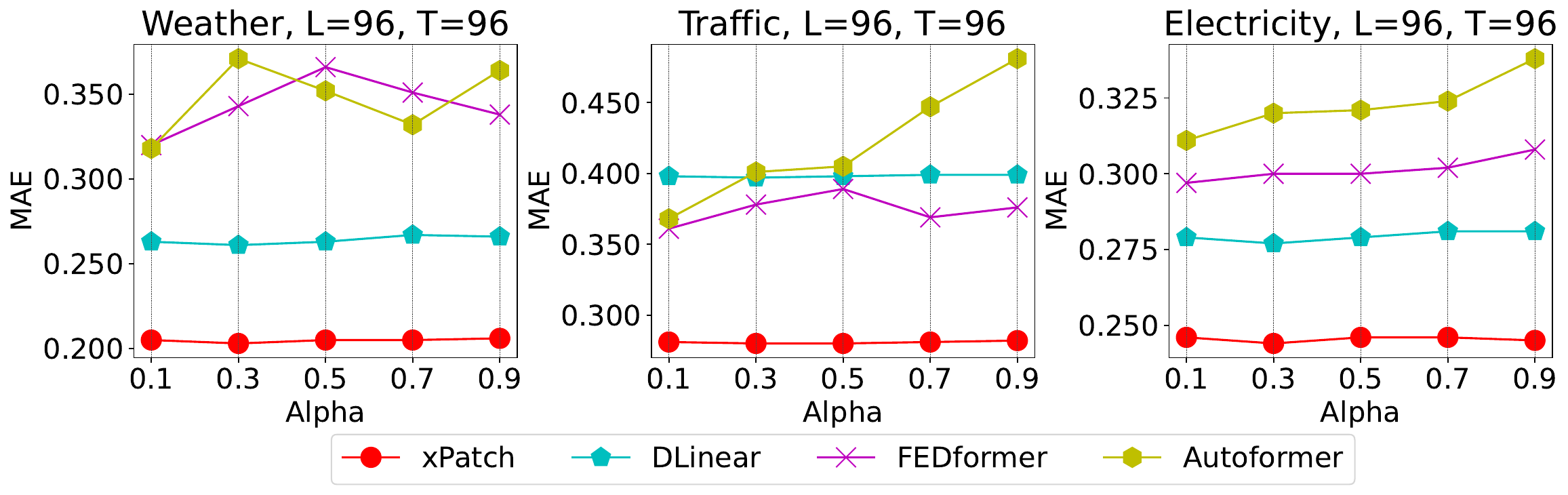}
\caption{Forecasting performance (MAE), lookback window $L = 96$, prediction horizon $T = 96$, alpha $\alpha = \{ 0.1, 0.3, 0.5, 0.7, 0.9 \}$.}
\label{fig:ablation-ema}
\end{figure}

\begin{table*}[ht]
  \scriptsize
  \centering
    \setlength\tabcolsep{3pt}
  \begin{tabular}{c|c|cccc|cccc|cccc|cccc|cccc}
    \hline
    \multicolumn{2}{c}{Method}      &
    \multicolumn{2}{|c}{xPatch}   & \multicolumn{2}{c}{xPatch*} & 
    \multicolumn{2}{|c}{PatchTST}   & \multicolumn{2}{c}{PatchTST*} & 
    \multicolumn{2}{|c}{DLinear}    & \multicolumn{2}{c}{DLinear*} & 
    \multicolumn{2}{|c}{FEDformer}  & \multicolumn{2}{c}{FEDformer*} & 
    \multicolumn{2}{|c}{Autoformer} & \multicolumn{2}{c}{Autoformer*} \\
    \hline
    \multicolumn{2}{c|}{Metric}
    & MSE & MAE   & MSE   & MAE & MSE & MAE   & MSE   & MAE & MSE & MAE & MSE & MAE & MSE & MAE & MSE & MAE & MSE & MAE & MSE & MAE \\
    \hline
    \multirow{4}{*}{\rotatebox{90}{ETTh1}}
    & 96  & 0.385 & 0.390 & \textbf{0.376} & \textbf{0.386} & 0.393 & 0.407 & \textbf{0.386} & \textbf{0.400} & 0.386 & 0.400 & \textbf{0.382} & \textbf{0.395} & \textbf{0.376} & 0.419 & 0.379 & \textbf{0.415} & \textbf{0.449} & \textbf{0.459} & 0.470 & 0.465 \\
    & 192 & 0.423 & 0.409 & \textbf{0.417} & \textbf{0.407} & 0.445 & 0.434 & \textbf{0.441} & \textbf{0.430} & \textbf{0.437} & \textbf{0.432} & 0.439 & 0.433 & \textbf{0.420} & 0.448 & 0.422 & \textbf{0.445} & 0.500 & 0.482 & \textbf{0.454} & \textbf{0.456} \\
    & 336 & 0.450 & 0.426 & \textbf{0.449} & \textbf{0.425} & 0.483 & 0.451 & \textbf{0.478} & \textbf{0.445} & \textbf{0.481} & \textbf{0.459} & 0.491 & 0.467 & 0.459 & \textbf{0.465} & \textbf{0.454} & 0.466 & 0.521 & 0.496 & \textbf{0.480} & \textbf{0.475} \\
    & 720 & 0.483 & 0.460 & \textbf{0.470} & \textbf{0.456} & 0.479 & 0.470 & \textbf{0.474} & \textbf{0.466} & \textbf{0.519} & 0.516 & 0.524 & \textbf{0.515} & 0.506 & 0.507 & \textbf{0.485} & \textbf{0.498} & 0.514 & 0.512 & \textbf{0.500} & \textbf{0.504} \\
    \hline
    \multirow{4}{*}{\rotatebox{90}{ETTh2}}
    & 96  & \textbf{0.232} & \textbf{0.300} & 0.233 & \textbf{0.300} & 0.293 & 0.342 & \textbf{0.291} & \textbf{0.340} & 0.333 & 0.387 & \textbf{0.329} & \textbf{0.384} & 0.346 & 0.388 & \textbf{0.335} & \textbf{0.382} & \textbf{0.358} & \textbf{0.397} & 0.371 & 0.407 \\
    & 192 & \textbf{0.289} & \textbf{0.338} & 0.291 & \textbf{0.338} & 0.377 & 0.393 & \textbf{0.371} & \textbf{0.390} & 0.477 & 0.476 & \textbf{0.431} & \textbf{0.443} & 0.429 & \textbf{0.439} & \textbf{0.426} & 0.443 & \textbf{0.456} & \textbf{0.452} & 0.457 & 0.457 \\
    & 336 & \textbf{0.339} & \textbf{0.376} & 0.344 & 0.377 & 0.380 & 0.408 & \textbf{0.375} & \textbf{0.407} & 0.594 & 0.541 & \textbf{0.445} & \textbf{0.454} & 0.496 & 0.487 & \textbf{0.470} & \textbf{0.472} & 0.482 & 0.486 & \textbf{0.467} & \textbf{0.474} \\
    & 720 & \textbf{0.406} & 0.430 & 0.407 & \textbf{0.427} & 0.411 & 0.433 & \textbf{0.408} & \textbf{0.432} & 0.831 & 0.657 & \textbf{0.776} & \textbf{0.632} & 0.463 & \textbf{0.474} & \textbf{0.460} & 0.475 & 0.515 & 0.511 & \textbf{0.454} & \textbf{0.477} \\
    \hline
    \multirow{4}{*}{\rotatebox{90}{ETTm1}}
    & 96  & 0.312 & 0.349 & \textbf{0.311} & \textbf{0.346} & \textbf{0.320} & 0.359 & \textbf{0.320} & \textbf{0.357} & \textbf{0.345} & \textbf{0.372} & 0.346 & \textbf{0.372} & 0.379 & 0.419 & \textbf{0.336} & \textbf{0.388} & 0.505 & 0.475 & \textbf{0.402} & \textbf{0.431} \\
    & 192 & 0.355 & 0.372 & \textbf{0.348} & \textbf{0.368} & 0.365 & \textbf{0.381} & \textbf{0.363} & \textbf{0.381} & \textbf{0.380} & 0.389 & 0.387 & \textbf{0.287} & 0.426 & 0.441 & \textbf{0.376} & \textbf{0.413} & \textbf{0.553} & \textbf{0.496} & 0.569 & 0.505 \\
    & 336 & 0.392 & 0.395 & \textbf{0.388} & \textbf{0.391} & \textbf{0.391} & \textbf{0.401} & \textbf{0.391} & 0.404 & 0.413 & \textbf{0.413} & \textbf{0.412} & 0.414 & 0.445 & 0.459 & \textbf{0.434} & \textbf{0.448} & 0.621 & 0.537 & \textbf{0.529} & \textbf{0.495} \\
    & 720 & 0.466 & 0.431 & \textbf{0.461} & \textbf{0.430} & 0.455 & \textbf{0.436} & \textbf{0.451} & 0.439 & \textbf{0.474} & \textbf{0.453} & 0.475 & \textbf{0.453} & 0.543 & 0.490 & \textbf{0.478} & \textbf{0.470} & 0.671 & 0.561 & \textbf{0.590} & \textbf{0.524} \\
    \hline
    \multirow{4}{*}{\rotatebox{90}{ETTm2}}
    & 96  & 0.165 & 0.250 & \textbf{0.164} & \textbf{0.248} & 0.177 & 0.259 & \textbf{0.176} & \textbf{0.258} & 0.193 & 0.292 & \textbf{0.186} & \textbf{0.277} & 0.203 & 0.287 & \textbf{0.181} & \textbf{0.275} & 0.255 & 0.339 & \textbf{0.210} & \textbf{0.297} \\
    & 192 & \textbf{0.230} & 0.293 & \textbf{0.230} & \textbf{0.291} & 0.248 & 0.306 & \textbf{0.240} & \textbf{0.303} & 0.284 & 0.362 & \textbf{0.256} & \textbf{0.328} & 0.269 & 0.328 & \textbf{0.248} & \textbf{0.319} & \textbf{0.281} & \textbf{0.340} & 0.283 & \textbf{0.340} \\
    & 336 & \textbf{0.292} & 0.333 & \textbf{0.292} & \textbf{0.331} & 0.313 & 0.346 & \textbf{0.300} & \textbf{0.341} & 0.369 & 0.427 & \textbf{0.324} & \textbf{0.374} & 0.325 & 0.366 & \textbf{0.314} & \textbf{0.362} & 0.339 & 0.372 & \textbf{0.330} & \textbf{0.368} \\
    & 720 & \textbf{0.381} & 0.384 & \textbf{0.381} & \textbf{0.383} & \textbf{0.399} & \textbf{0.397} & 0.403 & 0.400 & 0.554 & 0.522 & \textbf{0.511} & \textbf{0.498} & \textbf{0.421} & \textbf{0.415} & 0.427 & 0.421 & 0.433 & 0.432 & \textbf{0.431} & \textbf{0.427} \\
    \hline
    \multirow{4}{*}{\rotatebox{90}{Weather}}
    & 96  & 0.170 & 0.205 & \textbf{0.168} & \textbf{0.203} & 0.177 & 0.218 & \textbf{0.175} & \textbf{0.217} & \textbf{0.196} & \textbf{0.255} & 0.199 & 0.261 & \textbf{0.217} & \textbf{0.296} & 0.241 & 0.320 & 0.266 & 0.336 & \textbf{0.249} & \textbf{0.332} \\
    & 192 & 0.218 & 0.248 & \textbf{0.214} & \textbf{0.245} & 0.225 & 0.259 & \textbf{0.223} & \textbf{0.257} & \textbf{0.237} & 0.296 & \textbf{0.237} & \textbf{0.294} & 0.276 & \textbf{0.336} & \textbf{0.273} & 0.342 & \textbf{0.307} & \textbf{0.367} & 0.326 & 0.380 \\
    & 336 & 0.240 & 0.277 & \textbf{0.236} & \textbf{0.273} & 0.277 & 0.297 & \textbf{0.276} & \textbf{0.296} & 0.283 & 0.335 & \textbf{0.270} & \textbf{0.329} & \textbf{0.339} & \textbf{0.380} & 0.348 & 0.382 & 0.359 & 0.395 & \textbf{0.339} & \textbf{0.379} \\
    & 720 & 0.310 & 0.322 & \textbf{0.309} & \textbf{0.321} & 0.350 & \textbf{0.345} & \textbf{0.349} & 0.346 & 0.345 & \textbf{0.381} & \textbf{0.342} & \textbf{0.381} & 0.403 & 0.428 & \textbf{0.395} & \textbf{0.406} & \textbf{0.419} & \textbf{0.428} & 0.482 & 0.470 \\
    \hline
    \multirow{4}{*}{\rotatebox{90}{Traffic}}
    & 96  & 0.489 & 0.281 & \textbf{0.481} & \textbf{0.280} & 0.446 & 0.283 & \textbf{0.438} & \textbf{0.279} & \textbf{0.650} & \textbf{0.396} & 0.656 & 0.397 & 0.587 & 0.366 & \textbf{0.571} & \textbf{0.361} & 0.613 & 0.388 & \textbf{0.558} & \textbf{0.368} \\
    & 192 & 0.485 & 0.276 & \textbf{0.484} & \textbf{0.275} & 0.453 & 0.285 & \textbf{0.449} & \textbf{0.282} & 0.598 & 0.370 & \textbf{0.596} & \textbf{0.367} & \textbf{0.604} & \textbf{0.373} & 0.613 & 0.386 & \textbf{0.616} & \textbf{0.382} & 0.635 & 0.397 \\
    & 336 & \textbf{0.500} & 0.281 & 0.504 & \textbf{0.279} & 0.467 & 0.291 & \textbf{0.464} & \textbf{0.289} & 0.605 & 0.373 & \textbf{0.599} & \textbf{0.370} & 0.621 & 0.383 & \textbf{0.609} & \textbf{0.375} & \textbf{0.622} & \textbf{0.337} & 0.626 & 0.382 \\
    & 720 & \textbf{0.534} & 0.295 & 0.540 & \textbf{0.293} & 0.500 & 0.309 & \textbf{0.496} & \textbf{0.307} & 0.645 & 0.394 & \textbf{0.641} & \textbf{0.393} & \textbf{0.626} & \textbf{0.382} & 0.643 & 0.397 & 0.660 & 0.408 & \textbf{0.648} & \textbf{0.392} \\
    \hline
    \multirow{4}{*}{\rotatebox{90}{Electricity}}
    & 96  & 0.166 & 0.248 & \textbf{0.159} & \textbf{0.244} & 0.166 & 0.252 & \textbf{0.164} & \textbf{0.251} & 0.197 & 0.282 & \textbf{0.194} & \textbf{0.277} & 0.193 & 0.308 & \textbf{0.181} & \textbf{0.297} & 0.201 & 0.317 & \textbf{0.199} & \textbf{0.311} \\
    & 192 & 0.165 & 0.252 & \textbf{0.160} & \textbf{0.248} & 0.174 & 0.260 & \textbf{0.172} & \textbf{0.259} & 0.196 & 0.285 & \textbf{0.189} & \textbf{0.278} & 0.201 & 0.315 & \textbf{0.195} & \textbf{0.309} & 0.222 & 0.334 & \textbf{0.217} & \textbf{0.326} \\
    & 336 & 0.186 & 0.269 & \textbf{0.182} & \textbf{0.267} & 0.190 & 0.277 & \textbf{0.188} & \textbf{0.276} & 0.209 & 0.301 & \textbf{0.205} & \textbf{0.295} & 0.214 & 0.329 & \textbf{0.211} & \textbf{0.324} & \textbf{0.231} & \textbf{0.338} & 0.241 & 0.342 \\
    & 720 & 0.222 & 0.300 & \textbf{0.216} & \textbf{0.298} & \textbf{0.230} & \textbf{0.311} & \textbf{0.230} & 0.312 & 0.245 & 0.333 & \textbf{0.241} & \textbf{0.329} & \textbf{0.246} & \textbf{0.355} & 0.249 & 0.356 & 0.254 & 0.361 & \textbf{0.249} & \textbf{0.357} \\
    \hline
    \multirow{4}{*}{\rotatebox{90}{Exchange}}
    & 96  & 0.084 & 0.201 & \textbf{0.082} & \textbf{0.199} & 0.080 & \textbf{0.196} & \textbf{0.079} & \textbf{0.196} & 0.088 & 0.218 & \textbf{0.078} & \textbf{0.198} & 0.148 & 0.278 & \textbf{0.109} & \textbf{0.238} & 0.197 & 0.323 & \textbf{0.131} & \textbf{0.264} \\
    & 192 & 0.182 & 0.301 & \textbf{0.177} & \textbf{0.298} & 0.171 & 0.293 & \textbf{0.168} & \textbf{0.291} & 0.176 & 0.315 & \textbf{0.156} & \textbf{0.292} & 0.271 & 0.380 & \textbf{0.213} & \textbf{0.333} & 0.300 & 0.369 & \textbf{0.252} & \textbf{0.368} \\
    & 336 & \textbf{0.349} & \textbf{0.424} & \textbf{0.349} & 0.425 & \textbf{0.317} & \textbf{0.406} & 0.319 & 0.407 & 0.313 & 0.427 & \textbf{0.300} & \textbf{0.414} & 0.460 & 0.500 & \textbf{0.408} & \textbf{0.462} & 0.509 & 0.524 & \textbf{0.470} & \textbf{0.512} \\
    & 720 & 0.897 & 0.716 & \textbf{0.891} & \textbf{0.711} & 0.887 & 0.703 & \textbf{0.817} & \textbf{0.679} & 0.839 & 0.695 & \textbf{0.785} & \textbf{0.671} & 1.195 & 0.841 & \textbf{1.144} & \textbf{0.820} & 1.447 & 0.941 & \textbf{1.098} & \textbf{0.813} \\
    \hline
    \multirow{4}{*}{\rotatebox{90}{ILI}}
    & 24  & 1.541 & 0.755 & \textbf{1.378} & \textbf{0.685} & 1.691 & 0.816 & \textbf{1.468} & \textbf{0.729} & \textbf{2.398} & \textbf{1.040} & 2.592 & 1.092 & 3.228 & 1.260 & \textbf{2.646} & \textbf{1.062} & 3.483 & 1.287 & \textbf{3.403} & \textbf{1.254} \\
    & 36  & 1.468 & 0.734 & \textbf{1.315} & \textbf{0.681} & 1.415 & 0.762 & \textbf{1.343} & \textbf{0.695} & \textbf{2.646} & \textbf{1.088} & 2.738 & 1.125 & 2.679 & 1.080 & \textbf{2.492} & \textbf{0.971} & 3.103 & 1.148 & \textbf{2.720} & \textbf{1.051} \\
    & 48  & \textbf{1.439} & \textbf{0.743} & 1.459 & 0.747 & 1.754 & 0.819 & \textbf{1.617} & \textbf{0.809} & \textbf{2.614} & \textbf{1.086} & 2.665 & 1.098 & 2.622 & 1.078 & \textbf{2.521} & \textbf{1.017} & \textbf{2.669} & \textbf{1.085} & 2.737 & 1.098 \\
    & 60  & \textbf{1.574} & \textbf{0.778} & 1.616 & 0.787 & 1.645 & 0.820 & \textbf{1.569} & \textbf{0.764} & 2.804 & 1.146 & \textbf{2.787} & \textbf{1.136} & 2.857 & 1.157 & \textbf{2.716} & \textbf{1.097} & \textbf{2.770} & \textbf{1.125} & 2.889 & 1.139 \\
    \hline
    \multicolumn{2}{c|}{Gain} & & & 1.35\% & 1.01\% & & & 1.91\% & 1.25\% & & & 2.93\% & 3.00\% & & & 4.67\% & 2.90\% & & & 4.96\% & 2.30\% \\
    \hline
    \end{tabular}
  \caption{Comparison of forecasting errors between the baselines and the models with the EMA decomposition module with unified lookback window $L = 36$ for the ILI dataset, and $L = 96$ for all other datasets.
  The model name with * denotes the model with EMA decomposition.
    }
    \label{tab:ema-extended}
\end{table*}

From the results in Figure \ref{fig:ablation-ema}, for Autoformer and FEDformer we found optimal $\alpha = 0.1$, while for xPatch, DLinear, and PatchTST we select $\alpha = 0.3$.

To conclude, for the transformer-based models that employ Autoformer's series decomposition blocks, the best $\alpha = 0.1$, while for the models that initially decompose data into trend and seasonality patterns and predict them separately, the best $\alpha = 0.3$.
The $\alpha$ hyperparameter is consistent and robust to all datasets inside the mentioned two categories.
Additionally, setting $\alpha$ as a learnable parameter showed a marginal performance improvement with a significant training time increase.
Therefore, we decided to use a fixed $\alpha$ parameter.

Table \ref{tab:ema-extended} presents the full comparative analysis between the original state-of-the-art models and versions that incorporate the exponential decomposition module.
The effectiveness of EMA was assessed on Autoformer, FEDformer, and DLinear models that were initially designed with the decomposition unit, and additionally on PatchTST.
We reproduce the models according to their official code and configurations, only replacing the decomposition module with EMA.
All models are tested in unified experimental settings: lookback $L=36$ for the ILI dataset and $L=96$ for all other datasets.


\section{Dual Flow Net}
\label{app:dual-flow}
\begin{figure}[ht]
\centering
\includegraphics[width=1\columnwidth]{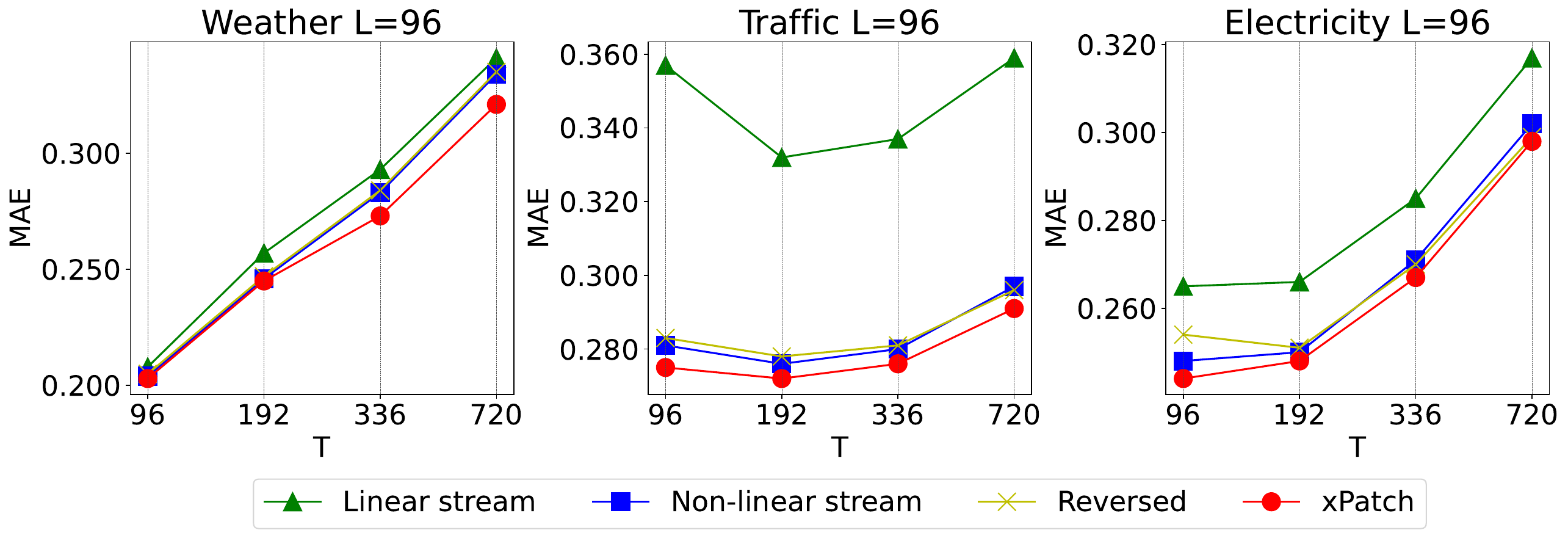}
\caption{Separate forecasting performance (MAE) of xPatch linear and non-linear streams, lookback window $L = 96$ and prediction horizon $T = \{ 96, 192, 336, 720 \}$.}
\label{fig:dual-flow}
\end{figure}

We explore the impact of the dual flow network in xPatch architecture and assess the contribution of each stream.
Figure \ref{fig:dual-flow} compares the dual flow network with separate non-linear and linear streams to analyze the contribution and behavior of each flow.
The four possible configurations:
\begin{itemize}
    \item Original: Seasonality $\rightarrow$ non-linear stream, Trend $\rightarrow$ linear stream
    \item Reversed: Seasonality $\rightarrow$ linear stream, Trend $\rightarrow$ non-linear stream
    \item Non-linear only: Seasonality $\rightarrow$ non-linear stream, Trend $\rightarrow$ non-linear stream
    \item Linear only: Seasonality $\rightarrow$ linear stream, Trend $\rightarrow$ linear stream
\end{itemize}

The results reveal that the model benefits from the original configuration of the dual flow architecture, effectively selecting between linear and non-linear features, leading to improved forecasting performance.

\section{Ablation Study on Arctangent Loss}
\label{app:arctan_loss}
We find that the scaling coefficient $\rho_{CARD}(i) = i^{-\frac{1}{2}}$ from Equation (\ref{equ:scalable_loss}) exhibits a too rapid decrease rate for the long-term time series forecasting task.
Therefore, we propose a novel approach by exploring the arctangent function, which features a slower increase rate compared to the exponential functions that were analyzed in the CARD \cite{wang2024card} paper.

Initially, we employ the negative arctangent function $-arctan(i)$ as a decreasing function required for our task.
Subsequently, we perform a vertical translation to ensure that the function equals 1 when $i=1$.
In other words, we shift the entire graph of the function along the y-axis by $y$, solving the equation $-\arctan(1) + y = 1$, which yields $y = \frac{\pi}{4} + 1$.

Therefore, the arctangent loss $\mathcal{L}_{arctan}$ between the predicted univariate sequence $\hat{x}_{1:T}^{(i)}$ and the ground truth observations $x_{1:T}^{(i)}$, where $T$ is future prediction length, $\rho_{arctan}(i)$ is loss scaling coefficient, and $m$ is arctangent scaling parameter is denoted as:
\begin{flalign}
    \mathcal{L}_{arctan} = \frac{1}{T} \sum _{i=1}^T \rho_{arctan}(i) || \hat{x}_{1:T}^{(i)} - x_{1:T}^{(i)} || \\
    \rho_{arctan}(i) = -m(\arctan(i)-\frac{\pi}{4})+1
\end{flalign}

Subsequently, we investigate the arctangent loss function scaling parameter.
In Table \ref{tab:arctan}, we compare different arctangent loss scaling parameters $m = \{ 1, \frac{1}{2}, \frac{1}{3}, \frac{1}{4} \}$, where $m=1$ is the original arctangent loss function and $m=\frac{1}{4}$ is the function closest to the original MAE loss function without scaling.

\begin{table}[th]
  \scriptsize
  \centering
    \setlength\tabcolsep{3pt}
  \begin{tabular}{c|c|cc|cc|cc|cc}
    \hline
    \multicolumn{2}{c}{Function} &
    \multicolumn{2}{|c}{m=1} & 
    \multicolumn{2}{|c}{m=0.5} & 
    \multicolumn{2}{|c}{m=0.33} & 
    \multicolumn{2}{|c}{m=0.25} \\
    \hline
    \multicolumn{2}{c|}{Metric}
    & MSE & MAE & MSE & MAE & MSE & MAE & MSE & MAE \\
    \hline
    \multirow{4}{*}{\rotatebox{90}{ETTh2}}
    & 96  & \textbf{0.226} & \textbf{0.297} & \textbf{0.226} & \textbf{0.297} & \textbf{0.226} & \textbf{0.297} & \textbf{0.226} & \textbf{0.297} \\
    & 192 & \textbf{0.275} & \textbf{0.330} & 0.276 & \textbf{0.330} & 0.276 & \textbf{0.330} & 0.276 & \textbf{0.330} \\
    & 336 & \textbf{0.312} & \textbf{0.360} & 0.313 & \textbf{0.360} & 0.313 & \textbf{0.360} & 0.313 & \textbf{0.360} \\
    & 720 & \textbf{0.384} & \textbf{0.418} & \textbf{0.384} & \textbf{0.418} & \textbf{0.384} & \textbf{0.418} & \textbf{0.384} & \textbf{0.418} \\
    \hline
    \multirow{4}{*}{\rotatebox{90}{ETTm2}}
    & 96  & \textbf{0.153} & \textbf{0.240} & 0.154 & \textbf{0.240} & 0.154 & \textbf{0.240} & 0.154 & \textbf{0.240} \\
    & 192 & \textbf{0.213} & \textbf{0.280} & \textbf{0.213} & \textbf{0.280} & \textbf{0.213} & \textbf{0.280} & \textbf{0.213} & \textbf{0.280} \\
    & 336 & \textbf{0.264} & \textbf{0.315} & \textbf{0.264} & \textbf{0.315} & \textbf{0.264} & \textbf{0.315} & \textbf{0.264} & \textbf{0.315} \\
    & 720 & \textbf{0.338} & \textbf{0.363} & \textbf{0.338} & \textbf{0.363} & \textbf{0.338} & 0.364 & \textbf{0.338} & 0.364 \\
    \hline
    \multirow{4}{*}{\rotatebox{90}{Weather}}
    & 96  & \textbf{0.146} & \textbf{0.185} & \textbf{0.146} & \textbf{0.185} & \textbf{0.146} & \textbf{0.185} & \textbf{0.146} & \textbf{0.185} \\
    & 192 & \textbf{0.189} & \textbf{0.227} & \textbf{0.189} & \textbf{0.227} & \textbf{0.189} & \textbf{0.227} & \textbf{0.189} & \textbf{0.227} \\
    & 336 & \textbf{0.218} & \textbf{0.260} & \textbf{0.218} & \textbf{0.260} & \textbf{0.218} & \textbf{0.260} & \textbf{0.218} & \textbf{0.260} \\
    & 720 & \textbf{0.291} & \textbf{0.315} & \textbf{0.291} & \textbf{0.315} & \textbf{0.291} & \textbf{0.315} & \textbf{0.291} & \textbf{0.315} \\
    \hline
    \end{tabular}
  \caption{Comparison of forecasting errors between xPatch versions with different arctangent loss scaling coefficient $m = \{ 1, \frac{1}{2}, \frac{1}{3}, \frac{1}{4} \}$.
    }
    \label{tab:arctan}
\end{table}

The experiment results indicate that there is no need to scale the arctangent loss, therefore we maintain $m=1$ without scaling.
The arctangent scaling coefficient is denoted as:
\begin{flalign}
\rho_{arctan}(i) = -\arctan(i)+\frac{\pi}{4}+1
\end{flalign}

\begin{figure}[ht]
\centering
\includegraphics[width=1\columnwidth]{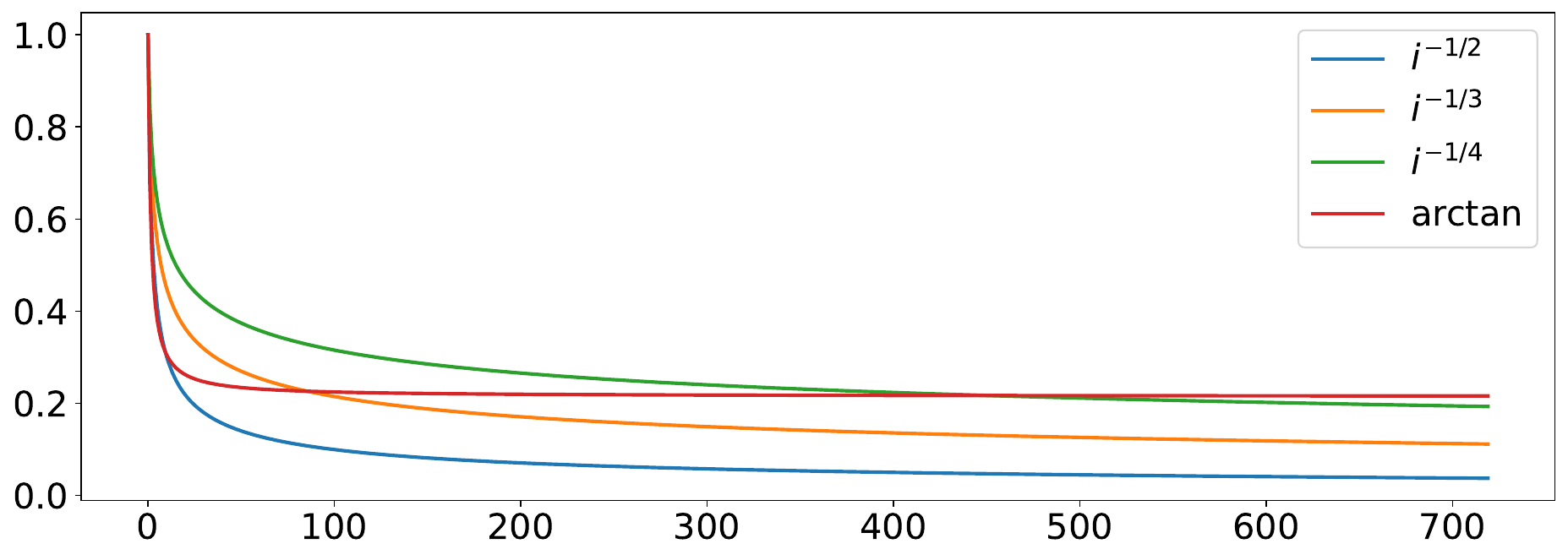}
\caption{Comparison between the arctangent function and exponential functions.}
\label{fig:loss}
\end{figure}

Figure \ref{fig:loss} illustrates the comparison between arctangent function and exponential functions $i^{- \frac{1}{2}}, i^{- \frac{1}{3}}$, and $i^{- \frac{1}{4}}$ analyzed in CARD \cite{wang2024card}.

\begin{table*}[ht]
  \scriptsize
  \centering
    \setlength\tabcolsep{1pt}
  \begin{tabular}{c|c|cccc|cccc|cccc|cccc|cccc|cccc}
    \hline
    \multicolumn{2}{c}{Method}      &
    \multicolumn{2}{|c}{xPatch}    & \multicolumn{2}{c}{xPatch*} & 
    \multicolumn{2}{|c}{CARD}  & \multicolumn{2}{c}{CARD*} & 
    \multicolumn{2}{|c}{PatchTST} & \multicolumn{2}{c}{PatchTST*} &
    \multicolumn{2}{|c}{DLinear}    & \multicolumn{2}{c}{DLinear*} & 
    \multicolumn{2}{|c}{FEDformer}  & \multicolumn{2}{c}{FEDformer*} & 
    \multicolumn{2}{|c}{Autoformer} & \multicolumn{2}{c}{Autoformer*} \\
    \hline
    \multicolumn{2}{c|}{Metric}
    & MSE & MAE & MSE & MAE & MSE & MAE & MSE & MAE & MSE & MAE & MSE & MAE & MSE & MAE & MSE & MAE & MSE & MAE & MSE & MAE & MSE & MAE & MSE & MAE \\
    \hline
    \multirow{4}{*}{\rotatebox{90}{ETTh1}}
    & 96  & \textbf{0.375} & 0.394 & 0.376 & \textbf{0.386} & 0.383 & \textbf{0.391} & \textbf{0.382} & 0.393 & 0.393 & 0.407 & \textbf{0.385} & \textbf{0.398} & 0.386 & 0.400 & \textbf{0.381} & \textbf{0.387} & 0.376 & 0.419 & \textbf{0.370} & \textbf{0.400} & \textbf{0.449} & 0.459 & 0.451 & \textbf{0.440} \\
    & 192 & \textbf{0.413} & 0.412 & 0.417 & \textbf{0.407} & \textbf{0.435} & \textbf{0.420} & \textbf{0.435} & 0.421 & 0.445 & 0.434 & \textbf{0.443} & \textbf{0.427} & 0.437 & 0.432 & \textbf{0.430} & \textbf{0.417} & \textbf{0.420} & 0.448 & \textbf{0.420} & \textbf{0.432} & 0.500 & 0.482 & \textbf{0.458} & \textbf{0.444} \\
    & 336 & \textbf{0.438} & 0.430 & 0.449 & \textbf{0.425} & 0.479 & 0.442 & \textbf{0.474} & \textbf{0.439} & \textbf{0.483} & 0.451 & \textbf{0.483} & \textbf{0.444} & 0.481 & 0.459 & \textbf{0.478} & \textbf{0.449} & \textbf{0.459} & 0.465 & 0.462 & \textbf{0.455} & \textbf{0.521} & 0.496 & 0.537 & \textbf{0.487} \\
    & 720 & \textbf{0.466} & 0.464 & 0.470 & \textbf{0.456} & 0.471 & 0.461 & \textbf{0.462} & \textbf{0.455} & 0.479 & 0.470 & \textbf{0.474} & \textbf{0.460} & 0.519 & 0.516 & \textbf{0.508} & \textbf{0.499} & 0.506 & 0.507 & \textbf{0.498} & \textbf{0.495} & \textbf{0.514} & \textbf{0.512} & 0.571 & 0.532 \\
    \hline
    \multirow{4}{*}{\rotatebox{90}{ETTh2}}
    & 96  & 0.236 & 0.311 & \textbf{0.233} & \textbf{0.300} & \textbf{0.281} & \textbf{0.330} & 0.283 & 0.332 & 0.293 & 0.342 & \textbf{0.292} & \textbf{0.336} & 0.333 & 0.387 & \textbf{0.294} & \textbf{0.344} & 0.346 & 0.388 & \textbf{0.328} & \textbf{0.370} & 0.358 & 0.397 & \textbf{0.337} & \textbf{0.375} \\
    & 192 & 0.292 & 0.346 & \textbf{0.291} & \textbf{0.338} & \textbf{0.363} & \textbf{0.381} & \textbf{0.363} & 0.383 & 0.377 & 0.393 & \textbf{0.371} & \textbf{0.385} & 0.477 & 0.476 & \textbf{0.379} & \textbf{0.396} & 0.429 & 0.439 & \textbf{0.414} & \textbf{0.423} & 0.456 & 0.452 & \textbf{0.422} & \textbf{0.427} \\
    & 336 & \textbf{0.344} & 0.386 & \textbf{0.344} & \textbf{0.377} & 0.411 & 0.418 & \textbf{0.378} & \textbf{0.401} & 0.380 & 0.408 & \textbf{0.378} & \textbf{0.402} & 0.594 & 0.541 & \textbf{0.436} & \textbf{0.442} & 0.496 & 0.487 & \textbf{0.462} & \textbf{0.463} & 0.482 & 0.486 & \textbf{0.439} & \textbf{0.449} \\
    & 720 & 0.414 & 0.440 & \textbf{0.407} & \textbf{0.427} & 0.416 & 0.431 & \textbf{0.399} & \textbf{0.421} & 0.411 & 0.433 & \textbf{0.406} & \textbf{0.427} & 0.831 & 0.657 & \textbf{0.584} & \textbf{0.534} & 0.463 & 0.474 & \textbf{0.433} & \textbf{0.455} & 0.515 & 0.511 & \textbf{0.427} & \textbf{0.448} \\
    \hline
    \multirow{4}{*}{\rotatebox{90}{ETTm1}}
    & 96  & 0.329 & 0.373 & \textbf{0.311} & \textbf{0.346} & \textbf{0.316} & \textbf{0.347} & \textbf{0.316} & \textbf{0.347} & 0.320 & 0.359 & \textbf{0.310} & \textbf{0.338} & 0.345 & 0.372 & \textbf{0.331} & \textbf{0.350} & 0.379 & 0.419 & \textbf{0.345} & \textbf{0.382} & 0.505 & 0.475 & \textbf{0.407} & \textbf{0.422} \\
    & 192 & 0.351 & 0.386 & \textbf{0.348} & \textbf{0.368} & \textbf{0.363} & 0.370 & \textbf{0.363} & \textbf{0.368} & \textbf{0.365} & 0.381 & 0.367 & \textbf{0.367} & 0.380 & 0.389 & \textbf{0.376} & \textbf{0.373} & 0.426 & 0.441 & \textbf{0.390} & \textbf{0.407} & \textbf{0.553} & 0.496 & 0.555 & \textbf{0.483} \\
    & 336 & 0.390 & 0.409 & \textbf{0.388} & \textbf{0.391} & \textbf{0.392} & \textbf{0.390} & 0.394 & 0.391 & 0.391 & 0.401 & \textbf{0.390} & \textbf{0.388} & 0.413 & 0.413 & \textbf{0.407} & \textbf{0.395} & 0.445 & 0.459 & \textbf{0.431} & \textbf{0.430} & 0.621 & 0.537 & \textbf{0.487} & \textbf{0.458} \\
    & 720 & \textbf{0.458} & 0.445 & 0.461 & \textbf{0.430} & \textbf{0.458} & \textbf{0.425} & 0.462 & 0.428 & \textbf{0.455} & 0.436 & 0.459 & \textbf{0.428} & 0.474 & 0.453 & \textbf{0.469} & \textbf{0.433} & 0.543 & 0.490 & \textbf{0.482} & \textbf{0.465} & 0.671 & 0.561 & \textbf{0.488} & \textbf{0.462} \\
    \hline
    \multirow{4}{*}{\rotatebox{90}{ETTm2}}
    & 96  & 0.167 & 0.258 & \textbf{0.164} & \textbf{0.248} & \textbf{0.169} & \textbf{0.248} & \textbf{0.169} & \textbf{0.248} & 0.177 & 0.259 & \textbf{0.175} & \textbf{0.252} & 0.193 & 0.292 & \textbf{0.182} & \textbf{0.257} & 0.203 & 0.287 & \textbf{0.185} & \textbf{0.272} & 0.255 & 0.339 & \textbf{0.218} & \textbf{0.302} \\
    & 192 & 0.232 & 0.301 & \textbf{0.230} & \textbf{0.291} & \textbf{0.234} & \textbf{0.292} & 0.236 & 0.293 & 0.248 & 0.306 & \textbf{0.244} & \textbf{0.297} & 0.284 & 0.362 & \textbf{0.244} & \textbf{0.302} & 0.269 & 0.328 & \textbf{0.252} & \textbf{0.314} & 0.281 & 0.340 & \textbf{0.270} & \textbf{0.328} \\
    & 336 & \textbf{0.291} & 0.338 & 0.292 & \textbf{0.331} & 0.294 & 0.339 & \textbf{0.293} & \textbf{0.329} & 0.313 & 0.346 & \textbf{0.305} & \textbf{0.337} & 0.369 & 0.427 & \textbf{0.306} & \textbf{0.346} & 0.325 & 0.366 & \textbf{0.317} & \textbf{0.356} & 0.339 & 0.372 & \textbf{0.322} & \textbf{0.361} \\
    & 720 & \textbf{0.378} & 0.391 & 0.381 & \textbf{0.383} & \textbf{0.390} & \textbf{0.388} & 0.392 & \textbf{0.388} & \textbf{0.399} & 0.397 & \textbf{0.399} & \textbf{0.391} & 0.554 & 0.522 & \textbf{0.415} & \textbf{0.413} & 0.421 & 0.415 & \textbf{0.415} & \textbf{0.408} & 0.433 & 0.432 & \textbf{0.414} & \textbf{0.411} \\
    \hline
    \multirow{4}{*}{\rotatebox{90}{Weather}}
    & 96  & 0.173 & 0.218 & \textbf{0.170} & \textbf{0.212} & \textbf{0.150} & \textbf{0.188} & 0.154 & 0.193 & 0.177 & 0.218 & \textbf{0.176} & \textbf{0.208} & \textbf{0.196} & 0.255 & 0.207 & \textbf{0.233} & \textbf{0.217} & \textbf{0.296} & 0.228 & 0.298 & 0.266 & 0.336 & \textbf{0.230} & \textbf{0.288} \\
    & 192 & 0.217 & 0.256 & \textbf{0.210} & \textbf{0.248} & \textbf{0.202} & \textbf{0.238} & 0.205 & 0.240 & 0.225 & 0.259 & \textbf{0.222} & \textbf{0.249} & \textbf{0.237} & 0.296 & 0.243 & \textbf{0.268} & 0.276 & 0.336 & \textbf{0.266} & \textbf{0.322} & 0.307 & 0.367 & \textbf{0.296} & \textbf{0.343} \\
    & 336 & 0.238 & 0.283 & \textbf{0.226} & \textbf{0.273} & \textbf{0.260} & \textbf{0.282} & 0.264 & 0.286 & 0.277 & 0.297 & \textbf{0.276} & \textbf{0.289} & \textbf{0.283} & 0.335 & 0.286 & \textbf{0.305} & \textbf{0.339} & \textbf{0.380} & 0.349 & 0.387 & 0.359 & 0.395 & \textbf{0.347} & \textbf{0.372} \\
    & 720 & 0.310 & 0.329 & \textbf{0.309} & \textbf{0.321} & 0.343 & 0.353 & \textbf{0.342} & \textbf{0.337} & \textbf{0.350} & 0.345 & \textbf{0.350} & \textbf{0.339} & \textbf{0.345} & 0.381 & \textbf{0.345} & \textbf{0.355} & 0.403 & 0.428 & \textbf{0.383} & \textbf{0.412} & \textbf{0.419} & 0.428 & 0.421 & \textbf{0.423} \\
    \hline
    \multirow{4}{*}{\rotatebox{90}{Traffic}}
    & 96  & 0.490 & 0.299 & \textbf{0.471} & \textbf{0.275} & 0.419 & 0.269 & \textbf{0.395} & \textbf{0.250} & \textbf{0.446} & 0.283 & 0.472 & \textbf{0.270} & \textbf{0.650} & 0.396 & 0.687 & \textbf{0.364} & \textbf{0.587} & 0.366 & 0.596 & \textbf{0.347} & \textbf{0.613} & 0.388 & 0.647 & \textbf{0.374} \\
    & 192 & 0.488 & 0.293 & \textbf{0.478} & \textbf{0.272} & 0.443 & 0.276 & \textbf{0.418} & \textbf{0.258} & \textbf{0.453} & 0.285 & 0.476 & \textbf{0.274} & \textbf{0.598} & 0.370 & 0.645 & \textbf{0.340} & \textbf{0.604} & 0.373 & 0.617 & \textbf{0.360} & \textbf{0.616} & 0.382 & 0.656 & \textbf{0.381} \\
    & 336 & \textbf{0.495} & 0.298 & 0.501 & \textbf{0.276} & 0.460 & 0.283 & \textbf{0.437} & \textbf{0.265} & \textbf{0.467} & 0.291 & 0.493 & \textbf{0.280} & \textbf{0.605} & 0.373 & 0.647 & \textbf{0.342} & \textbf{0.621} & 0.383 & 0.627 & \textbf{0.358} & \textbf{0.622} & \textbf{0.337} & 0.638 & 0.366 \\
    & 720 & 0.548 & 0.335 & \textbf{0.547} & \textbf{0.291} & 0.490 & 0.299 & \textbf{0.485} & \textbf{0.291} & \textbf{0.500} & 0.309 & 0.526 & \textbf{0.299} & \textbf{0.645} & 0.394 & 0.676 & \textbf{0.362} & \textbf{0.626} & 0.382 & 0.653 & \textbf{0.368} & \textbf{0.660} & 0.408 & 0.666 & \textbf{0.380} \\
    \hline
    \multirow{4}{*}{\rotatebox{90}{Electricity}}
    & 96  & \textbf{0.159} & 0.251 & \textbf{0.159} & \textbf{0.244} & \textbf{0.141} & \textbf{0.233} & 0.144 & 0.235 & \textbf{0.166} & 0.252 & 0.172 & \textbf{0.248} & \textbf{0.197} & 0.282 & 0.198 & \textbf{0.269} & 0.193 & 0.308 & \textbf{0.187} & \textbf{0.293} & 0.201 & 0.317 & \textbf{0.196} & \textbf{0.304} \\
    & 192 & \textbf{0.160} & 0.255 & \textbf{0.160} & \textbf{0.248} & 0.160 & 0.250 & \textbf{0.158} & \textbf{0.247} & \textbf{0.174} & 0.260 & 0.179 & \textbf{0.256} & \textbf{0.196} & 0.285 & \textbf{0.196} & \textbf{0.272} & 0.201 & 0.315 & \textbf{0.197} & \textbf{0.304} & 0.222 & 0.334 & \textbf{0.217} & \textbf{0.322} \\
    & 336 & 0.183 & 0.275 & \textbf{0.182} & \textbf{0.267} & \textbf{0.173} & \textbf{0.263} & \textbf{0.173} & 0.265 & \textbf{0.190} & 0.277 & 0.194 & \textbf{0.272} & \textbf{0.209} & 0.301 & \textbf{0.209} & \textbf{0.287} & 0.214 & 0.329 & \textbf{0.209} & \textbf{0.315} & \textbf{0.231} & \textbf{0.338} & 0.243 & 0.341 \\
    & 720 & 0.226 & 0.313 & \textbf{0.216} & \textbf{0.298} & \textbf{0.197} & \textbf{0.284} & 0.201 & 0.286 & \textbf{0.230} & 0.311 & 0.234 & \textbf{0.306} & 0.245 & 0.333 & \textbf{0.244} & \textbf{0.319} & 0.246 & 0.355 & \textbf{0.244} & \textbf{0.346} & \textbf{0.254} & 0.361 & 0.258 & \textbf{0.351} \\
    \hline
    \multicolumn{2}{c|}{Gain} & & & 0.95\% & 3.89\% & & & 0.76\% & 1.06\% & & & 0.55\% & 2.62\% & & & 4.61\% & 8.89\% & & & 2.69\% & 3.97\% & & & 4.69\% & 5.31\% \\
    \hline
    \end{tabular}
  \caption{Comparison of forecasting errors between the baselines and the models trained with arctangent loss with unified lookback window $L = 96$.
  The model name with * denotes the model trained with arctangent loss.}
    \label{tab:loss-extended}
\end{table*}

Considering that the furthest predicting horizon under the common LTSF setting is $i=720$, the scaling coefficient of signal decay-based loss $\rho_{\text{CARD}}(720) \approx 0.037$, whereas the scaling coefficient of the arctangent loss $\rho_{\text{arctan}}(720) \approx 0.216$.
Figure \ref{fig:loss} highlights that the arctangent function exhibits a steeper initial curve compared to exponential functions, indicating an overall slower decay rate.

The effectiveness of the arctangent loss function was evaluated on PatchTST, DLinear, FEDformer, and Autoformer models that were trained with MSE loss objective, and additionally on the CARD model comparing arctangent loss with the signal decay-based loss.
Table \ref{tab:loss-extended} presents the full comparative analysis between the original state-of-the-art models and versions that are trained using the proposed arctangent training loss.
All models are tested with lookback $L=96$ for all datasets.

\section{Ablation Study on Sigmoid Learning Rate Adjustment Scheme}
\label{app:sigmoid}
In Equation \ref{equ:cosine}, CARD employs linear warm-up initially and then adjusts the learning rate according to one cycle of the cosine function for the remaining epochs.
While we recognize the effectiveness of the initial warm-up approach, we propose that the entire adjustment scheme should be encapsulated within a single function featuring a non-linear warm-up.
Therefore, we chose to investigate the sigmoid function.

A logistic function is a common sigmoid curve with the following equation:
\begin{flalign}
    f(x) = \frac{L}{1+e^{-k(x-x_0)}}
\end{flalign}
where $x_0$ is the $x$ value of the function's midpoint, $L$ is the supremum of the values of the function, and k is the logistic growth rate or steepness of the curve.

To calculate the learning rate $\alpha$ for the current epoch $t$, the supremum $L$ corresponds to the initialized learning rate $\alpha_0$, while the midpoint $t_0$ serves as a warm-up coefficient denoted by $w$, as it extends the rising curve of the sigmoid function:
\begin{flalign}
\label{equ:sig1}
    \alpha_t = \frac{\alpha_0}{1+e^{-k(t-w)}}
\end{flalign}
To extend the function to slowly decrease after some point, Equation (\ref{equ:sig1}) requires updating with the subtraction of another sigmoid function featuring a smaller growth rate.
Before delving into the analysis of the second sigmoid function, we examine the variable $e$:
\begin{flalign}
    e^{-k(t-w)} = e^{-kt+kw}
\end{flalign}
Since the $\alpha_t$ function at $t=0$ is intended to be equal to 0, we can revise the $-kt$ term to $-\frac{kt}{s}$, where $s$ represents the smoothing coefficient:
\begin{flalign}
    e^{-\frac{kt}{s}+kw} = e^{-\frac{k}{s}(t-sw)}
\end{flalign}
Hence, we introduce the novel sigmoid learning rate adjustment scheme.
The learning rate $\alpha_t$ at epoch $t$, with an initial learning rate $\alpha_0$, logistic growth rate $k$, decreasing curve smoothing rate $s$, and warm-up coefficient $w$, is calculated as follows:
\begin{flalign}
    \label{equ:sigmoid2}
    \alpha_t = \frac{\alpha_0}{1+e^{-k(t-w)}} - \frac{\alpha_0}{1+e^{-\frac{k}{s}(t-sw)}}
\end{flalign}
Note that the rationale behind incorporating this specific smoothing coefficient $s$ is to ensure that $\alpha_t(0) = 0$, as the coefficient diminishes with $t=0$.

Equation (\ref{equ:sigmoid2}) contains three hyperparameters: the logistic growth rate $k$, the decreasing curve smoothing rate $s$, and the warm-up coefficient $w$.
In Table \ref{tab:sigmoid}, we compare different sigmoid learning rate adjustment technique hyperparameters $k = \{ 0.3, 0.5, 0.8 \}$, $s = \{ 5, 8, 10 \}$, $w = \{ 5, 8, 10 \}$.

\begin{table}[th]
  \scriptsize
  \centering
    \setlength\tabcolsep{1pt}
  \begin{tabular}{c|c|cc|cc|cc|cc|cc|cc}
    \hline
    \multicolumn{2}{c}{Function} &
    \multicolumn{2}{|c}{0.3,5,5} & 
    \multicolumn{2}{|c}{0.3,10,10} & 
    \multicolumn{2}{|c}{0.5,5,5} & 
    \multicolumn{2}{|c}{0.5,10,10} & 
    \multicolumn{2}{|c}{0.8,8,8} & 
    \multicolumn{2}{|c}{0.8,5,10} \\
    \hline
    \multicolumn{2}{c|}{Metric}
    & MSE & MAE & MSE & MAE & MSE & MAE & MSE & MAE & MSE & MAE & MSE & MAE \\
    \hline
    \multirow{4}{*}{\rotatebox{90}{ETTh1}}
    & 96  & 0.359 & 0.389 & 0.355 & \textbf{0.379} & 0.362 & 0.392 & \textbf{0.354} & \textbf{0.379} & \textbf{0.354} & \textbf{0.379} & 0.355 & 0.381\\
    & 192 & 0.378 & 0.401 & \textbf{0.376} & 0.396 & 0.379 & 0.403 & \textbf{0.376} & \textbf{0.395} & 0.377 & 0.396 & 0.377 & 0.396\\
    & 336 & 0.394 & 0.419 & 0.394 & 0.417 & 0.392 & 0.416 & \textbf{0.391} & \textbf{0.415} & 0.392 & 0.416 & 0.392 & 0.416\\
    & 720 & 0.443 & 0.463 & \textbf{0.442} & \textbf{0.459} & 0.444 & 0.463 & \textbf{0.442} & \textbf{0.459} & 0.444 & 0.460 & 0.443 & \textbf{0.459}\\
    \hline
    \multirow{4}{*}{\rotatebox{90}{ETTh2}}
    & 96  & 0.228 & 0.299 & \textbf{0.226} & \textbf{0.296} & 0.229 & 0.300 & \textbf{0.226} & 0.297 & \textbf{0.226} & 0.297 & \textbf{0.226} & 0.297\\
    & 192 & 0.278 & 0.333 & \textbf{0.275} & \textbf{0.330} & 0.278 & 0.333 & \textbf{0.275} & \textbf{0.330} & \textbf{0.275} & 0.331 & \textbf{0.275} & \textbf{0.330}\\
    & 336 & 0.315 & 0.363 & 0.315 & 0.361 & 0.315 & 0.363 & \textbf{0.312} & \textbf{0.360} & 0.315 & 0.361 & \textbf{0.312} & \textbf{0.360}\\
    & 720 & 0.387 & 0.420 & \textbf{0.384} & \textbf{0.418} & 0.388 & 0.421 & \textbf{0.384} & \textbf{0.418} & \textbf{0.384} & \textbf{0.418} & 0.385 & 0.419\\
    \hline
    \multirow{4}{*}{\rotatebox{90}{ETTm1}}
    & 96  & 0.276 & \textbf{0.330} & \textbf{0.275} & \textbf{0.330} & 0.276 & \textbf{0.330} & \textbf{0.275} & \textbf{0.330} & \textbf{0.275} & \textbf{0.330} & \textbf{0.275} & \textbf{0.330}\\
    & 192 & 0.317 & 0.356 & \textbf{0.315} & 0.356 & 0.317 & 0.357 & \textbf{0.315} & \textbf{0.355} & 0.316 & 0.356 & 0.316 & 0.356\\
    & 336 & 0.356 & 0.377 & \textbf{0.355} & \textbf{0.376} & 0.356 & 0.377 & \textbf{0.355} & \textbf{0.376} & \textbf{0.355} & \textbf{0.376} & \textbf{0.355} & \textbf{0.376}\\
    & 720 & 0.420 & 0.412 & \textbf{0.419} & 0.412 & 0.421 & 0.412 & \textbf{0.419} & \textbf{0.411} & \textbf{0.419} & \textbf{0.411} & 0.422 & 0.414\\
    \hline
    \multirow{4}{*}{\rotatebox{90}{ETTm2}}
    & 96  & 0.154 & \textbf{0.240} & \textbf{0.153} & \textbf{0.240} & 0.154 & \textbf{0.240} & \textbf{0.153} & \textbf{0.240} & \textbf{0.153} & \textbf{0.240} & \textbf{0.153} & \textbf{0.240}\\
    & 192 & 0.214 & 0.281 & 0.215 & 0.281 & 0.215 & 0.281 & \textbf{0.213} & \textbf{0.280} & 0.214 & 0.281 & 0.214 & 0.282\\
    & 336 & 0.265 & \textbf{0.315} & 0.265 & \textbf{0.315} & 0.265 & \textbf{0.315} & \textbf{0.264} & \textbf{0.315} & 0.265 & \textbf{0.315} & 0.265 & 0.316\\
    & 720 & 0.340 & 0.365 & \textbf{0.338} & \textbf{0.363} & 0.341 & 0.365 & \textbf{0.338} & \textbf{0.363} & \textbf{0.338} & \textbf{0.363} & 0.339 & \textbf{0.363}\\
    \hline
    \multirow{4}{*}{\rotatebox{90}{Weather}}
    & 96  & 0.147 & 0.190 & 0.148 & \textbf{0.185} & 0.147 & 0.190 & \textbf{0.146} & \textbf{0.185} & \textbf{0.146} & \textbf{0.185} & \textbf{0.146} & 0.186\\
    & 192 & 0.191 & 0.228 & \textbf{0.189} & \textbf{0.227} & 0.191 & 0.228 & \textbf{0.189} & \textbf{0.227} & \textbf{0.189} & \textbf{0.227} & \textbf{0.189} & \textbf{0.227}\\
    & 336 & 0.220 & 0.261 & 0.219 & \textbf{0.260} & 0.221 & 0.261 & \textbf{0.218} & \textbf{0.260} & 0.219 & \textbf{0.260} & 0.219 & \textbf{0.260}\\
    & 720 & 0.293 & 0.317 & 0.292 & \textbf{0.315} & 0.294 & 0.318 & \textbf{0.291} & \textbf{0.315} & 0.292 & \textbf{0.315} & 0.292 & 0.316\\
    \hline
    \end{tabular}
  \caption{Comparison of forecasting errors between xPatch versions trained with different sigmoid learning rate hyperparameters (k,s,w).
    }
    \label{tab:sigmoid}
\end{table}

\begin{figure}[ht]
\centering
\includegraphics[width=1\columnwidth]{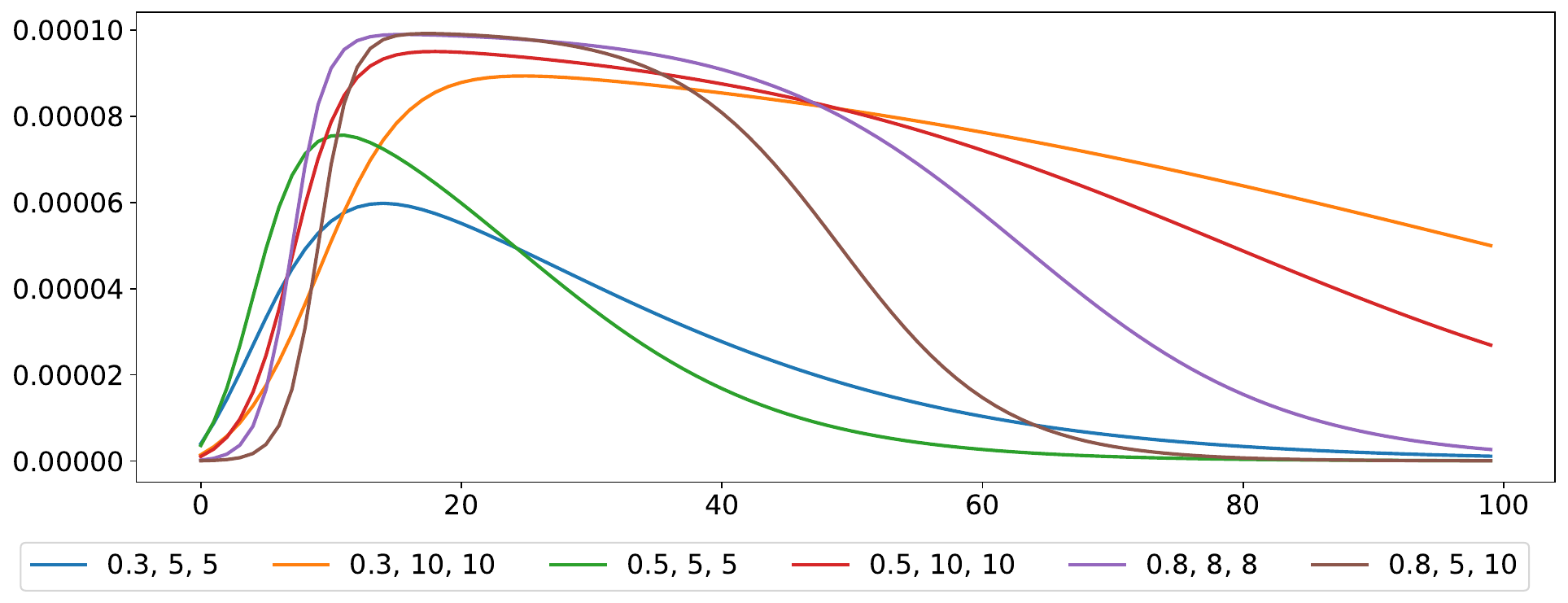}
\caption{Sigmoid function with different (k, s, w) parameters.}
\label{fig:lr_abl}
\end{figure}

Figure \ref{fig:lr_abl} demonstrates the Sigmoid function with different hyperparameters.

\begin{figure}[ht]
\centering
\includegraphics[width=1\columnwidth]{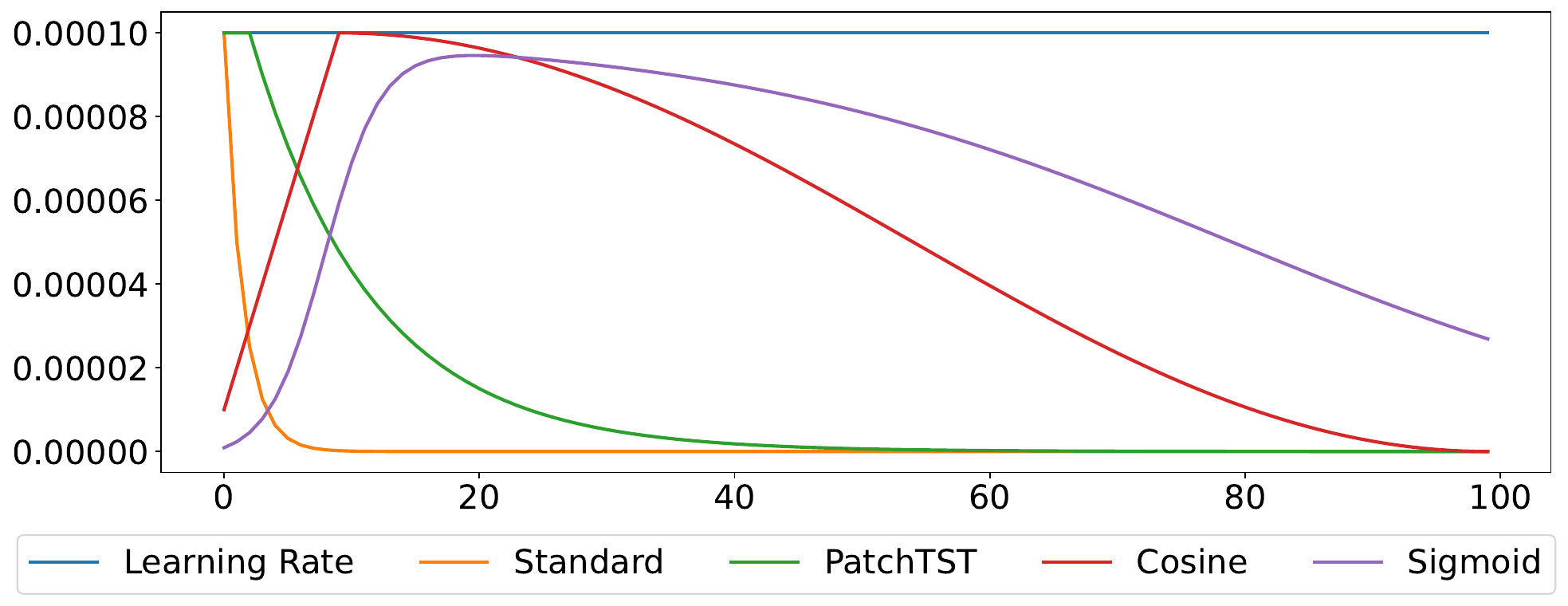}
\caption{LTSF learning rate adjustment strategies.}
\label{fig:lr}
\end{figure}

Following the experiment results, we follow the $k=0.5, s=10$, and $w=10$ hyperparameter combination.
Figure \ref{fig:lr} demonstrates Standard (\ref{equ:standard}), PatchTST (\ref{equ:patchtst}), Cosine (\ref{equ:cosine}), and Sigmoid (\ref{equ:sigmoid}) learning rate adjustment schemes.
The initial learning rate is set to $\alpha=0.0001$, the number of warm-up epochs $w=10$, the logistic growth rate $k=0.5$, and the decreasing curve smoothing rate $s=10$.

The effectiveness of the sigmoid adjustment approach was evaluated on PatchTST and CARD models since these are the only models that utilize long training with 100 epochs and that introduced their own learning rate adjustment techniques.
Table \ref{tab:lr-extended} presents the full comparative analysis between the original state-of-the-art models and versions that are trained using the proposed sigmoid learning rate adjustment scheme.
All models are tested with lookback $L=96$ for all datasets.

\begin{table*}[th]
  \scriptsize
  \centering
    \setlength\tabcolsep{3pt}
  \begin{tabular}{c|c|cccc|cccc|cccc}
    \hline
    \multicolumn{2}{c}{Method}      &
    \multicolumn{2}{|c}{xPatch}   & \multicolumn{2}{c}{xPatch*} & 
    \multicolumn{2}{|c}{CARD}   & \multicolumn{2}{c}{CARD*} & 
    \multicolumn{2}{|c}{PatchTST}    & \multicolumn{2}{c}{PatchTST*} \\
    \hline
    \multicolumn{2}{c|}{Metric}
    & MSE & MAE & MSE & MAE & MSE & MAE & MSE & MAE & MSE & MAE & MSE & MAE \\
    \hline
    \multirow{4}{*}{\rotatebox{90}{ETTh1}}
    & 96  & 0.385 & 0.395 & \textbf{0.376} & \textbf{0.386} & 0.383 & 0.391 & \textbf{0.379} & \textbf{0.390} & 0.393 & 0.407 & \textbf{0.377} & \textbf{0.396} \\
    & 192 & 0.422 & 0.408 & \textbf{0.417} & \textbf{0.407} & 0.435 & 0.420 & \textbf{0.433} & \textbf{0.419} & 0.445 & 0.434 & \textbf{0.429} & \textbf{0.424} \\
    & 336 & \textbf{0.445} & \textbf{0.423} & 0.449 & 0.425 & 0.479 & 0.442 & \textbf{0.475} & \textbf{0.438} & 0.483 & 0.451 & \textbf{0.472} & \textbf{0.443} \\
    & 720 & \textbf{0.469} & \textbf{0.455} & 0.470 & 0.456 & 0.471 & 0.461 & \textbf{0.470} & \textbf{0.459} & 0.479 & 0.470 & \textbf{0.475} & \textbf{0.466} \\
    \hline
    \multirow{4}{*}{\rotatebox{90}{ETTh2}}
    & 96  & 0.234 & 0.301 & \textbf{0.233} & \textbf{0.300} & \textbf{0.281} & \textbf{0.330} & 0.286 & 0.333 & 0.293 & 0.342 & \textbf{0.290} & \textbf{0.339} \\
    & 192 & \textbf{0.289} & \textbf{0.337} & 0.291 & 0.338 & 0.363 & 0.381 & \textbf{0.360} & \textbf{0.379} & 0.377 & 0.393 & \textbf{0.364} & \textbf{0.386} \\
    & 336 & \textbf{0.341} & \textbf{0.376} & 0.344 & 0.377 & 0.411 & 0.418 & \textbf{0.369} & \textbf{0.396} & 0.380 & 0.408 & \textbf{0.373} & \textbf{0.405} \\
    & 720 & \textbf{0.407} & 0.428 & \textbf{0.407} & \textbf{0.427} & 0.416 & 0.431 & \textbf{0.404} & \textbf{0.425} & 0.411 & \textbf{0.433} & \textbf{0.409} & \textbf{0.433} \\
    \hline
    \multirow{4}{*}{\rotatebox{90}{ETTm1}}
    & 96  & 0.318 & 0.353 & \textbf{0.311} & \textbf{0.346} & 0.316 & 0.347 & \textbf{0.314} & \textbf{0.344} & \textbf{0.320} & \textbf{0.359} & \textbf{0.320} & \textbf{0.359} \\
    & 192 & 0.350 & \textbf{0.368} & \textbf{0.348} & \textbf{0.368} & \textbf{0.363} & 0.370 & \textbf{0.363} & \textbf{0.369} & \textbf{0.365} & \textbf{0.381} & 0.366 & 0.384 \\
    & 336 & \textbf{0.386} & \textbf{0.391} & 0.388 & \textbf{0.391} & \textbf{0.392} & 0.390 & \textbf{0.392} & \textbf{0.389} & \textbf{0.391} & \textbf{0.401} & 0.392 & 0.403 \\
    & 720 & \textbf{0.460} & \textbf{0.428} & 0.461 & 0.430 & \textbf{0.458} & \textbf{0.425} & 0.462 & 0.427 & 0.455 & \textbf{0.436} & \textbf{0.454} & 0.438 \\
    \hline
    \multirow{4}{*}{\rotatebox{90}{ETTm2}}
    & 96  & 0.168 & 0.251 & \textbf{0.164} & \textbf{0.248} & 0.169 & \textbf{0.248} & \textbf{0.168} & \textbf{0.248} & 0.177 & 0.259 & \textbf{0.175} & \textbf{0.258} \\
    & 192 & 0.231 & 0.292 & \textbf{0.230} & \textbf{0.291} & 0.234 & 0.292 & \textbf{0.233} & \textbf{0.291} & 0.248 & 0.306 & \textbf{0.243} & \textbf{0.302} \\
    & 336 & 0.293 & \textbf{0.331} & \textbf{0.292} & \textbf{0.331} & 0.294 & 0.339 & \textbf{0.292} & \textbf{0.329} & 0.313 & 0.346 & \textbf{0.302} & \textbf{0.341} \\
    & 720 & \textbf{0.380} & 0.384 & 0.381 & \textbf{0.383} & \textbf{0.390} & \textbf{0.388} & 0.392 & \textbf{0.388} & \textbf{0.399} & 0.397 & 0.400 & \textbf{0.396} \\
    \hline
    \multirow{4}{*}{\rotatebox{90}{Weather}}
    & 96  & 0.178 & \textbf{0.212} & \textbf{0.170} & \textbf{0.212} & \textbf{0.150} & \textbf{0.188} & 0.154 & 0.192 & 0.177 & 0.218 & \textbf{0.173} & \textbf{0.215} \\
    & 192 & 0.223 & 0.251 & \textbf{0.210} & \textbf{0.248} & \textbf{0.202} & \textbf{0.238} & 0.205 & 0.240 & 0.225 & 0.259 & \textbf{0.219} & \textbf{0.256} \\
    & 336 & 0.241 & 0.276 & \textbf{0.226} & \textbf{0.273} & \textbf{0.260} & \textbf{0.282} & 0.264 & 0.285 & 0.277 & \textbf{0.297} & \textbf{0.275} & \textbf{0.297} \\
    & 720 & 0.313 & 0.323 & \textbf{0.309} & \textbf{0.321} & 0.343 & 0.353 & \textbf{0.342} & \textbf{0.337} & \textbf{0.350} & \textbf{0.345} & \textbf{0.350} & 0.346 \\
    \hline
    \multirow{4}{*}{\rotatebox{90}{Traffic}}
    & 96  & 0.482 & 0.286 & \textbf{0.471} & \textbf{0.275} & 0.419 & 0.269 & \textbf{0.412} & \textbf{0.257} & 0.446 & 0.283 & \textbf{0.432} & \textbf{0.276} \\
    & 192 & 0.479 & 0.280 & \textbf{0.478} & \textbf{0.272} & 0.443 & 0.276 & \textbf{0.430} & \textbf{0.263} & 0.453 & 0.285 & \textbf{0.443} & \textbf{0.279} \\
    & 336 & \textbf{0.498} & 0.284 & 0.501 & \textbf{0.276} & 0.460 & 0.283 & \textbf{0.445} & \textbf{0.270} & 0.467 & 0.291 & \textbf{0.455} & \textbf{0.286} \\
    & 720 & \textbf{0.530} & 0.284 & 0.501 & \textbf{0.291} & 0.490 & 0.299 & \textbf{0.473} & \textbf{0.286} & 0.500 & 0.309 & \textbf{0.490} & \textbf{0.304} \\
    \hline
    \multirow{4}{*}{\rotatebox{90}{Electricity}}
    & 96  & 0.166 & 0.249 & \textbf{0.159} & \textbf{0.244} & 0.141 & 0.233 & \textbf{0.136} & \textbf{0.229} & 0.166 & 0.252 & \textbf{0.165} & \textbf{0.251} \\
    & 192 & 0.166 & 0.253 & \textbf{0.160} & \textbf{0.248} & 0.160 & 0.250 & \textbf{0.155} & \textbf{0.245} & 0.174 & 0.260 & \textbf{0.172} & \textbf{0.259} \\
    & 336 & 0.189 & 0.272 & \textbf{0.182} & \textbf{0.267} & 0.173 & 0.263 & \textbf{0.168} & \textbf{0.259} & 0.190 & 0.277 & \textbf{0.189} & \textbf{0.276} \\
    & 720 & 0.229 & 0.306 & \textbf{0.216} & \textbf{0.298} & 0.197 & 0.284 & \textbf{0.193} & \textbf{0.280} & \textbf{0.230} & \textbf{0.311} & 0.231 & 0.313 \\
    \hline
    \multicolumn{2}{c|}{Gain} & & & 1.42\% & 1.04\% & & & 1.14\% & 1.34\% & & & 1.33\% & 0.74\% \\
    \hline
    \end{tabular}
  \caption{Comparison of forecasting errors between the baselines and the models trained with sigmoid learning rate adjustment strategy with unified lookback window $L = 96$.
  The model name with * denotes the model trained using the sigmoid learning rate adjustment technique.
    }
    \label{tab:lr-extended}
\end{table*}

\section{Lookback Window}
\label{app:lookback}
\begin{figure}[ht]
\centering
\includegraphics[width=1\columnwidth]{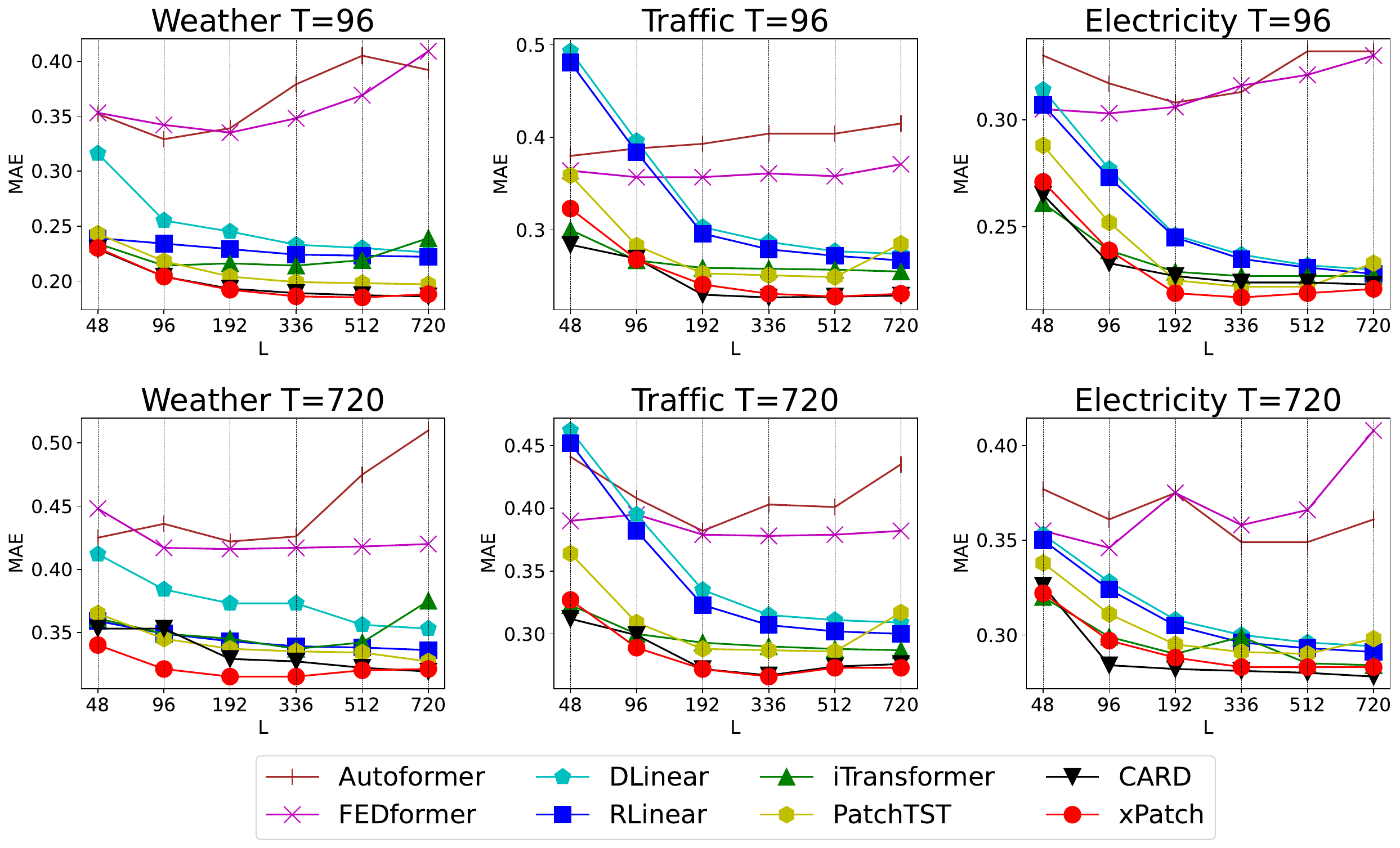}
\caption{Forecasting performance (MAE), lookback window $L = \{ 48, 96, 192, 336, 512, 720 \}$ and prediction horizon $T = 96$.}
\label{fig:lookback}
\end{figure}

In theory, a longer lookback window should increase the receptive field, potentially leading to improved forecasting performance.
However, most of the transformer-based solutions do not follow this assumption.
Transformer-based baselines do not necessarily benefit from longer historical data, which indicates the transformer architecture's ineffectiveness in capturing temporal dependencies.
The only transformer-based model that has a strong preservation of temporal relation is PatchTST.
Therefore, we hypothesize that this capability is largely attributed to the patching mechanism and can be better leveraged with other architectures, such as CNNs.
Figure \ref{fig:lookback} illustrates the ability of different models to learn from a longer lookback window.

\section{Instance Normalization}
\label{app:revin}
Table \ref{tab:revin} presents a comprehensive comparative analysis between the original state-of-the-art models and versions trained without using RevIN instance normalization.
All models are tested with a lookback $L=96$ for all datasets.

\begin{table}[th!]
  \scriptsize
  \centering
    \setlength\tabcolsep{1pt}
  \begin{tabular}{c|c|cccc|cccc|cccc}
    \hline
    \multicolumn{2}{c}{Method}      &
    \multicolumn{2}{|c}{xPatch}    & \multicolumn{2}{c}{xPatch*} & 
    \multicolumn{2}{|c}{CARD}  & \multicolumn{2}{c}{CARD*} & 
    \multicolumn{2}{|c}{PatchTST} & \multicolumn{2}{c}{PatchTST*} \\
    \hline
    \multicolumn{2}{c|}{Metric}
    & MSE & MAE & MSE & MAE & MSE & MAE & MSE & MAE & MSE & MAE & MSE & MAE \\
    \hline
    \multirow{4}{*}{\rotatebox{90}{ETTh2}}
    & 96  & 0.263 & 0.324 & \textbf{0.233} & \textbf{0.300} & 0.377 & 0.421 & \textbf{0.281} & \textbf{0.330} & 0.317 & 0.371 & \textbf{0.293} & \textbf{0.342} \\
    & 192 & 0.324 & 0.356 & \textbf{0.291} & \textbf{0.338} & 0.796 & 0.647 & \textbf{0.363} & \textbf{0.381} & 0.427 & 0.441 & \textbf{0.377} & \textbf{0.393} \\
    & 336 & 0.412 & 0.414 & \textbf{0.344} & \textbf{0.377} & 0.668 & 0.587 & \textbf{0.411} & \textbf{0.418} & 0.495 & 0.488 & \textbf{0.380} & \textbf{0.408} \\
    & 720 & 0.643 & 0.556 & \textbf{0.407} & \textbf{0.427} & 1.002 & 0.744 & \textbf{0.416} & \textbf{0.431} & 0.756 & 0.622 & \textbf{0.411} & \textbf{0.433} \\
    \hline
    \multirow{4}{*}{\rotatebox{90}{ETTm2}}
    & 96  & 0.176 & 0.261 & \textbf{0.164} & \textbf{0.248} & 0.203 & 0.297 & \textbf{0.169} & \textbf{0.248} & 0.187 & 0.274 & \textbf{0.177} & \textbf{0.259} \\
    & 192 & 0.250 & 0.310 & \textbf{0.230} & \textbf{0.291} & 0.372 & 0.412 & \textbf{0.234} & \textbf{0.292} & 0.264 & 0.323 & \textbf{0.248} & \textbf{0.306} \\
    & 336 & 0.331 & 0.358 & \textbf{0.292} & \textbf{0.331} & 0.484 & 0.484 & \textbf{0.294} & \textbf{0.339} & 0.354 & 0.377 & \textbf{0.313} & \textbf{0.346} \\
    & 720 & 0.443 & 0.421 & \textbf{0.381} & \textbf{0.383} & 0.729 & 0.607 & \textbf{0.390} & \textbf{0.388} & 0.462 & 0.446 & \textbf{0.399} & \textbf{0.397} \\
    \hline
    \multirow{4}{*}{\rotatebox{90}{Weather}}
    & 96  & 0.170 & 0.212 & \textbf{0.168} & \textbf{0.203} & 0.168 & 0.215 & \textbf{0.150} & \textbf{0.188} & 0.186 & 0.243 & \textbf{0.177} & \textbf{0.218} \\
    & 192 & \textbf{0.210} & 0.248 & 0.214 & \textbf{0.245} & 0.224 & 0.269 & \textbf{0.202} & \textbf{0.238} & \textbf{0.222} & 0.275 & 0.225 & \textbf{0.259} \\
    & 336 & \textbf{0.226} & \textbf{0.273} & 0.236 & \textbf{0.273} & 0.277 & 0.309 & \textbf{0.260} & \textbf{0.282} & \textbf{0.269} & 0.313 & 0.277 & \textbf{0.297} \\
    & 720 & \textbf{0.291} & 0.323 & 0.309 & \textbf{0.321} & 0.347 & 0.355 & \textbf{0.343} & \textbf{0.353} & \textbf{0.331} & 0.360 & 0.350 & \textbf{0.345} \\
    \hline
    \end{tabular}
  \caption{Comparison of forecasting errors between the baselines and the models with RevIN instance normalization with unified lookback window $L = 96$.
  The model name with * denotes the model trained with RevIN instance normalization.
    }
    \label{tab:revin}
\end{table}

The forecasting performance of xPatch was enhanced by the RevIN module, with improvements of 8.67\% in MSE and 6.53\% in MAE, respectively.
For CARD, instance normalization improved accuracy by 28.69\% in MSE and 23.22\% in MAE, while for PatchTST the gains are 9.96\% in MSE and 10.07\% in MAE, respectively.

The greater benefit of instance normalization observed in CARD \cite{wang2024card} and PatchTST \cite{yuqietal2023patch} can be attributed to xPatch's use of EMA decomposition.
According to non-stationary transformer \cite{liu2022non}, statistical methods like ARIMA \cite{box2015time} employ moving average decomposition for stationarization, while most recent state-of-the-art solutions rely on RevIN \cite{kim2021reversible}.
Consequently, xPatch incorporates two mechanisms to address data non-stationarity and distribution shift, which explains its superior performance compared to CARD and PatchTST even without the use of RevIN.

\section{Full Results}
\label{app:full-experiments}

Table \ref{tab:full-experiments} demonstrates the multivariate LTSF results averaged over three random seeds following the unified long-term forecasting protocol proposed by TimesNet \cite{wu2023timesnet} with unified lookback length $L=36$ for ILI dataset, and $L=96$ for all remaining datasets.

Baseline results for DLinear, TimesNet, ETSformer, FEDformer, and Autoformer are collected from the TimesNet official repository \cite{wu2023timesnet}, which is fairly structured according to each model's official code settings.
Results for CARD, TimeMixer, iTransformer, and MICN, which were implemented under the same experimental settings, are collected from their respective official papers.
The official manuscripts report the results of RLinear and PatchTST with a lookback length $L=336$.
Therefore, we reproduce these experiments using unified settings based on their publicly available official implementations.

Table \ref{tab:full-experiments-search} demonstrates the multivariate LTSF results averaged over three random seeds under hyperparameter searching with the best lookback length in $L=\{36,104,148\}$ for ILI dataset, and $L=\{96,192,336,512,720\}$ for all remaining datasets.

Baseline results for CARD, TimeMixer, PatchTST, MICN, and DLinear are collected from their official papers, as they either include a hyperparameter search for their models or provide the best results according to our hyperparameter search.
For iTransformer and RLinear, we reproduce the experiments with hyperparameter search based on their publicly available official implementations.
Additionally, we reproduced experiments for models that omit experiments on specific datasets (Exchange-rate, Solar-energy, and ILI).

Table \ref{tab:fair-experiments} demonstrates the fair multivariate LTSF results without using the “drop-last” trick averaged over three random seeds following the unified long-term forecasting protocol proposed by TFB \cite{qiu2024tfb} with the best lookback length in $L=\{36,104,148\}$ for ILI dataset, and $L=\{96,192,336,512,720\}$ for all remaining datasets.

Baseline results for PDF \cite{dai2024periodicity}, FITS \cite{xu2023fits}, and the remaining models are collected from the TFB \cite{qiu2024tfb} official repository, which is fairly structured according to each model's official code settings.

\begin{table*}[th!]
  \scriptsize
  \centering
  \setlength\tabcolsep{3pt}
  \begin{tabular}{c|c|cc|cc|cc|cc|cc|cc|cc|cc|cc|cc}
    \hline
    \multicolumn{2}{c}{\multirow{2}{*}{Models}}                                 &
    \multicolumn{2}{|c}{\textbf{xPatch}}    & \multicolumn{2}{|c}{CARD}         &
    \multicolumn{2}{|c}{TimeMixer}          & \multicolumn{2}{|c}{iTransformer} & 
    \multicolumn{2}{|c}{RLinear}            & \multicolumn{2}{|c}{PatchTST}     & 
    \multicolumn{2}{|c}{MICN}               & \multicolumn{2}{|c}{DLinear}      &
    \multicolumn{2}{|c}{TimesNet}           & \multicolumn{2}{|c}{ETSformer}    \\
    \multicolumn{2}{c}{\multirow{2}{*}{}}                                       &
    \multicolumn{2}{|c}{\textbf{(ours)}}    & \multicolumn{2}{|c}{(2024)}       & 
    \multicolumn{2}{|c}{(2024)}             & \multicolumn{2}{|c}{(2024)}       & 
    \multicolumn{2}{|c}{(2023)}             & \multicolumn{2}{|c}{(2023)}       & 
    \multicolumn{2}{|c}{(2023)}             & \multicolumn{2}{|c}{(2023)}       & 
    \multicolumn{2}{|c}{(2023)}             & \multicolumn{2}{|c}{(2022)}       \\
    \hline
    \multicolumn{2}{c|}{Metric}
    & MSE & MAE & MSE & MAE & MSE & MAE & MSE & MAE & MSE & MAE & MSE & MAE & MSE & MAE & MSE & MAE & MSE & MAE & MSE & MAE \\
    \hline
    \multirow{5}{*}{\rotatebox{90}{ETTh1}}
    & 96  & \underline{0.376} & \textbf{0.386} & 0.383 & \underline{0.391} & \textbf{0.375} & 0.400 & 0.386 & 0.405 & 0.380 & 0.392 & 0.393 & 0.407 & 0.421 & 0.431 & 0.386 & 0.400 & 0.384 & 0.402 & 0.494 & 0.479 \\
    & 192 & \textbf{0.417} & \textbf{0.407} & 0.435 & \underline{0.420} & \underline{0.429} & 0.421 & 0.441 & 0.436 & 0.433 & \underline{0.420} & 0.445 & 0.434 & 0.474 & 0.487 & 0.437 & 0.432 & 0.436 & 0.429 & 0.538 & 0.504 \\
    & 336 & \textbf{0.449} & \textbf{0.425} & 0.479 & 0.442 & 0.484 & 0.458 & 0.487 & 0.458 & \underline{0.470} & \underline{0.437} & 0.483 & 0.451 & 0.569 & 0.551 & 0.481 & 0.459 & 0.491 & 0.469 & 0.574 & 0.521 \\
    & 720 & \underline{0.470} & \textbf{0.456} & 0.471 & 0.461 & 0.498 & 0.482 & 0.503 & 0.491 & \textbf{0.467} & \underline{0.460} & 0.479 & 0.470 & 0.770 & 0.672 & 0.519 & 0.516 & 0.521 & 0.500 & 0.562 & 0.535 \\
    \cline{2-22}
    & Avg & \textbf{0.428} & \textbf{0.419} & 0.442 & 0.429 & 0.447 & 0.440 & 0.454 & 0.448 & \underline{0.438} & \underline{0.427} & 0.450 & 0.441 & 0.559 & 0.535 & 0.456 & 0.452 & 0.458 & 0.450 & 0.542 & 0.510 \\
    \hline
    \multirow{5}{*}{\rotatebox{90}{ETTh2}}
    & 96  & \textbf{0.233} & \textbf{0.300} & 0.281 & \underline{0.330} & 0.289 & 0.341 & 0.297 & 0.349 & \underline{0.278} & 0.333 & 0.293 & 0.342 & 0.299 & 0.364 & 0.333 & 0.387 & 0.340 & 0.374 & 0.340 & 0.391 \\
    & 192 & \textbf{0.291} & \textbf{0.338} & 0.363 & \underline{0.381} & 0.372 & 0.392 & 0.380 & 0.400 & \underline{0.360} & 0.387 & 0.377 & 0.393 & 0.441 & 0.454 & 0.477 & 0.476 & 0.402 & 0.414 & 0.430 & 0.439 \\
    & 336 & \textbf{0.344} & \textbf{0.377} & 0.411 & 0.418 & 0.386 & 0.414 & 0.428 & 0.432 & \underline{0.379} & 0.410 & 0.380 & \underline{0.408} & 0.654 & 0.567 & 0.594 & 0.541 & 0.452 & 0.452 & 0.485 & 0.479 \\
    & 720 & \textbf{0.407} & \textbf{0.427} & 0.416 & \underline{0.431} & 0.412 & 0.434 & 0.427 & 0.445 & 0.430 & 0.445 & \underline{0.411} & 0.433 & 0.956 & 0.716 & 0.831 & 0.657 & 0.462 & 0.468 & 0.500 & 0.497 \\
    \cline{2-22}
    & Avg & \textbf{0.319} & \textbf{0.361} & 0.368 & \underline{0.390} & 0.365 & 0.395 & 0.383 & 0.407 & \underline{0.362} & 0.394 & 0.365 & 0.394 & 0.588 & 0.525 & 0.559 & 0.515 & 0.414 & 0.427 & 0.439 & 0.452 \\
    \hline
    \multirow{5}{*}{\rotatebox{90}{ETTm1}}
    & 96  & \textbf{0.311} & \textbf{0.346} & \underline{0.316} & \underline{0.347} & 0.320 & 0.357 & 0.334 & 0.368 & 0.351 & 0.369 & 0.320 & 0.359 & \underline{0.316} & 0.362 & 0.345 & 0.372 & 0.338 & 0.375 & 0.375 & 0.398 \\
    & 192 & \textbf{0.348} & \textbf{0.368} & 0.363 & \underline{0.370} & \underline{0.361} & 0.381 & 0.377 & 0.391 & 0.388 & 0.386 & 0.365 & 0.381 & 0.363 & 0.390 & 0.380 & 0.389 & 0.374 & 0.387 & 0.408 & 0.410 \\
    & 336 & \textbf{0.388} & \underline{0.391} & 0.392 & \textbf{0.390} & \underline{0.390} & 0.404 & 0.426 & 0.420 & 0.420 & 0.407 & 0.391 & 0.401 & 0.408 & 0.426 & 0.413 & 0.413 & 0.410 & 0.411 & 0.435 & 0.428 \\
    & 720 & 0.461 & \underline{0.430} & 0.458 & \textbf{0.425} & \textbf{0.454} & 0.441 & 0.491 & 0.459 & 0.478 & 0.440 & \underline{0.455} & 0.436 & 0.481 & 0.476 & 0.474 & 0.453 & 0.478 & 0.450 & 0.499 & 0.462 \\
    \cline{2-22}
    & Avg & \textbf{0.377} & \underline{0.384} & 0.382 & \textbf{0.383} & \underline{0.381} & 0.396 & 0.407 & 0.410 & 0.409 & 0.401 & 0.383 & 0.394 & 0.392 & 0.414 & 0.403 & 0.407 & 0.400 & 0.406 & 0.429 & 0.425 \\
    \hline
    \multirow{5}{*}{\rotatebox{90}{ETTm2}}
    & 96  & \textbf{0.164} & \textbf{0.248} & \underline{0.169} & \textbf{0.248} & 0.175 & \underline{0.258} & 0.180 & 0.264 & 0.182 & 0.265 & 0.177 & 0.259 & 0.179 & 0.275 & 0.193 & 0.292 & 0.187 & 0.267 & 0.189 & 0.280 \\
    & 192 & \textbf{0.230} & \textbf{0.291} & \underline{0.234} & \underline{0.292} & 0.237 & 0.299 & 0.250 & 0.309 & 0.247 & 0.305 & 0.248 & 0.306 & 0.307 & 0.376 & 0.284 & 0.362 & 0.249 & 0.309 & 0.253 & 0.319 \\
    & 336 & \textbf{0.292} & \textbf{0.331} & \underline{0.294} & \underline{0.339} & 0.298 & 0.340 & 0.311 & 0.348 & 0.309 & 0.343 & 0.313 & 0.346 & 0.325 & 0.388 & 0.369 & 0.427 & 0.321 & 0.351 & 0.314 & 0.357 \\
    & 720 & \textbf{0.381} & \textbf{0.383} & \underline{0.390} & \underline{0.388} & 0.391 & 0.396 & 0.412 & 0.407 & 0.405 & 0.397 & 0.399 & 0.397 & 0.502 & 0.490 & 0.554 & 0.522 & 0.408 & 0.403 & 0.414 & 0.413 \\
    \cline{2-22}
    & Avg & \textbf{0.267} & \textbf{0.313} & \underline{0.272} & \underline{0.317} & 0.275 & 0.323 & 0.288 & 0.332 & 0.286 & 0.328 & 0.284 & 0.327 & 0.328 & 0.382 & 0.350 & 0.401 & 0.291 & 0.333 & 0.293 & 0.342 \\
    \hline
    \multirow{5}{*}{\rotatebox{90}{Weather}}
    & 96  & 0.168 & \underline{0.203} & \textbf{0.150} & \textbf{0.188} & 0.163 & 0.209 & 0.174 & 0.214 & 0.194 & 0.234 & 0.177 & 0.218 & \underline{0.161} & 0.229 & 0.196 & 0.255 & 0.172 & 0.220 & 0.197 & 0.281 \\
    & 192 & 0.214 & \underline{0.245} & \textbf{0.202} & \textbf{0.238} & \underline{0.208} & 0.250 & 0.221 & 0.254 & 0.238 & 0.269 & 0.225 & 0.259 & 0.220 & 0.281 & 0.237 & 0.296 & 0.219 & 0.261 & 0.237 & 0.312 \\
    & 336 & \textbf{0.236} & \textbf{0.273} & 0.260 & \underline{0.282} & \underline{0.251} & 0.287 & 0.278 & 0.296 & 0.287 & 0.304 & 0.277 & 0.297 & 0.278 & 0.331 & 0.283 & 0.335 & 0.280 & 0.306 & 0.298 & 0.353 \\
    & 720 & \textbf{0.309} & \textbf{0.321} & 0.343 & 0.353 & 0.339 & \underline{0.341} & 0.358 & 0.349 & 0.355 & 0.349 & 0.350 & 0.345 & \underline{0.311} & 0.356 & 0.345 & 0.381 & 0.365 & 0.359 & 0.352 & 0.388 \\
    \cline{2-22}
    & Avg & \textbf{0.232} & \textbf{0.261} & \underline{0.239} & \underline{0.265} & 0.240 & 0.272 & 0.258 & 0.278 & 0.269 & 0.289 & 0.257 & 0.280 & 0.243 & 0.299 & 0.265 & 0.317 & 0.259 & 0.287 & 0.271 & 0.334 \\
    \hline
    \multirow{5}{*}{\rotatebox{90}{Traffic}}
    & 96  & 0.471 & 0.275 & \underline{0.419} & \underline{0.269} & 0.462 & 0.285 & \textbf{0.395} & \textbf{0.268} & 0.646 & 0.384 & 0.446 & 0.283 & 0.519 & 0.309 & 0.650 & 0.396 & 0.593 & 0.321 & 0.607 & 0.392 \\
    & 192 & 0.478 & \textbf{0.272} & \underline{0.443} & \underline{0.276} & 0.473 & 0.296 & \textbf{0.417} & \underline{0.276} & 0.598 & 0.360 & 0.453 & 0.285 & 0.537 & 0.315 & 0.598 & 0.370 & 0.617 & 0.336 & 0.621 & 0.399 \\
    & 336 & 0.501 & \textbf{0.276} & \underline{0.460} & \underline{0.283} & 0.498 & 0.296 & \textbf{0.433} & \underline{0.283} & 0.605 & 0.363 & 0.467 & 0.291 & 0.534 & 0.313 & 0.605 & 0.373 & 0.629 & 0.336 & 0.622 & 0.396 \\
    & 720 & 0.547 & \textbf{0.291} & \underline{0.490} & \underline{0.299} & 0.506 & 0.313 & \textbf{0.467} & 0.302 & 0.643 & 0.382 & 0.500 & 0.309 & 0.577 & 0.325 & 0.645 & 0.394 & 0.640 & 0.350 & 0.632 & 0.396 \\
    \cline{2-22}
    & Avg & 0.499 & \textbf{0.279} & \underline{0.453} & \underline{0.282} & 0.485 & 0.298 & \textbf{0.428} & \underline{0.282} & 0.623 & 0.372 & 0.467 & 0.292 & 0.542 & 0.316 & 0.625 & 0.383 & 0.620 & 0.336 & 0.621 & 0.396 \\
    \hline
    \multirow{5}{*}{\rotatebox{90}{Electricity}}
    & 96  & 0.159 & 0.244 & \textbf{0.141} & \textbf{0.233} & 0.153 & 0.247 & \underline{0.148} & \underline{0.240} & 0.197 & 0.273 & 0.166 & 0.252 & 0.164 & 0.269 & 0.197 & 0.282 & 0.168 & 0.272 & 0.187 & 0.304 \\
    & 192 & \textbf{0.160} & \textbf{0.248} & \textbf{0.160} & \underline{0.250} & 0.166 & 0.256 & 0.162 & 0.253 & 0.196 & 0.276 & 0.174 & 0.260 & 0.177 & 0.285 & 0.196 & 0.285 & 0.184 & 0.289 & 0.199 & 0.315 \\
    & 336 & 0.182 & \underline{0.267} & \textbf{0.173} & \textbf{0.263} & 0.185 & 0.277 & \underline{0.178} & 0.269 & 0.211 & 0.291 & 0.190 & 0.277 & 0.193 & 0.304 & 0.209 & 0.301 & 0.198 & 0.300 & 0.212 & 0.329 \\
    & 720 & 0.216 & \underline{0.298} & \textbf{0.197} & \textbf{0.284} & 0.225 & 0.310 & 0.225 & 0.317 & 0.253 & 0.324 & 0.230 & 0.311 & \underline{0.212} & 0.321 & 0.245 & 0.333 & 0.220 & 0.320 & 0.233 & 0.345 \\
    \cline{2-22}
    & Avg & 0.179 & \underline{0.264} & \textbf{0.168} & \textbf{0.258} & 0.182 & 0.273 & \underline{0.178} & 0.270 & 0.214 & 0.291 & 0.190 & 0.275 & 0.187 & 0.295 & 0.212 & 0.300 & 0.193 & 0.295 & 0.208 & 0.323 \\
    \hline
    \multirow{5}{*}{\rotatebox{90}{Exchange}}
    & 96  & \underline{0.082} & \underline{0.199} & 0.084 & 0.202 & 0.087 & 0.206 & 0.086 & 0.206 & \underline{0.082} & 0.200 & \textbf{0.080} & \textbf{0.196} & 0.102 & 0.235 & 0.088 & 0.218 & 0.107 & 0.234 & 0.085 & 0.204 \\
    & 192 & 0.177 & 0.298 & 0.174 & \underline{0.295} & 0.193 & 0.310 & 0.177 & 0.299 & 0.179 & 0.300 & \textbf{0.171} & \textbf{0.293} & \underline{0.172} & 0.316 & 0.176 & 0.315 & 0.226 & 0.344 & 0.182 & 0.303 \\
    & 336 & 0.349 & 0.425 & 0.342 & 0.421 & 0.345 & 0.425 & 0.331 & 0.417 & 0.346 & 0.423 & 0.317 & \textbf{0.406} & \textbf{0.272} & \underline{0.407} & \underline{0.313} & 0.427 & 0.367 & 0.448 & 0.348 & 0.428 \\
    & 720 & 0.891 & 0.711 & 0.841 & \underline{0.689} & 1.008 & 0.747 & 0.847 & 0.691 & 0.913 & 0.717 & 0.887 & 0.703 & \textbf{0.714} & \textbf{0.658} & \underline{0.839} & 0.695 & 0.964 & 0.746 & 1.025 & 0.774 \\
    \cline{2-22}
    & Avg & 0.375 & 0.408 & 0.360 & \underline{0.402} & 0.408 & 0.422 & 0.360 & 0.403 & 0.380 & 0.410 & 0.364 & \textbf{0.400} & \textbf{0.315} & 0.404 & \underline{0.354} & 0.414 & 0.416 & 0.443 & 0.410 & 0.427 \\
    \hline
    \multirow{5}{*}{\rotatebox{90}{Solar}}
    & 96  & 0.201 & \underline{0.215} & \underline{0.194} & \textbf{0.209} & \textbf{0.189} & 0.259 & 0.203 & 0.237 & 0.322 & 0.339 & 0.225 & 0.270 & 0.257 & 0.325 & 0.287 & 0.374 & 0.250 & 0.292 & 0.258 & 0.371 \\
    & 192 & 0.235 & \textbf{0.234} & 0.234 & \underline{0.235} & \textbf{0.222} & 0.283 & \underline{0.233} & 0.261 & 0.360 & 0.358 & 0.253 & 0.288 & 0.278 & 0.354 & 0.317 & 0.395 & 0.296 & 0.318 & 0.608 & 0.606 \\
    & 336 & 0.258 & \textbf{0.249} & 0.256 & \underline{0.253} & \textbf{0.231} & 0.292 & \underline{0.248} & 0.273 & 0.397 & 0.369 & 0.270 & 0.298 & 0.298 & 0.375 & 0.350 & 0.413 & 0.319 & 0.330 & 0.758 & 0.705 \\
    & 720 & 0.260 & \textbf{0.247} & 0.262 & \underline{0.257} & \textbf{0.223} & 0.285 & \underline{0.249} & 0.275 & 0.396 & 0.361 & 0.269 & 0.299 & 0.299 & 0.379 & 0.355 & 0.411 & 0.338 & 0.337 & 0.789 & 0.779 \\
    \cline{2-22}
    & Avg & 0.239 & \textbf{0.236} & 0.237 & \underline{0.239} & \textbf{0.216} & 0.280 & \underline{0.233} & 0.262 & 0.369 & 0.357 & 0.254 & 0.289 & 0.283 & 0.358 & 0.327 & 0.398 & 0.301 & 0.319 & 0.603 & 0.615 \\
    \hline
    \multirow{5}{*}{\rotatebox{90}{ILI}}
    & 24  & \textbf{1.378} & \textbf{0.685} & \underline{1.665} & \underline{0.803} & 1.897 & 0.840 & 2.915 & 1.146 & 2.517 & 1.002 & 1.691 & 0.816 & 2.684 & 1.112 & 2.398 & 1.040 & 2.317 & 0.934 & 2.527 & 1.020 \\
    & 36 & \textbf{1.315} & \textbf{0.681} & 2.200 & 0.890 & \underline{1.405} & \underline{0.747} & 2.924 & 1.157 & 2.443 & 0.960 & 1.415 & 0.762 & 2.667 & 1.068 & 2.646 & 1.088 & 1.972 & 0.920 & 2.615 & 1.007 \\
    & 48 & \textbf{1.459} & \textbf{0.747} & 1.875 & 0.821 & \underline{1.640} & \underline{0.809} & 2.835 & 1.134 & 2.344 & 0.950 & 1.754 & 0.819 & 2.558 & 1.052 & 2.614 & 1.086 & 2.238 & 0.940 & 2.359 & 0.972 \\
    & 60 & \textbf{1.616} & \textbf{0.787} & 1.923 & 0.853 & 1.890 & 0.884 & 2.996 & 1.178 & 2.503 & 0.999 & \underline{1.645} & \underline{0.820} & 2.747 & 1.110 & 2.804 & 1.146 & 2.027 & 0.928 & 2.487 & 1.016 \\
    \cline{2-22}
    & Avg & \textbf{1.442} & \textbf{0.725} & 1.916 & 0.842 & 1.708 & 0.820 & 2.918 & 1.154 & 2.452 & 0.978 & \underline{1.626} & \underline{0.804} & 2.664 & 1.086 & 2.616 & 1.090 & 2.139 & 0.931 & 2.497 & 1.004 \\
    \hline
    \multicolumn{2}{c|}{$1^{st}$ Count} & \textbf{26} & \textbf{34} & \underline{7} & \underline{11} & \underline{7} & 0 & 5 & 1 & 1 & 0 & 2 & 4 & 3 & 1 & 0 & 0 & 0 & 0 & 0 & 0 \\
    \hline
    \end{tabular}
  \caption{Full long-term forecasting results with unified lookback window $L = 36$ for the ILI dataset, and $L = 96$ for all other datasets.
  The best model is \textbf{boldface} and the second best is \underline{underlined}.}
    \label{tab:full-experiments}
\end{table*}

\begin{table*}[th!]
  \scriptsize
  \centering
  \setlength\tabcolsep{3pt}
  \begin{tabular}{c|c|cc|cc|cc|cc|cc|cc|cc|cc|cc|cc}
    \hline
    \multicolumn{2}{c}{\multirow{2}{*}{Models}}                                 &
    \multicolumn{2}{|c}{\textbf{xPatch}}    & \multicolumn{2}{|c}{CARD}         &
    \multicolumn{2}{|c}{TimeMixer}          & \multicolumn{2}{|c}{iTransformer} & 
    \multicolumn{2}{|c}{RLinear}            & \multicolumn{2}{|c}{PatchTST}     & 
    \multicolumn{2}{|c}{MICN}               & \multicolumn{2}{|c}{DLinear}      &
    \multicolumn{2}{|c}{TimesNet}           & \multicolumn{2}{|c}{ETSformer}    \\
    \multicolumn{2}{c}{\multirow{2}{*}{}}                                       &
    \multicolumn{2}{|c}{\textbf{(ours)}}    & \multicolumn{2}{|c}{(2024)}       & 
    \multicolumn{2}{|c}{(2024)}             & \multicolumn{2}{|c}{(2024)}       & 
    \multicolumn{2}{|c}{(2023)}             & \multicolumn{2}{|c}{(2023)}       & 
    \multicolumn{2}{|c}{(2023)}             & \multicolumn{2}{|c}{(2023)}       & 
    \multicolumn{2}{|c}{(2023)}             & \multicolumn{2}{|c}{(2022)}       \\
    \hline
    \multicolumn{2}{c|}{Metric}
    & MSE & MAE & MSE & MAE & MSE & MAE & MSE & MAE & MSE & MAE & MSE & MAE & MSE & MAE & MSE & MAE & MSE & MAE & MSE & MAE \\
    \hline
    \multirow{5}{*}{\rotatebox{90}{ETTh1}}
    & 96  & \textbf{0.354} & \textbf{0.379} & 0.368 & 0.396 & \underline{0.361} & \underline{0.390} & 0.396 & 0.425 & 0.364 & 0.391 & 0.370 & 0.400 & 0.398 & 0.427 & 0.375 & 0.399 & 0.384 & 0.402 & 0.494 & 0.479 \\
    & 192 & \textbf{0.376} & \textbf{0.395} & 0.406 & 0.418 & 0.409 & \underline{0.414} & 0.430 & 0.450 & 0.418 & 0.429 & 0.413 & 0.429 & 0.430 & 0.453 & \underline{0.405} & 0.416 & 0.436 & 0.429 & 0.538 & 0.504 \\
    & 336 & \textbf{0.391} & \textbf{0.415} & \underline{0.415} & \underline{0.424} & 0.430 & 0.429 & 0.479 & 0.485 & 0.418 & 0.425 & 0.422 & 0.440 & 0.440 & 0.460 & 0.439 & 0.443 & 0.491 & 0.469 & 0.574 & 0.521 \\
    & 720 & \underline{0.442} & \underline{0.459} & \textbf{0.416} & \textbf{0.448} & 0.445 & 0.460 & 0.700 & 0.608 & 0.450 & 0.462 & 0.447 & 0.468 & 0.491 & 0.509 & 0.472 & 0.490 & 0.521 & 0.500 & 0.562 & 0.535 \\
    \cline{2-22}
    & Avg & \textbf{0.391} & \textbf{0.412} & \underline{0.401} & \underline{0.422} & 0.411 & 0.423 & 0.501 & 0.492 & 0.413 & 0.427 & 0.413 & 0.434 & 0.440 & 0.462 & 0.423 & 0.437 & 0.458 & 0.450 & 0.542 & 0.510 \\
    \hline
    \multirow{5}{*}{\rotatebox{90}{ETTh2}}
    & 96  & \textbf{0.226} & \textbf{0.297} & 0.262 & \underline{0.327} & 0.271 & 0.330 & 0.311 & 0.363 & \underline{0.255} & \underline{0.327} & 0.274 & 0.337 & 0.299 & 0.364 & 0.289 & 0.353 & 0.340 & 0.374 & 0.340 & 0.391 \\
    & 192 & \textbf{0.275} & \textbf{0.330} & 0.322 & \underline{0.369} & 0.317 & 0.402 & 0.391 & 0.413 & \underline{0.317} & 0.371 & 0.341 & 0.382 & 0.422 & 0.441 & 0.383 & 0.418 & 0.402 & 0.414 & 0.430 & 0.439 \\
    & 336 & \textbf{0.312} & \textbf{0.360} & 0.326 & \underline{0.378} & 0.332 & 0.396 & 0.415 & 0.437 & \underline{0.324} & 0.385 & 0.329 & 0.384 & 0.447 & 0.474 & 0.448 & 0.465 & 0.452 & 0.452 & 0.485 & 0.479 \\
    & 720 & 0.384 & \underline{0.418} & \underline{0.373} & 0.419 & \textbf{0.342} & \textbf{0.408} & 0.424 & 0.455 & 0.414 & 0.445 & 0.379 & 0.422 & 0.442 & 0.467 & 0.605 & 0.551 & 0.462 & 0.468 & 0.500 & 0.497 \\
    \cline{2-22}
    & Avg & \textbf{0.299} & \textbf{0.351} & 0.321 & \underline{0.373} & \underline{0.316} & 0.384 & 0.385 & 0.417 & 0.328 & 0.382 & 0.331 & 0.381 & 0.403 & 0.437 & 0.431 & 0.447 & 0.414 & 0.427 & 0.439 & 0.452 \\
    \hline
    \multirow{5}{*}{\rotatebox{90}{ETTm1}}
    & 96  & \textbf{0.275} & \textbf{0.330} & \underline{0.288} & \underline{0.332} & 0.291 & 0.340 & 0.313 & 0.366 & 0.310 & 0.350 & 0.293 & 0.346 & 0.316 & 0.364 & 0.299 & 0.343 & 0.338 & 0.375 & 0.375 & 0.398 \\
    & 192 & \textbf{0.315} & \textbf{0.355} & 0.332 & \underline{0.357} & \underline{0.327} & 0.365 & 0.349 & 0.388 & 0.337 & 0.366 & 0.333 & 0.370 & 0.363 & 0.390 & 0.335 & 0.365 & 0.374 & 0.387 & 0.408 & 0.410 \\
    & 336 & \textbf{0.355} & \textbf{0.376} & 0.364 & \textbf{0.376} & \underline{0.360} & \underline{0.381} & 0.381 & 0.411 & 0.369 & 0.384 & 0.369 & 0.392 & 0.408 & 0.426 & 0.369 & 0.386 & 0.410 & 0.411 & 0.435 & 0.428 \\
    & 720 & 0.419 & \underline{0.411} & \textbf{0.414} & \textbf{0.407} & \underline{0.415} & 0.417 & 0.448 & 0.449 & 0.419 & \underline{0.411} & 0.416 & 0.420 & 0.459 & 0.464 & 0.425 & 0.421 & 0.478 & 0.450 & 0.499 & 0.462 \\
    \cline{2-22}
    & Avg & \textbf{0.341} & \textbf{0.368} & 0.350 & \textbf{0.368} & \underline{0.348} & \underline{0.376} & 0.373 & 0.404 & 0.359 & 0.378 & 0.353 & 0.382 & 0.387 & 0.411 & 0.357 & 0.379 & 0.400 & 0.406 & 0.429 & 0.425 \\
    \hline
    \multirow{5}{*}{\rotatebox{90}{ETTm2}}
    & 96  & \textbf{0.153} & \textbf{0.240} & \underline{0.159} & \underline{0.246} & 0.164 & 0.254 & 0.180 & 0.274 & 0.163 & 0.251 & 0.166 & 0.256 & 0.179 & 0.275 & 0.167 & 0.260 & 0.187 & 0.267 & 0.189 & 0.280 \\
    & 192 & \textbf{0.213} & \textbf{0.280} & \underline{0.214} & \underline{0.285} & 0.223 & 0.295 & 0.238 & 0.311 & 0.218 & 0.289 & 0.223 & 0.296 & 0.262 & 0.326 & 0.224 & 0.303 & 0.249 & 0.309 & 0.253 & 0.319 \\
    & 336 & \textbf{0.264} & \textbf{0.315} & \underline{0.266} & \underline{0.319} & 0.279 & 0.330 & 0.294 & 0.349 & 0.271 & 0.325 & 0.274 & 0.329 & 0.305 & 0.353 & 0.281 & 0.342 & 0.321 & 0.351 & 0.314 & 0.357 \\
    & 720 & \textbf{0.338} & \textbf{0.363} & 0.379 & 0.390 & \underline{0.359} & \underline{0.383} & 0.382 & 0.406 & 0.360 & 0.385 & 0.362 & 0.385 & 0.389 & 0.407 & 0.397 & 0.421 & 0.408 & 0.403 & 0.414 & 0.413 \\
    \cline{2-22}
    & Avg & \textbf{0.242} & \textbf{0.300} & 0.255 & \underline{0.310} & 0.256 & 0.316 & 0.274 & 0.335 & \underline{0.253} & 0.313 & 0.256 & 0.317 & 0.284 & 0.340 & 0.267 & 0.332 & 0.291 & 0.333 & 0.293 & 0.342 \\
    \hline
    \multirow{5}{*}{\rotatebox{90}{Weather}}
    & 96  & \underline{0.146} & \textbf{0.185} & \textbf{0.145} & \underline{0.186} & 0.147 & 0.197 & 0.191 & 0.239 & 0.171 & 0.223 & 0.149 & 0.198 & 0.161 & 0.229 & 0.176 & 0.237 & 0.172 & 0.220 & 0.197 & 0.281 \\
    & 192 & \underline{0.189} & \textbf{0.227} & \textbf{0.187} & \textbf{0.227} & \underline{0.189} & \underline{0.239} & 0.219 & 0.263 & 0.215 & 0.259 & 0.194 & 0.241 & 0.220 & 0.281 & 0.220 & 0.282 & 0.219 & 0.261 & 0.237 & 0.312 \\
    & 336 & \textbf{0.218} & \underline{0.260} & \underline{0.238} & \textbf{0.258} & 0.241 & 0.280 & 0.284 & 0.311 & 0.261 & 0.293 & 0.245 & 0.282 & 0.278 & 0.331 & 0.265 & 0.319 & 0.280 & 0.306 & 0.298 & 0.353 \\
    & 720 & \textbf{0.291} & \textbf{0.315} & \underline{0.308} & \underline{0.321} & 0.310 & 0.330 & 0.389 & 0.375 & 0.322 & 0.338 & 0.314 & 0.334 & 0.311 & 0.356 & 0.323 & 0.362 & 0.365 & 0.359 & 0.352 & 0.388 \\
    \cline{2-22}
    & Avg & \textbf{0.211} & \textbf{0.247} & \underline{0.220} & \underline{0.248} & 0.222 & 0.262 & 0.271 & 0.297 & 0.242 & 0.278 & 0.226 & 0.264 & 0.243 & 0.299 & 0.246 & 0.300 & 0.259 & 0.287 & 0.271 & 0.334 \\
    \hline
    \multirow{5}{*}{\rotatebox{90}{Traffic}}
    & 96  & 0.364 & \underline{0.233} & \textbf{0.341} & \textbf{0.229} & 0.360 & 0.249 & \underline{0.348} & 0.255 & 0.395 & 0.272 & 0.360 & 0.249 & 0.519 & 0.309 & 0.410 & 0.282 & 0.593 & 0.321 & 0.607 & 0.392 \\
    & 192 & 0.377 & \textbf{0.241} & \underline{0.367} & \underline{0.243} & 0.375 & 0.250 & \textbf{0.366} & 0.266 & 0.406 & 0.276 & 0.379 & 0.256 & 0.537 & 0.315 & 0.423 & 0.287 & 0.617 & 0.336 & 0.621 & 0.399 \\
    & 336 & 0.388 & \textbf{0.243} & 0.388 & \underline{0.254} & \underline{0.385} & 0.270 & \textbf{0.383} & 0.273 & 0.415 & 0.281 & 0.392 & 0.264 & 0.534 & 0.313 & 0.436 & 0.296 & 0.629 & 0.336 & 0.622 & 0.396 \\
    & 720 & 0.437 & \textbf{0.273} & \underline{0.427} & \underline{0.276} & 0.430 & 0.281 & \textbf{0.413} & 0.287 & 0.453 & 0.302 & 0.432 & 0.286 & 0.577 & 0.325 & 0.466 & 0.315 & 0.640 & 0.350 & 0.632 & 0.396 \\
    \cline{2-22}
    & Avg & 0.392 & \textbf{0.248} & \underline{0.381} & \underline{0.251} & 0.388 & 0.263 & \textbf{0.378} & 0.270 & 0.417 & 0.283 & 0.391 & 0.264 & 0.542 & 0.316 & 0.434 & 0.295 & 0.620 & 0.336 & 0.621 & 0.396 \\
    \hline
    \multirow{5}{*}{\rotatebox{90}{Electricity}}
    & 96  & \textbf{0.126} & \textbf{0.217} & \underline{0.129} & 0.223 & \underline{0.129} & 0.224 & 0.131 & 0.227 & 0.136 & 0.231 & \underline{0.129} & \underline{0.222} & 0.164 & 0.269 & 0.140 & 0.237 & 0.168 & 0.272 & 0.187 & 0.304 \\
    & 192 & \textbf{0.140} & \underline{0.232} & 0.154 & 0.245 & \textbf{0.140} & \textbf{0.220} & 0.152 & 0.248 & 0.149 & 0.243 & \underline{0.147} & 0.240 & 0.177 & 0.285 & 0.153 & 0.249 & 0.184 & 0.289 & 0.199 & 0.315 \\
    & 336 & \textbf{0.156} & \textbf{0.249} & \underline{0.161} & 0.257 & \underline{0.161} & \underline{0.255} & 0.170 & 0.267 & 0.165 & 0.259 & 0.163 & 0.259 & 0.193 & 0.304 & 0.169 & 0.267 & 0.198 & 0.300 & 0.212 & 0.329 \\
    & 720 & \underline{0.190} & \underline{0.281} & \textbf{0.185} & \textbf{0.278} & 0.194 & 0.287 & \underline{0.190} & 0.284 & 0.205 & 0.293 & 0.197 & 0.290 & 0.212 & 0.321 & 0.203 & 0.301 & 0.220 & 0.320 & 0.233 & 0.345 \\
    \cline{2-22}
    & Avg & \textbf{0.153} & \textbf{0.245} & 0.157 & 0.251 & \underline{0.156} & \underline{0.247} & 0.161 & 0.257 & 0.164 & 0.257 & 0.159 & 0.253 & 0.187 & 0.295 & 0.166 & 0.264 & 0.193 & 0.295 & 0.208 & 0.323 \\
    \hline
    \multirow{5}{*}{\rotatebox{90}{Exchange}}
    & 96  & \textbf{0.081} & \textbf{0.197} & \underline{0.084} & \underline{0.202} & 0.096 & 0.219 & 0.108 & 0.239 & 0.090 & 0.211 & 0.086 & 0.208 & 0.102 & 0.235 & \textbf{0.081} & 0.203 & 0.107 & 0.234 & 0.085 & 0.204 \\
    & 192 & 0.178 & 0.298 & 0.174 & \underline{0.295} & 0.205 & 0.324 & 0.253 & 0.376 & 0.190 & 0.309 & 0.195 & 0.316 & \underline{0.172} & 0.316 & \textbf{0.157} & \textbf{0.293} & 0.226 & 0.344 & 0.182 & 0.303 \\
    & 336 & 0.339 & 0.418 & 0.342 & 0.421 & 0.378 & 0.456 & 0.390 & 0.471 & 0.368 & 0.434 & 0.342 & 0.425 & \textbf{0.272} & \textbf{0.407} & \underline{0.305} & \underline{0.414} & 0.367 & 0.448 & 0.348 & 0.428 \\
    & 720 & 0.867 & 0.701 & 0.841 & 0.689 & 1.205 & 0.810 & 1.080 & 0.789 & 1.045 & 0.755 & 0.998 & 0.756 & \underline{0.714} & \underline{0.658} & \textbf{0.643} & \textbf{0.601} & 0.964 & 0.746 & 1.025 & 0.774 \\
    \cline{2-22}
    & Avg & 0.366 & 0.404 & 0.360 & \underline{0.402} & 0.471 & 0.452 & 0.458 & 0.469 & 0.423 & 0.427 & 0.405 & 0.426 & \underline{0.315} & 0.404 & \textbf{0.297} & \textbf{0.378} & 0.416 & 0.443 & 0.410 & 0.427 \\
    \hline
    \multirow{5}{*}{\rotatebox{90}{Solar}}
    & 96  & \underline{0.173} & \textbf{0.197} & 0.179 & \underline{0.212} & \textbf{0.167} & 0.220 & 0.179 & 0.248 & 0.211 & 0.254 & 0.224 & 0.278 & 0.188 & 0.252 & 0.289 & 0.377 & 0.219 & 0.314 & 0.258 & 0.371 \\
    & 192 & \underline{0.193} & \textbf{0.216} & 0.200 & \underline{0.227} & \textbf{0.187} & 0.249 & 0.197 & 0.266 & 0.230 & 0.264 & 0.253 & 0.298 & 0.215 & 0.280 & 0.319 & 0.397 & 0.231 & 0.322 & 0.608 & 0.606 \\
    & 336 & \textbf{0.196} & \textbf{0.224} & 0.203 & \underline{0.226} & \underline{0.200} & 0.258 & 0.202 & 0.263 & 0.246 & 0.272 & 0.273 & 0.306 & 0.222 & 0.267 & 0.352 & 0.415 & 0.246 & 0.337 & 0.758 & 0.705 \\
    & 720 & \underline{0.212} & \textbf{0.219} & \textbf{0.209} & \underline{0.236} & 0.215 & 0.250 & \textbf{0.209} & 0.271 & 0.252 & 0.274 & 0.272 & 0.308 & 0.226 & 0.264 & 0.356 & 0.412 & 0.280 & 0.363 & 0.789 & 0.779 \\
    \cline{2-22}
    & Avg & \underline{0.194} & \textbf{0.214} & 0.198 & \underline{0.225} & \textbf{0.192} & 0.244 & 0.197 & 0.262 & 0.235 & 0.266 & 0.256 & 0.298 & 0.213 & 0.266 & 0.329 & 0.400 & 0.244 & 0.334 & 0.603 & 0.615 \\
    \hline
    \multirow{5}{*}{\rotatebox{90}{ILI}}
    & 24  & \textbf{1.188} & \textbf{0.638} & 1.665 & 0.803 & 1.693 & 0.872 & 2.743 & 1.150 & 1.734 & 0.864 & \underline{1.319} & \underline{0.754} & 2.684 & 1.112 & 2.215 & 1.081 & 2.317 & 0.934 & 2.527 & 1.020 \\
    & 36 & \textbf{1.226} & \textbf{0.653} & 2.200 & 0.890 & 2.002 & 0.942 & 2.887 & 1.183 & 1.771 & \underline{0.849} & \underline{1.579} & 0.870 & 2.507 & 1.013 & 1.963 & 0.963 & 1.972 & 0.920 & 2.615 & 1.007 \\
    & 48 & \textbf{1.254} & \textbf{0.686} & 1.875 & 0.821 & 2.086 & 0.937 & 2.998 & 1.206 & 1.777 & 0.864 & \underline{1.553} & \underline{0.815} & 2.423 & 1.012 & 2.130 & 1.024 & 2.238 & 0.940 & 2.359 & 0.972 \\
    & 60 & \textbf{1.455} & \textbf{0.773} & 1.923 & 0.853 & 2.102 & 0.946 & 3.160 & 1.234 & 1.929 & 0.919 & \underline{1.470} & \underline{0.788} & 2.653 & 1.085 & 2.368 & 1.096 & 2.027 & 0.928 & 2.487 & 1.016 \\
    \cline{2-22}
    & Avg & \textbf{1.281} & \textbf{0.688} & 1.916 & 0.842 & 1.971 & 0.924 & 2.947 & 1.193 & 1.803 & 0.874 & \underline{1.480} & \underline{0.807} & 2.567 & 1.056 & 2.169 & 1.041 & 2.139 & 0.931 & 2.497 & 1.004 \\
    \hline
    \multicolumn{2}{c|}{$1^{st}$ Count} & \textbf{31} & \textbf{39} & \underline{7} & \underline{8} & 5 & 2 & 5 & 0 & 0 & 0 & 0 & 0 & 1 & 1 & 4 & 3 & 0 & 0 & 0 & 0 \\
    \hline
    \end{tabular}
  \caption{Full long-term forecasting results under hyperparameter searching.
  The best model is \textbf{boldface} and the second best is \underline{underlined}.}
    \label{tab:full-experiments-search}
\end{table*}

\begin{table*}[th!]
  \scriptsize
  \centering
  \setlength\tabcolsep{3pt}
  \begin{tabular}{c|c|cc|cc|cc|cc|cc|cc|cc|cc|cc|cc}
    \hline
    \multicolumn{2}{c}{\multirow{2}{*}{Models}}                                 &
    \multicolumn{2}{|c}{\textbf{xPatch}}    & \multicolumn{2}{|c}{PDF}         &
    \multicolumn{2}{|c}{FITS}               & \multicolumn{2}{|c}{iTransformer} & 
    \multicolumn{2}{|c}{TimeMixer}          & \multicolumn{2}{|c}{PatchTST}     & 
    \multicolumn{2}{|c}{MICN}               & \multicolumn{2}{|c}{DLinear}      &
    \multicolumn{2}{|c}{TimesNet}           & \multicolumn{2}{|c}{FEDformer}    \\
    \multicolumn{2}{c}{\multirow{2}{*}{}}                                       &
    \multicolumn{2}{|c}{\textbf{(ours)}}    & \multicolumn{2}{|c}{(2024)}       & 
    \multicolumn{2}{|c}{(2024)}             & \multicolumn{2}{|c}{(2024)}       & 
    \multicolumn{2}{|c}{(2024)}             & \multicolumn{2}{|c}{(2023)}       & 
    \multicolumn{2}{|c}{(2023)}             & \multicolumn{2}{|c}{(2023)}       & 
    \multicolumn{2}{|c}{(2023)}             & \multicolumn{2}{|c}{(2022)}       \\
    \hline
    \multicolumn{2}{c|}{Metric}
    & MSE & MAE & MSE & MAE & MSE & MAE & MSE & MAE & MSE & MAE & MSE & MAE & MSE & MAE & MSE & MAE & MSE & MAE & MSE & MAE \\
    \hline
    \multirow{5}{*}{\rotatebox{90}{ETTh1}}
    & 96  & \underline{0.363} & \textbf{0.390} & \textbf{0.360} & \underline{0.391} & 0.376 & 0.396 & 0.386 & 0.405 & 0.372 & 0.401 & 0.377 & 0.397 & 0.378 & 0.412 & 0.379 & 0.403 & 0.389 & 0.412 & 0.379 & 0.419 \\
    & 192 & 0.404 & \textbf{0.414} & \textbf{0.392} & \textbf{0.414} & \underline{0.400} & \underline{0.418} & 0.424 & 0.440 & 0.413 & 0.430 & 0.409 & 0.425 & \underline{0.400} & 0.430 & 0.408 & 0.419 & 0.440 & 0.443 & 0.420 & 0.444 \\
    & 336 & 0.432 & \textbf{0.432} & \textbf{0.418} & \underline{0.435} & \underline{0.419} & \underline{0.435} & 0.449 & 0.460 & 0.438 & 0.450 & 0.431 & 0.444 & 0.428 & 0.447 & 0.440 & 0.440 & 0.523 & 0.487 & 0.458 & 0.466 \\
    & 720 & \underline{0.451} & \textbf{0.457} & 0.456 & 0.462 & \textbf{0.435} & \underline{0.458} & 0.495 & 0.487 & 0.486 & 0.484 & 0.457 & 0.477 & 0.474 & 0.499 & 0.471 & 0.493 & 0.521 & 0.495 & 0.474 & 0.488 \\
    \cline{2-22}
    & Avg & 0.413 & \textbf{0.423} & \textbf{0.407} & \underline{0.426} & \underline{0.408} & 0.427 & 0.439 & 0.448 & 0.427 & 0.441 & 0.419 & 0.436 & 0.420 & 0.447 & 0.425 & 0.439 & 0.468 & 0.459 & 0.433 & 0.454 \\
    \hline
    \multirow{5}{*}{\rotatebox{90}{ETTh2}}
    & 96  & \textbf{0.274} & \textbf{0.333} & \underline{0.276} & 0.341 & 0.277 & 0.345 & 0.297 & 0.348 & 0.281 & 0.351 & \textbf{0.274} & \underline{0.337} & 0.313 & 0.372 & 0.300 & 0.364 & 0.334 & 0.370 & 0.337 & 0.380 \\
    & 192 & \underline{0.336} & \textbf{0.374} & 0.339 & 0.382 & \textbf{0.331} & \underline{0.379} & 0.372 & 0.403 & 0.349 & 0.387 & 0.348 & 0.384 & 0.419 & 0.439 & 0.387 & 0.423 & 0.404 & 0.413 & 0.415 & 0.428 \\
    & 336 & \underline{0.366} & \underline{0.400} & 0.374 & 0.406 & \textbf{0.350} & \textbf{0.396} & 0.388 & 0.417 & \underline{0.366} & 0.413 & 0.377 & 0.416 & 0.474 & 0.475 & 0.490 & 0.487 & 0.389 & 0.435 & 0.389 & 0.457 \\
    & 720 & \underline{0.388} & \textbf{0.425} & 0.398 & \underline{0.433} & \textbf{0.382} & \textbf{0.425} & 0.424 & 0.444 & 0.401 & 0.436 & 0.406 & 0.441 & 0.723 & 0.600 & 0.704 & 0.597 & 0.434 & 0.448 & 0.483 & 0.488 \\
    \cline{2-22}
    & Avg & \underline{0.341} & \textbf{0.383} & 0.347 & 0.391 & \textbf{0.335} & \underline{0.386} & 0.370 & 0.403 & 0.349 & 0.397 & 0.351 & 0.395 & 0.482 & 0.472 & 0.470 & 0.468 & 0.390 & 0.417 & 0.406 & 0.438 \\
    \hline
    \multirow{5}{*}{\rotatebox{90}{ETTm1}}
    & 96  & \underline{0.287} & \textbf{0.330} & \textbf{0.286} & \underline{0.340} & 0.303 & 0.345 & 0.300 & 0.353 & 0.293 & 0.345 & 0.289 & 0.343 & 0.303 & 0.349 & 0.300 & 0.345 & 0.340 & 0.378 & 0.463 & 0.463 \\
    & 192 & \underline{0.328} & \textbf{0.356} & \textbf{0.321} & \underline{0.364} & 0.337 & 0.365 & 0.341 & 0.380 & 0.335 & 0.372 & 0.329 & 0.368 & 0.336 & 0.369 & 0.336 & 0.366 & 0.392 & 0.404 & 0.575 & 0.516 \\
    & 336 & 0.363 & \textbf{0.379} & \textbf{0.354} & \underline{0.383} & 0.368 & 0.384 & 0.374 & 0.396 & 0.368 & 0.386 & \underline{0.362} & 0.390 & 0.370 & 0.391 & 0.367 & 0.386 & 0.423 & 0.426 & 0.618 & 0.544 \\
    & 720 & 0.426 & 0.417 & \textbf{0.408} & \underline{0.415} & 0.420 & \textbf{0.413} & 0.429 & 0.430 & 0.426 & 0.417 & 0.416 & 0.423 & \underline{0.410} & 0.421 & 0.419 & 0.416 & 0.475 & 0.453 & 0.612 & 0.551 \\
    \cline{2-22}
    & Avg & 0.351 & \textbf{0.371} & \textbf{0.342} & \underline{0.376} & 0.357 & 0.377 & 0.361 & 0.390 & 0.356 & 0.380 & \underline{0.349} & 0.381 & 0.355 & 0.383 & 0.356 & 0.378 & 0.408 & 0.415 & 0.567 & 0.519 \\
    \hline
    \multirow{5}{*}{\rotatebox{90}{ETTm2}}
    & 96  & \textbf{0.157} & \textbf{0.243} & \underline{0.163} & \underline{0.251} & 0.165 & 0.254 & 0.175 & 0.266 & 0.165 & 0.256 & 0.165 & 0.255 & 0.173 & 0.271 & 0.164 & 0.255 & 0.189 & 0.265 & 0.216 & 0.309 \\
    & 192 & \textbf{0.216} & \textbf{0.285} & \underline{0.219} & \underline{0.290} & \underline{0.219} & 0.291 & 0.242 & 0.312 & 0.225 & 0.298 & 0.221 & 0.293 & 0.232 & 0.313 & 0.224 & 0.304 & 0.254 & 0.310 & 0.297 & 0.360 \\
    & 336 & \underline{0.271} & \textbf{0.323} & \textbf{0.269} & 0.330 & \underline{0.272} & \underline{0.326} & 0.282 & 0.337 & 0.277 & 0.332 & 0.276 & 0.327 & 0.303 & 0.367 & 0.277 & 0.337 & 0.313 & 0.345 & 0.366 & 0.400 \\
    & 720 & \underline{0.358} & \textbf{0.377} & \textbf{0.349} & 0.382 & 0.359 & \underline{0.381} & 0.375 & 0.394 & 0.360 & 0.387 & 0.362 & 0.381 & 0.467 & 0.477 & 0.371 & 0.401 & 0.413 & 0.402 & 0.459 & 0.450 \\
    \cline{2-22}
    & Avg & \underline{0.251} & \textbf{0.307} & \textbf{0.250} & \underline{0.313} & 0.254 & \underline{0.313} & 0.269 & 0.327 & 0.257 & 0.318 & 0.256 & 0.314 & 0.294 & 0.357 & 0.259 & 0.324 & 0.292 & 0.331 & 0.335 & 0.380 \\
    \hline
    \multirow{5}{*}{\rotatebox{90}{Weather}}
    & 96  & \textbf{0.144} & \textbf{0.184} & \underline{0.147} & \underline{0.196} & 0.172 & 0.225 & 0.157 & 0.207 & \underline{0.147} & 0.198 & 0.149 & \underline{0.196} & 0.172 & 0.232 & 0.170 & 0.230 & 0.168 & 0.214 & 0.175 & 0.242 \\
    & 192 & \textbf{0.188} & \textbf{0.227} & 0.193 & 0.240 & 0.215 & 0.261 & 0.200 & 0.248 & 0.192 & 0.243 & \underline{0.191} & \underline{0.239} & 0.214 & 0.270 & 0.216 & 0.275 & 0.219 & 0.262 & 0.274 & 0.344 \\
    & 336 & \textbf{0.236} & \textbf{0.266} & 0.245 & 0.280 & 0.261 & 0.295 & 0.252 & 0.287 & 0.247 & 0.284 & \underline{0.242} & \underline{0.279} & 0.259 & 0.309 & 0.258 & 0.307 & 0.278 & 0.302 & 0.331 & 0.374 \\
    & 720 & \textbf{0.309} & \textbf{0.319} & 0.323 & 0.334 & 0.326 & 0.341 & 0.320 & 0.336 & 0.318 & \underline{0.330} & \underline{0.312} & \underline{0.330} & \textbf{0.309} & 0.343 & 0.323 & 0.362 & 0.353 & 0.351 & 0.423 & 0.418 \\
    \cline{2-22}
    & Avg & \textbf{0.219} & \textbf{0.249} & 0.227 & 0.263 & 0.244 & 0.281 & 0.232 & 0.270 & 0.226 & 0.264 & \underline{0.224} & \underline{0.261} & 0.239 & 0.289 & 0.242 & 0.293 & 0.255 & 0.282 & 0.301 & 0.345 \\
    \hline
    \multirow{5}{*}{\rotatebox{90}{Traffic}}
    & 96  & \textbf{0.359} & \textbf{0.233} & 0.368 & \underline{0.252} & 0.400 & 0.280 & \underline{0.363} & 0.265 & 0.369 & 0.257 & 0.370 & 0.262 & 0.517 & 0.313 & 0.395 & 0.275 & 0.595 & 0.312 & 0.593 & 0.365 \\
    & 192 & \textbf{0.375} & \textbf{0.239} & \underline{0.382} & \underline{0.261} & 0.412 & 0.288 & 0.384 & 0.273 & 0.400 & 0.272 & 0.386 & 0.269 & 0.526 & 0.302 & 0.407 & 0.280 & 0.613 & 0.322 & 0.614 & 0.381 \\
    & 336 & \textbf{0.388} & \textbf{0.244} & \underline{0.393} & \underline{0.268} & 0.433 & 0.308 & 0.396 & 0.277 & 0.407 & 0.272 & 0.396 & 0.275 & 0.545 & 0.307 & 0.417 & 0.286 & 0.626 & 0.332 & 0.627 & 0.389 \\
    & 720 & \textbf{0.429} & \textbf{0.264} & 0.438 & 0.297 & 0.478 & 0.339 & 0.445 & 0.308 & 0.461 & 0.316 & \underline{0.435} & \underline{0.295} & 0.569 & 0.328 & 0.454 & 0.308 & 0.635 & 0.340 & 0.646 & 0.394 \\
    \cline{2-22}
    & Avg & \textbf{0.388} & \textbf{0.245} & \underline{0.395} & \underline{0.270} & 0.429 & 0.302 & 0.397 & 0.281 & 0.409 & 0.279 & 0.397 & 0.275 & 0.539 & 0.313 & 0.418 & 0.287 & 0.617 & 0.327 & 0.620 & 0.382 \\
    \hline
    \multirow{5}{*}{\rotatebox{90}{Electricity}}
    & 96  & \textbf{0.126} & \textbf{0.217} & \underline{0.128} & \underline{0.222} & 0.139 & 0.237 & 0.134 & 0.230 & 0.153 & 0.256 & 0.143 & 0.247 & 0.158 & 0.266 & 0.140 & 0.237 & 0.169 & 0.271 & 0.191 & 0.305 \\
    & 192 & \textbf{0.143} & \textbf{0.233} & \underline{0.147} & \underline{0.242} & 0.154 & 0.250 & 0.154 & 0.250 & 0.168 & 0.269 & 0.158 & 0.260 & 0.175 & 0.287 & 0.154 & 0.251 & 0.180 & 0.280 & 0.203 & 0.316 \\
    & 336 & \textbf{0.159} & \textbf{0.250} & \underline{0.165} & \underline{0.260} & 0.170 & 0.268 & 0.169 & 0.265 & 0.189 & 0.291 & 0.168 & 0.267 & 0.184 & 0.296 & 0.169 & 0.268 & 0.204 & 0.304 & 0.221 & 0.333 \\
    & 720 & \textbf{0.189} & \textbf{0.279} & 0.199 & 0.289 & 0.212 & 0.304 & \underline{0.194} & \underline{0.288} & 0.228 & 0.320 & 0.214 & 0.307 & 0.200 & 0.310 & 0.204 & 0.301 & 0.245 & 0.333 & 0.259 & 0.364 \\
    \cline{2-22}
    & Avg & \textbf{0.154} & \textbf{0.245} & \underline{0.160} & \underline{0.253} & 0.169 & 0.265 & 0.163 & 0.258 & 0.185 & 0.284 & 0.171 & 0.270 & 0.179 & 0.290 & 0.167 & 0.264 & 0.200 & 0.297 & 0.219 & 0.330 \\
    \hline
    \multirow{5}{*}{\rotatebox{90}{Exchange}}
    & 96  & \underline{0.080} & \textbf{0.197} & 0.083 & 0.200 & 0.082 & \underline{0.199} & 0.086 & 0.205 & 0.084 & 0.207 & \textbf{0.079} & 0.200 & \textbf{0.079} & 0.203 & \underline{0.080} & 0.202 & 0.112 & 0.242 & \textbf{0.079} & 0.203 \\
    & 192 & 0.172 & 0.293 & 0.172 & 0.294 & 0.173 & 0.295 & 0.177 & 0.299 & 0.178 & 0.300 & \underline{0.159} & \textbf{0.289} & \textbf{0.158} & 0.299 & 0.182 & 0.321 & 0.209 & 0.334 & \underline{0.159} & \underline{0.292} \\
    & 336 & 0.336 & 0.418 & 0.323 & 0.411 & 0.317 & 0.406 & 0.331 & 0.417 & 0.376 & 0.451 & \textbf{0.297} & \textbf{0.399} & \underline{0.300} & \underline{0.403} & 0.327 & 0.434 & 0.358 & 0.435 & 0.301 & 0.407 \\
    & 720 & 0.855 & 0.696 & 0.820 & 0.682 & 0.825 & 0.684 & 0.846 & 0.693 & 0.884 & 0.707 & 0.751 & \underline{0.650} & \underline{0.745} & 0.665 & \textbf{0.578} & \textbf{0.605} & 0.944 & 0.736 & 0.835 & 0.706 \\
    \cline{2-22}
    & Avg & 0.361 & 0.401 & 0.350 & 0.397 & 0.349 & 0.396 & 0.360 & 0.404 & 0.381 & 0.416 & 0.322 & \textbf{0.385} & \underline{0.321} & 0.393 & \textbf{0.292} & \underline{0.391} & 0.406 & 0.437 & 0.344 & 0.402 \\
    \hline
    \multirow{5}{*}{\rotatebox{90}{Solar}}
    & 96  & \underline{0.176} & \textbf{0.198} & 0.181 & 0.247 & 0.208 & 0.255 & 0.190 & 0.244 & 0.179 & \underline{0.232} & \textbf{0.170} & 0.234 & 0.190 & 0.250 & 0.199 & 0.265 & 0.198 & 0.270 & 0.472 & 0.559 \\
    & 192 & \textbf{0.190} & \textbf{0.209} & 0.200 & 0.259 & 0.229 & 0.267 & \underline{0.193} & \underline{0.257} & 0.201 & 0.259 & 0.204 & 0.302 & 0.226 & 0.284 & 0.220 & 0.282 & 0.206 & 0.276 & 0.415 & 0.477 \\
    & 336 & \underline{0.195} & \textbf{0.216} & 0.208 & 0.269 & 0.241 & 0.273 & 0.203 & 0.266 & \textbf{0.190} & \underline{0.256} & 0.212 & 0.293 & 0.259 & 0.308 & 0.234 & 0.295 & 0.208 & 0.284 & 1.008 & 0.839 \\
    & 720 & \underline{0.206} & \textbf{0.223} & 0.212 & 0.275 & 0.248 & 0.277 & 0.223 & 0.281 & \textbf{0.203} & \underline{0.261} & 0.215 & 0.307 & 0.341 & 0.365 & 0.243 & 0.301 & 0.232 & 0.294 & 0.642 & 0.620 \\
    \cline{2-22}
    & Avg & \textbf{0.192} & \textbf{0.212} & 0.200 & 0.263 & 0.232 & 0.268 & 0.202 & 0.262 & \underline{0.193} & \underline{0.252} & 0.200 & 0.284 & 0.254 & 0.302 & 0.224 & 0.286 & 0.211 & 0.281 & 0.634 & 0.624 \\
    \hline
    \multirow{5}{*}{\rotatebox{90}{ILI}}
    & 24 & \textbf{1.642} & \textbf{0.771} & 1.801 & 0.874 & 2.182 & 1.002 & \underline{1.783} & 0.846 & 1.804 & \underline{0.820} & 1.932 & 0.872 & 2.279 & 1.020 & 2.208 & 1.031 & 2.131 & 0.958 & 2.403 & 1.020 \\
    & 36 & \textbf{1.647} & \textbf{0.773} & \underline{1.743} & 0.867 & 2.241 & 1.029 & 1.746 & \underline{0.860} & 1.891 & 0.926 & 1.869 & 0.866 & 2.451 & 1.085 & 2.032 & 0.981 & 2.612 & 0.974 & 2.410 & 1.005 \\
    & 48 & \textbf{1.601} & \textbf{0.773} & 1.843 & 0.926 & 2.272 & 1.036 & \underline{1.716} & 0.898 & 1.752 & \underline{0.866} & 1.891 & 0.883 & 2.440 & 1.077 & 2.209 & 1.063 & 1.916 & 0.897 & 2.591 & 1.033 \\
    & 60 & \textbf{1.643} & \textbf{0.814} & 1.845 & 0.925 & 2.642 & 1.142 & 2.183 & 0.963 & \underline{1.831} & 0.930 & 1.914 & \underline{0.896} & 2.303 & 1.012 & 2.292 & 1.086 & 1.995 & 0.905 & 2.540 & 1.070 \\
    \cline{2-22}
    & Avg & \textbf{1.633} & \textbf{0.783} & \underline{1.808} & 0.898 & 2.334 & 1.052 & 1.857 & 0.892 & 1.820 & 0.886 & 1.902 & \underline{0.879} & 2.368 & 1.049 & 2.185 & 1.040 & 2.164 & 0.934 & 2.486 & 1.032 \\
    \hline
    \multicolumn{2}{c|}{$1^{st}$ Count} & \textbf{25} & \textbf{44} & \underline{12} & 1 & 5 & \underline{3} & 0 & 0 & 2 & 0 & 4 & \underline{3} & 3 & 0 & 2 & 1 & 0 & 0 & 1 & 0 \\
    \hline
    \end{tabular}
  \caption{Fair long-term forecasting results under hyperparameter searching without the “drop-last” trick.
  The best model is \textbf{boldface} and the second best is \underline{underlined}.}
    \label{tab:fair-experiments}
\end{table*}

\section{Visualizations}
\begin{figure*}[th]
     \centering
     \begin{subfigure}[b]{0.35\textwidth}
         \centering
         \includegraphics[width=1\columnwidth]{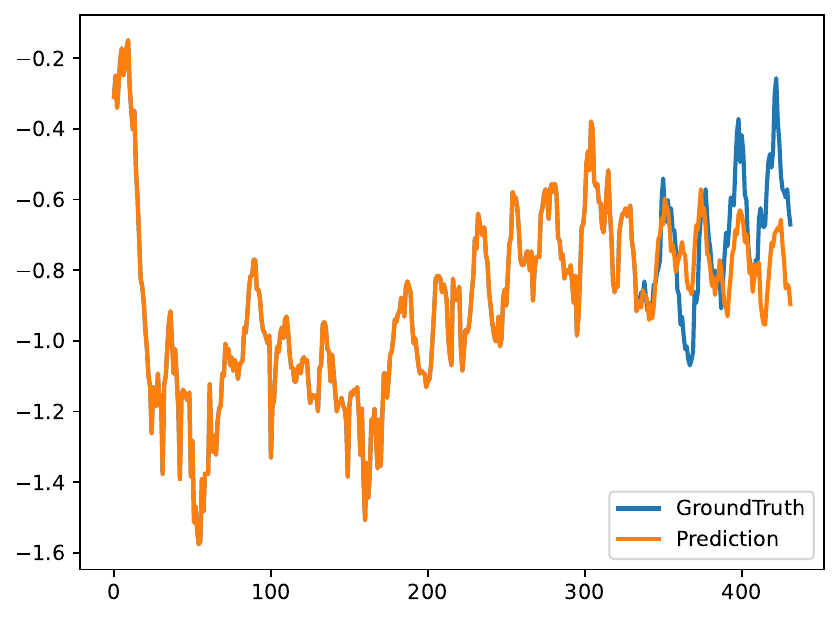}
         \caption{xPatch}
     \end{subfigure}
     \begin{subfigure}[b]{0.35\textwidth}
         \centering
         \includegraphics[width=1\columnwidth]{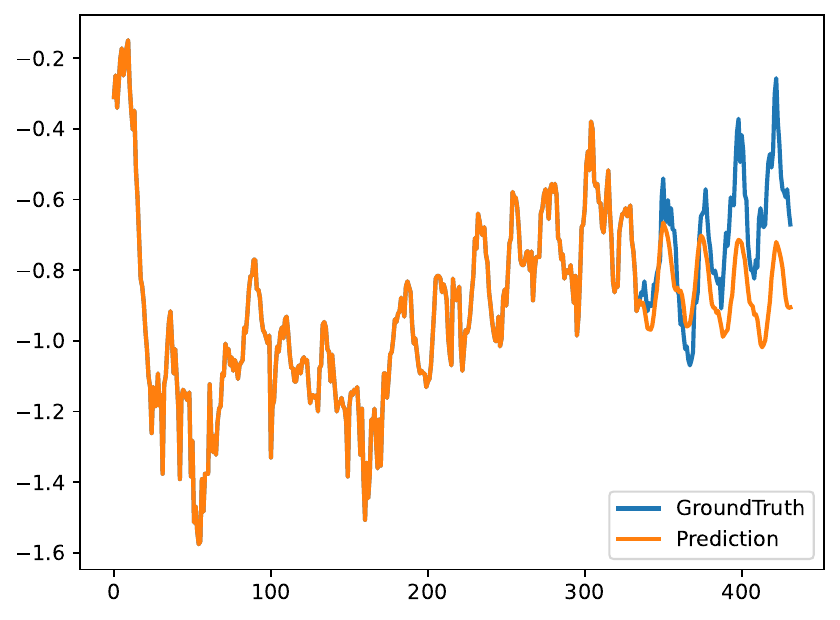}
         \caption{CARD}
     \end{subfigure}
     \begin{subfigure}[b]{0.35\textwidth}
         \centering
         \includegraphics[width=1\columnwidth]{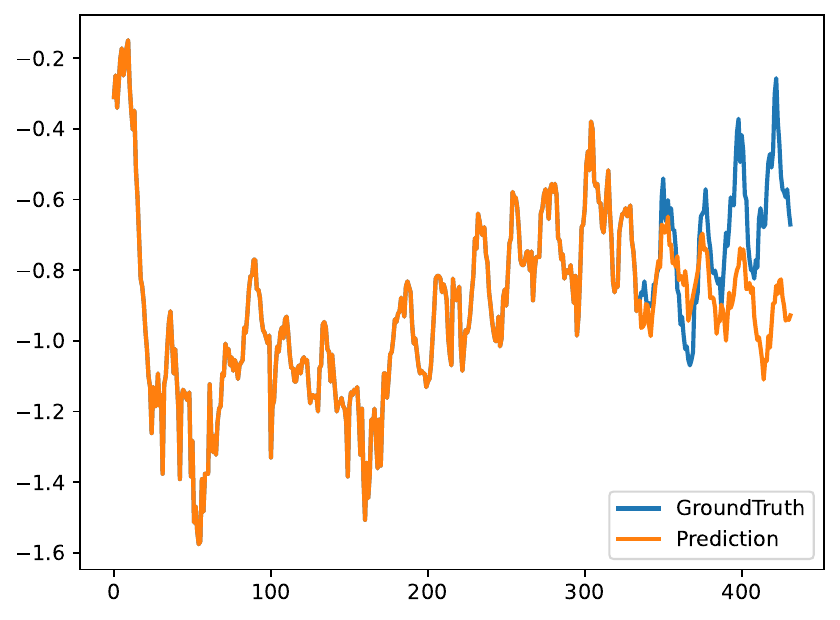}
         \caption{iTransformer}
     \end{subfigure}
     \begin{subfigure}[b]{0.35\textwidth}
         \centering
         \includegraphics[width=1\columnwidth]{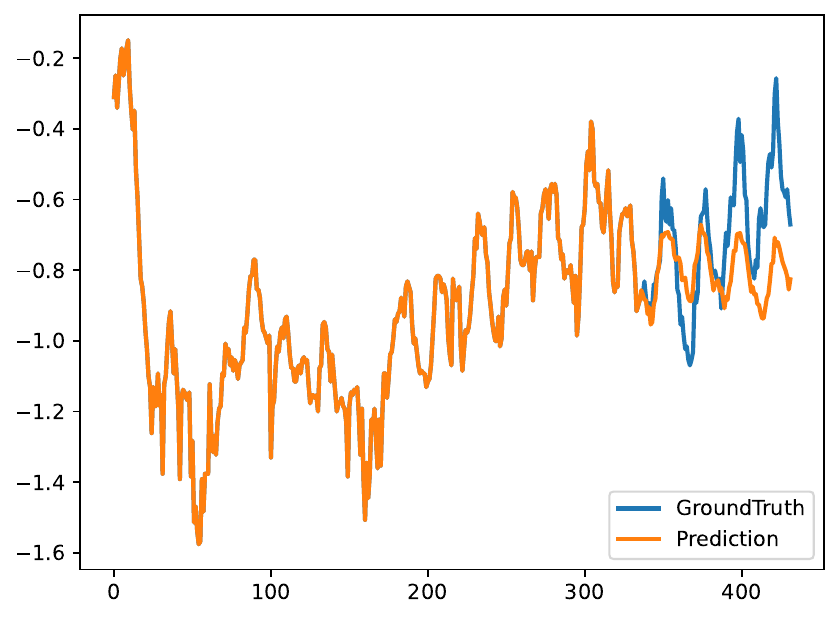}
         \caption{PatchTST}
     \end{subfigure}
        \caption{Sample prediction graph of the next T = 96 points with lookback window L = 336 from the ETTh1 dataset.}
        \label{fig:qual1}
\end{figure*}

\begin{figure*}[th]
     \centering
     \begin{subfigure}[b]{0.35\textwidth}
         \centering
         \includegraphics[width=1\columnwidth]{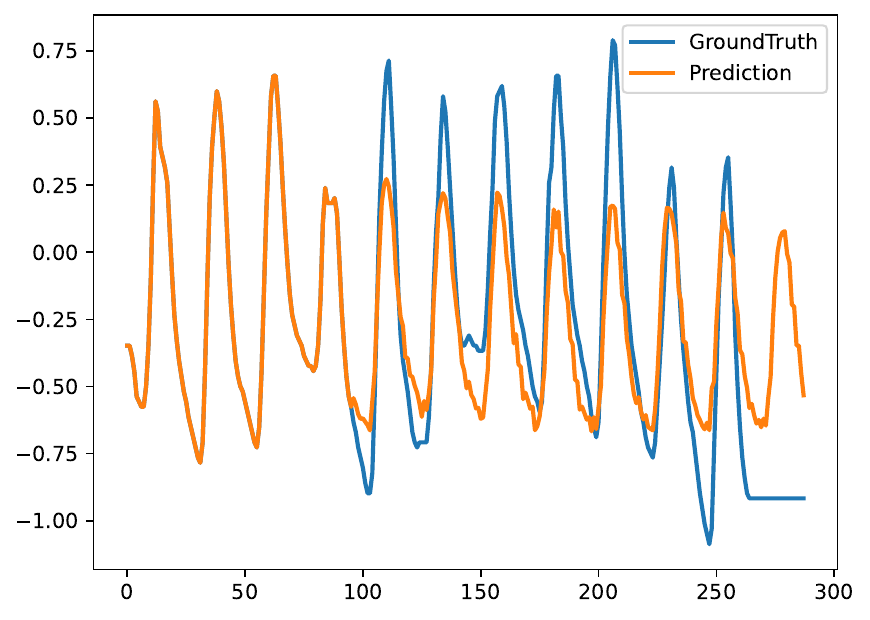}
         \caption{xPatch}
     \end{subfigure}
     \begin{subfigure}[b]{0.35\textwidth}
         \centering
         \includegraphics[width=1\columnwidth]{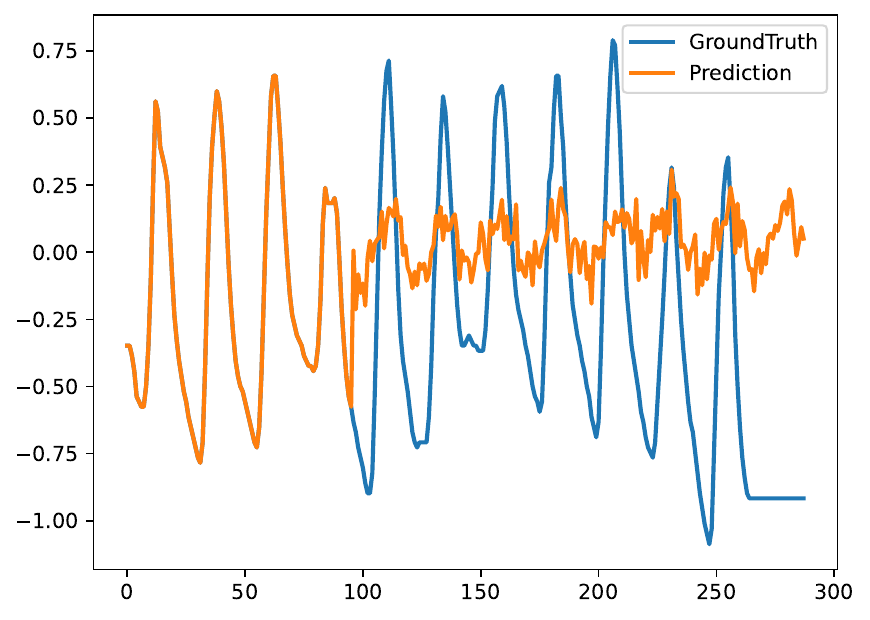}
         \caption{CARD}
     \end{subfigure}
     \begin{subfigure}[b]{0.35\textwidth}
         \centering
         \includegraphics[width=1\columnwidth]{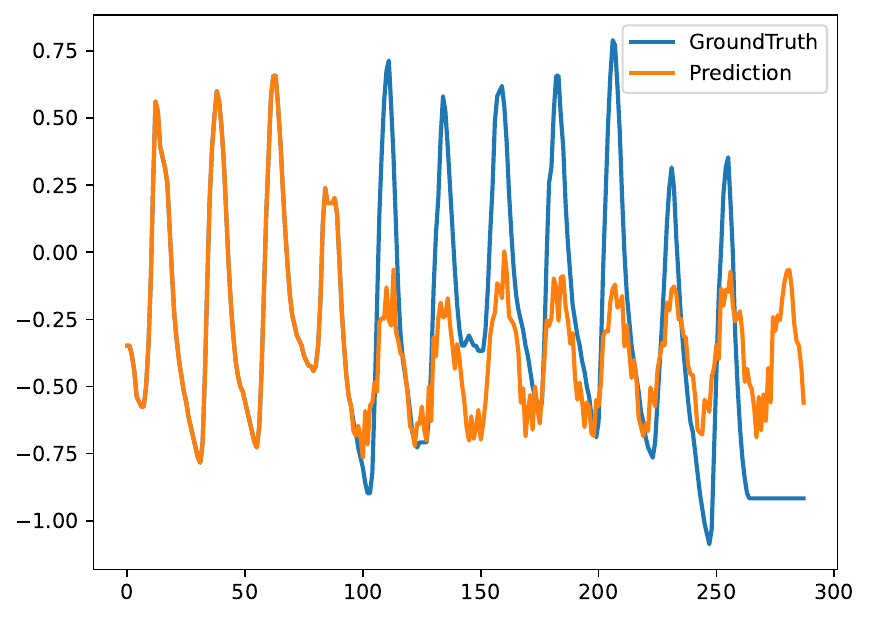}
         \caption{iTransformer}
     \end{subfigure}
     \begin{subfigure}[b]{0.35\textwidth}
         \centering
         \includegraphics[width=1\columnwidth]{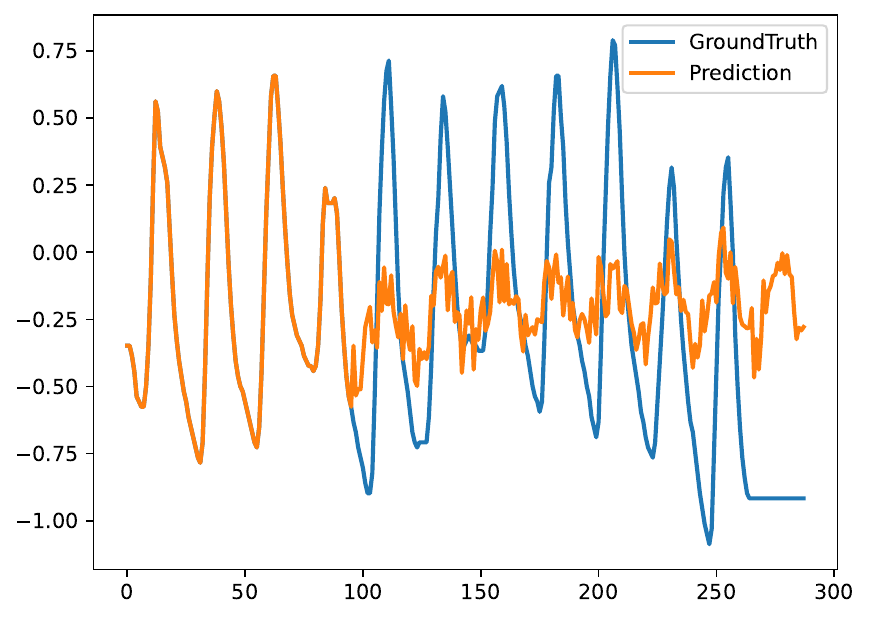}
         \caption{PatchTST}
     \end{subfigure}
        \caption{Sample prediction graph of the next T = 192 points with lookback window L = 96 from the ETTh2 dataset.}
        \label{fig:qual2}
\end{figure*}

\begin{figure*}[th]
     \centering
     \begin{subfigure}[b]{0.35\textwidth}
         \centering
         \includegraphics[width=1\columnwidth]{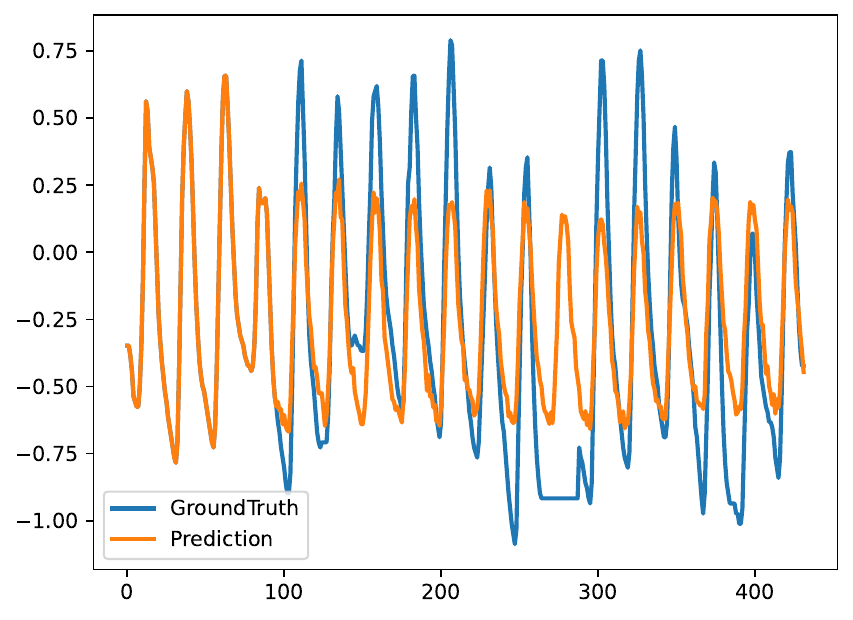}
         \caption{xPatch}
     \end{subfigure}
     \begin{subfigure}[b]{0.35\textwidth}
         \centering
         \includegraphics[width=1\columnwidth]{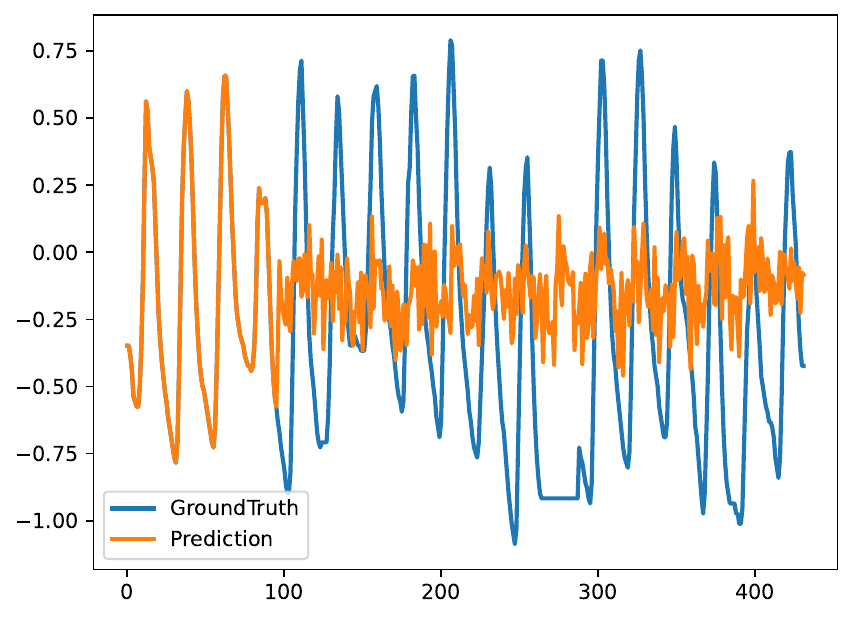}
         \caption{CARD}
     \end{subfigure}
     \begin{subfigure}[b]{0.35\textwidth}
         \centering
         \includegraphics[width=1\columnwidth]{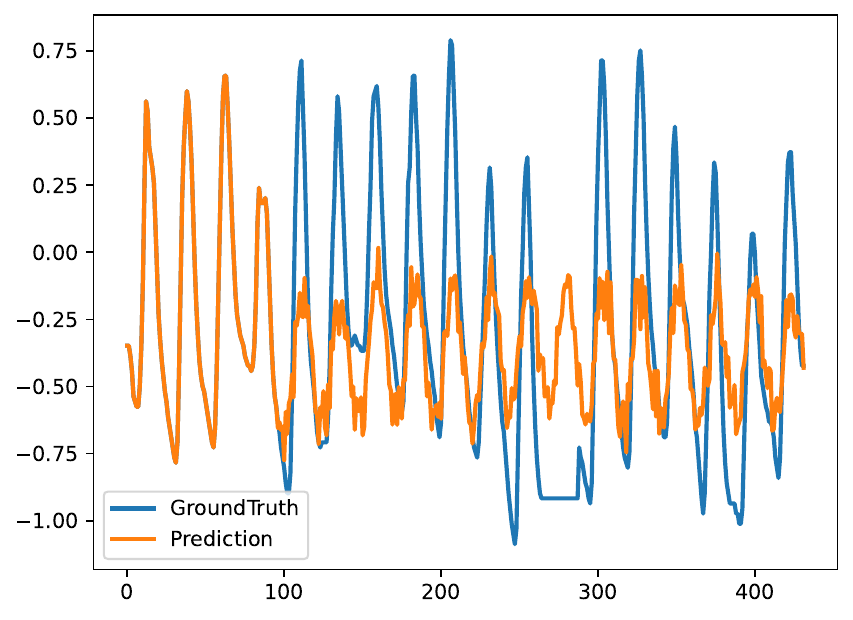}
         \caption{iTransformer}
     \end{subfigure}
     \begin{subfigure}[b]{0.35\textwidth}
         \centering
         \includegraphics[width=1\columnwidth]{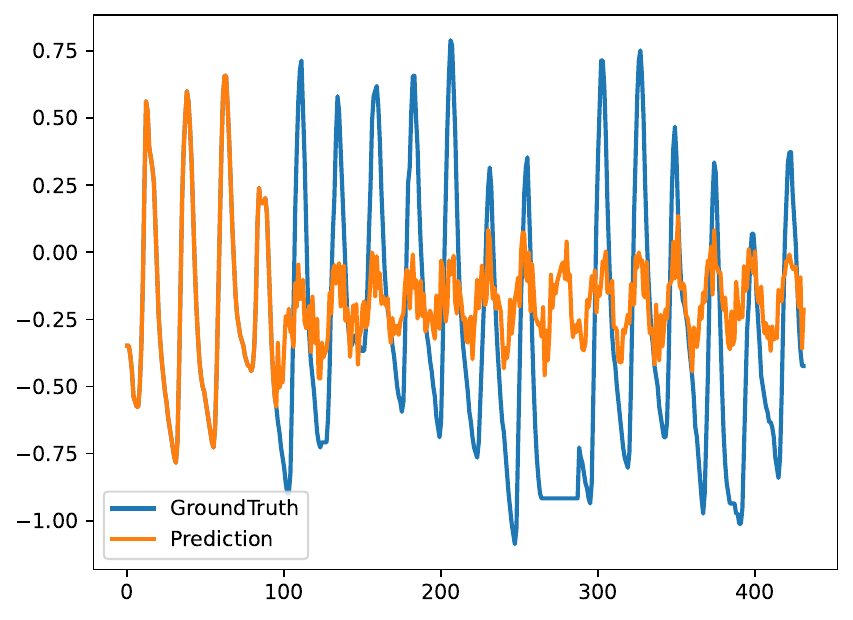}
         \caption{PatchTST}
     \end{subfigure}
        \caption{Sample prediction graph of the next T = 336 points with lookback window L = 96 from the ETTh2 dataset.}
        \label{fig:qual3}
\end{figure*}

\begin{figure*}[th]
     \centering
     \begin{subfigure}[b]{0.35\textwidth}
         \centering
         \includegraphics[width=1\columnwidth]{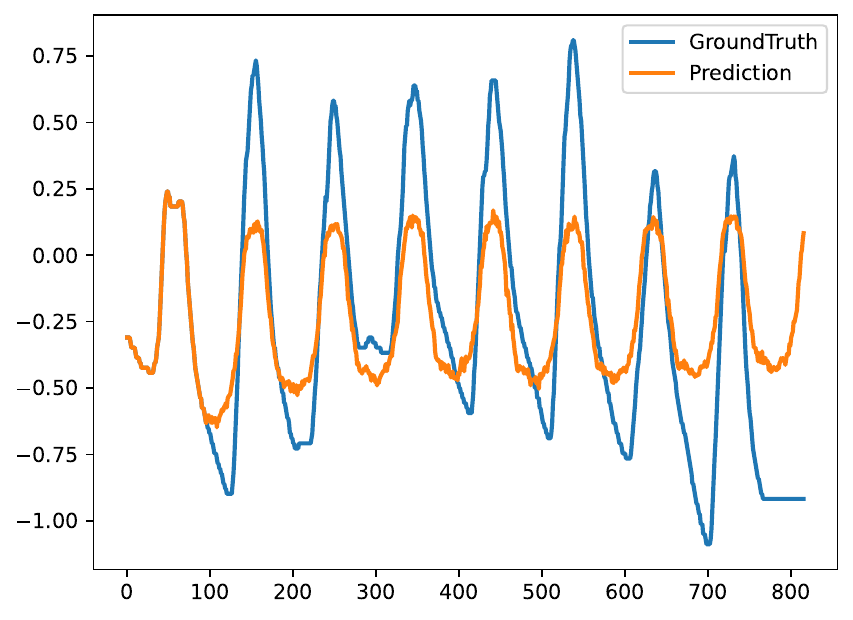}
         \caption{xPatch}
     \end{subfigure}
     \begin{subfigure}[b]{0.35\textwidth}
         \centering
         \includegraphics[width=1\columnwidth]{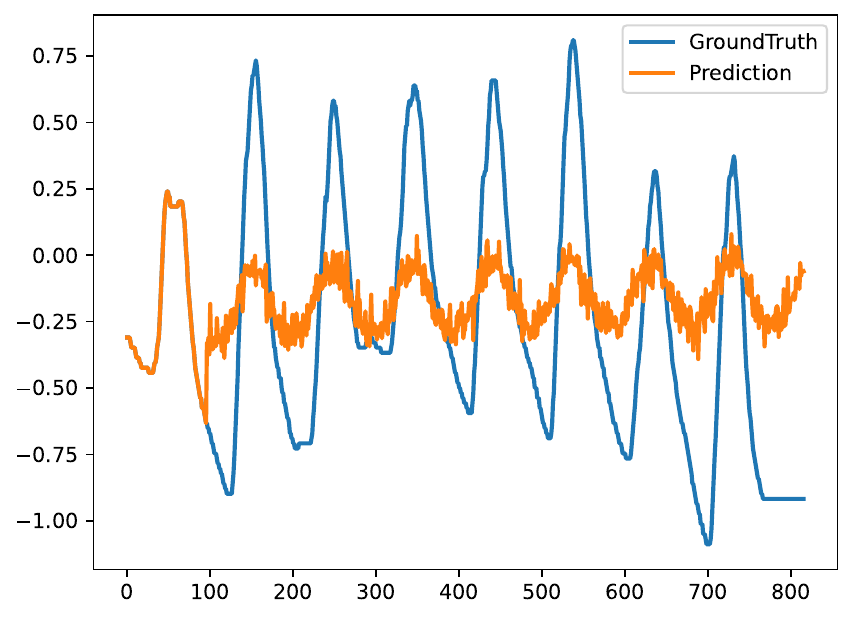}
         \caption{CARD}
     \end{subfigure}
     \begin{subfigure}[b]{0.35\textwidth}
         \centering
         \includegraphics[width=1\columnwidth]{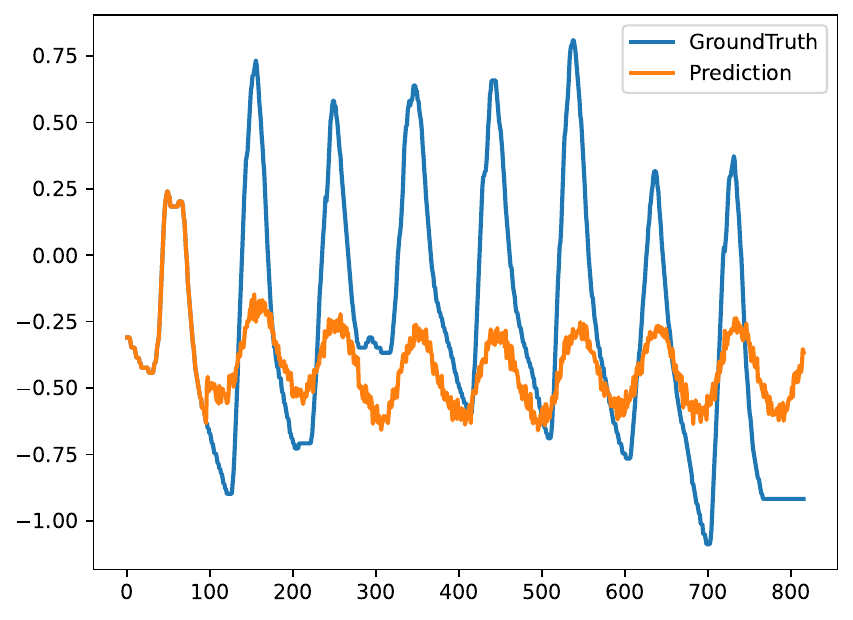}
         \caption{iTransformer}
     \end{subfigure}
     \begin{subfigure}[b]{0.35\textwidth}
         \centering
         \includegraphics[width=1\columnwidth]{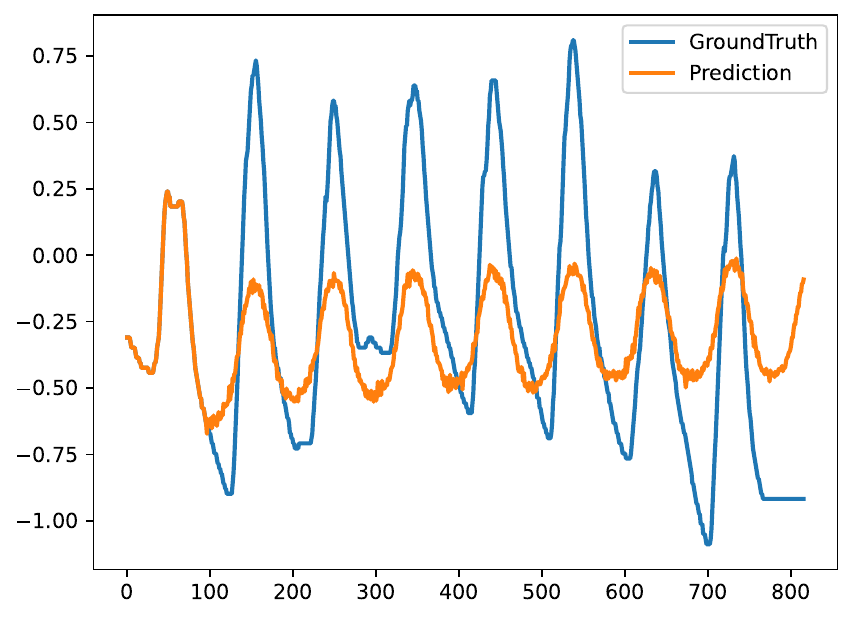}
         \caption{PatchTST}
     \end{subfigure}
        \caption{Sample prediction graph of the next T = 720 points with lookback window L = 96 from the ETTm2 dataset.}
        \label{fig:qual4}
\end{figure*}

\begin{figure*}[th]
     \centering
     \begin{subfigure}[b]{0.35\textwidth}
         \centering
         \includegraphics[width=1\columnwidth]{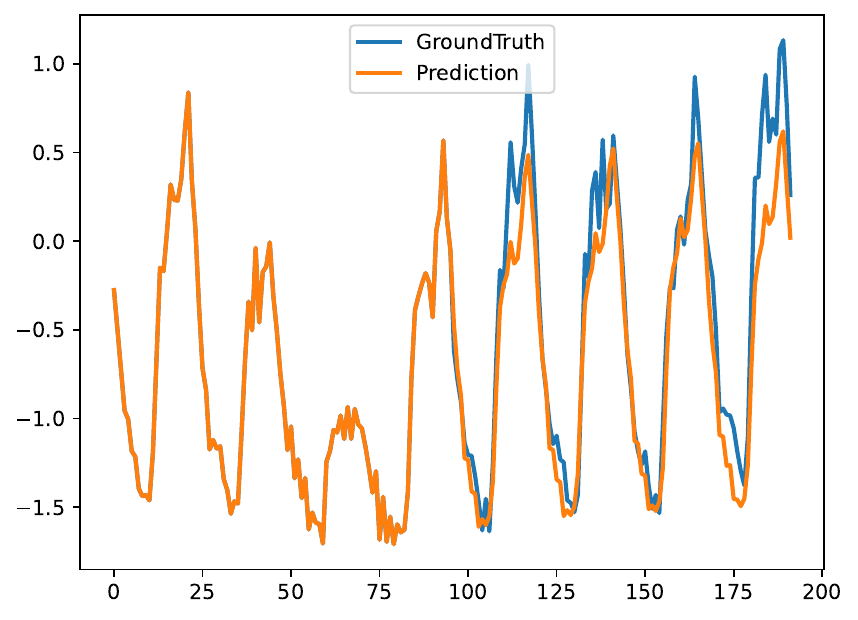}
         \caption{xPatch}
     \end{subfigure}
     \begin{subfigure}[b]{0.35\textwidth}
         \centering
         \includegraphics[width=1\columnwidth]{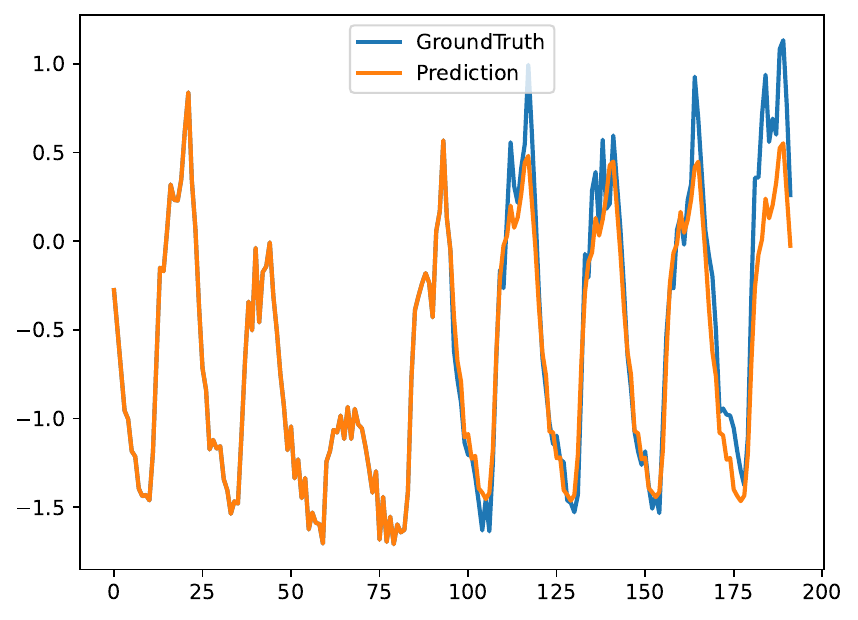}
         \caption{CARD}
     \end{subfigure}
     \begin{subfigure}[b]{0.35\textwidth}
         \centering
         \includegraphics[width=1\columnwidth]{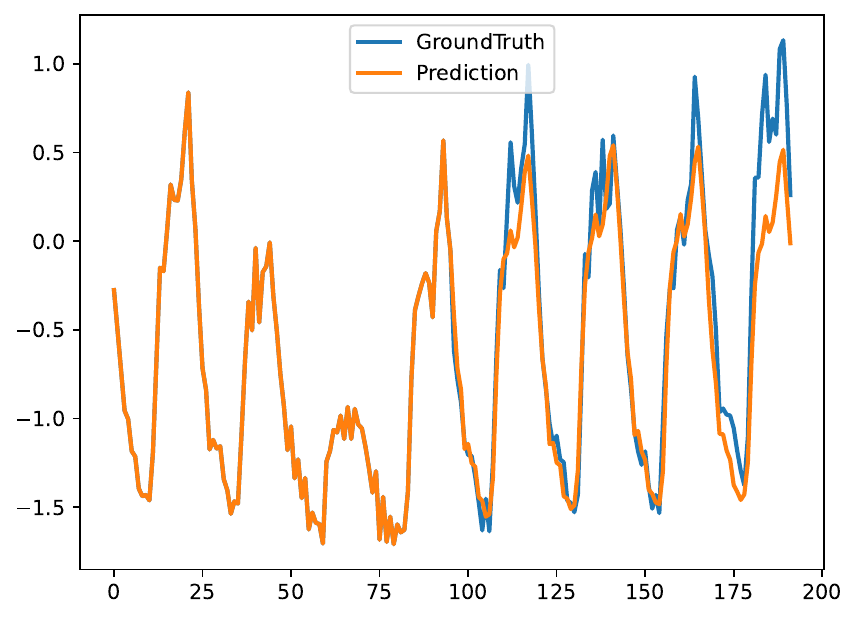}
         \caption{iTransformer}
     \end{subfigure}
     \begin{subfigure}[b]{0.35\textwidth}
         \centering
         \includegraphics[width=1\columnwidth]{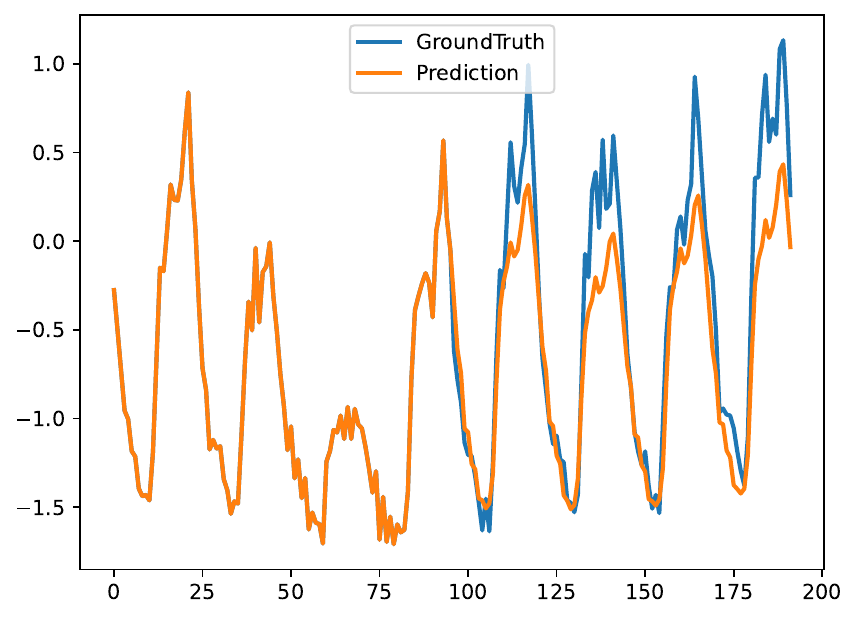}
         \caption{PatchTST}
     \end{subfigure}
        \caption{Sample prediction graph of the next T = 96 points with lookback window L = 96 from the Electricity dataset.}
        \label{fig:qual5}
\end{figure*}

\begin{figure*}[th]
     \centering
     \begin{subfigure}[b]{0.35\textwidth}
         \centering
         \includegraphics[width=1\columnwidth]{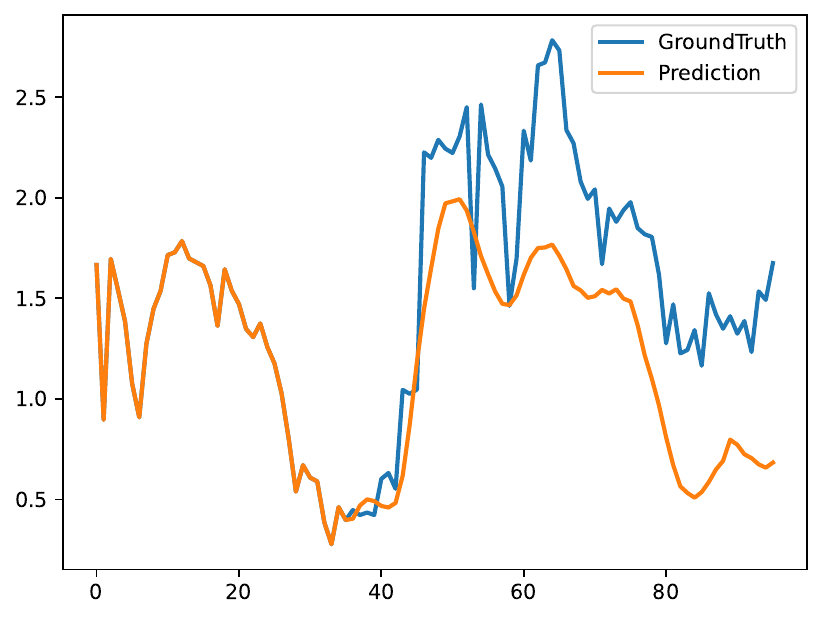}
         \caption{xPatch}
     \end{subfigure}
     \begin{subfigure}[b]{0.35\textwidth}
         \centering
         \includegraphics[width=1\columnwidth]{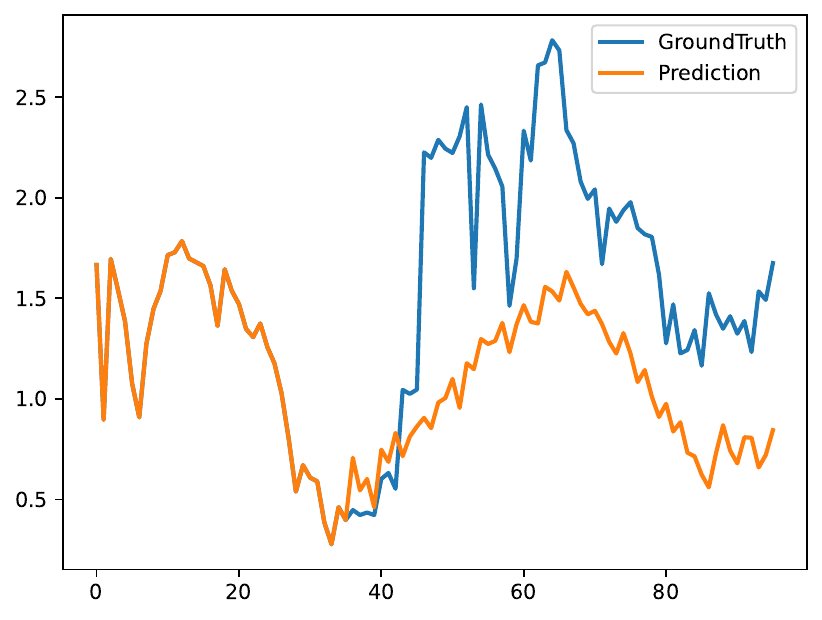}
         \caption{CARD}
     \end{subfigure}
     \begin{subfigure}[b]{0.35\textwidth}
         \centering
         \includegraphics[width=1\columnwidth]{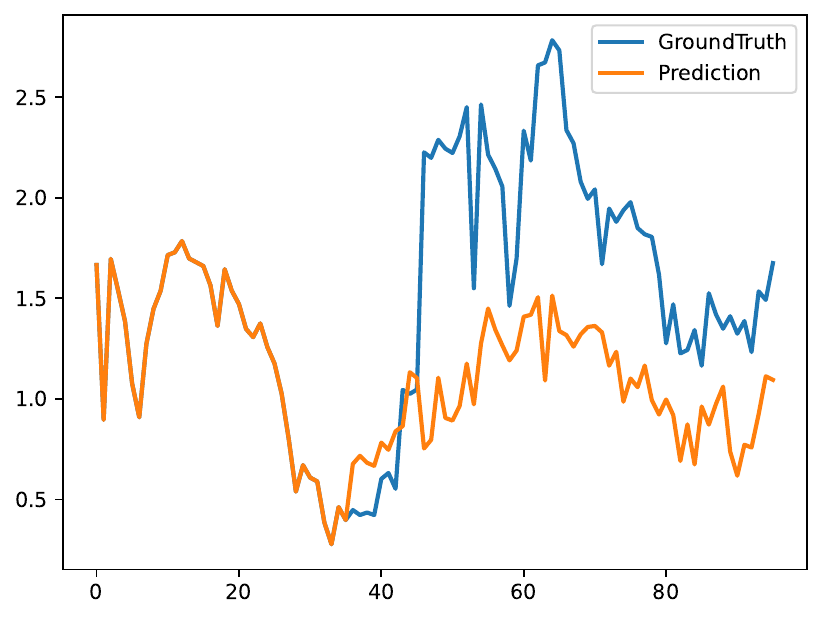}
         \caption{iTransformer}
     \end{subfigure}
     \begin{subfigure}[b]{0.35\textwidth}
         \centering
         \includegraphics[width=1\columnwidth]{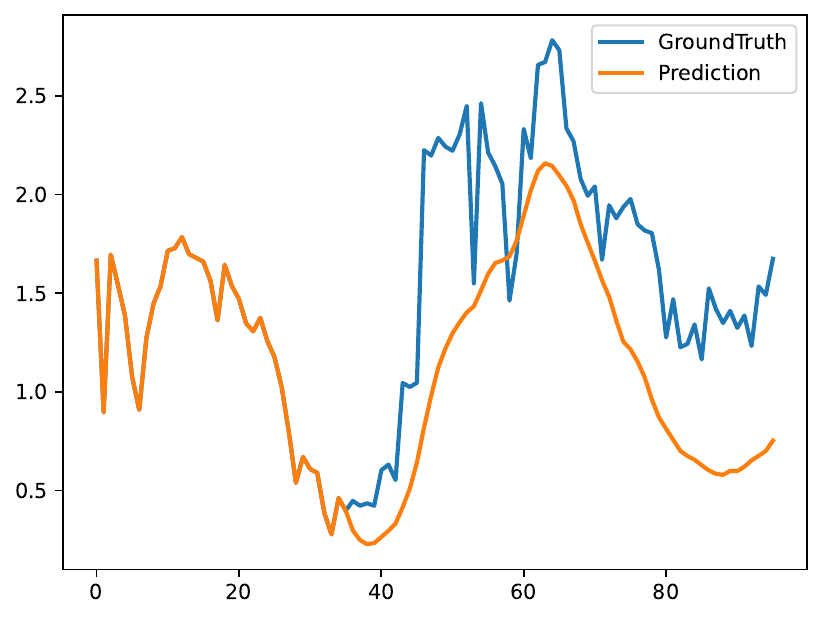}
         \caption{PatchTST}
     \end{subfigure}
        \caption{Sample prediction graph of the next T = 60 points with lookback window L = 36 from the Illness dataset.}
        \label{fig:qual6}
\end{figure*}

Figures \ref{fig:qual1}, \ref{fig:qual2}, \ref{fig:qual3}, \ref{fig:qual4}, \ref{fig:qual5}, \ref{fig:qual6} provide qualitative visualizations comparing the proposed xPatch model with recent state-of-the-art models, including CARD, iTransformer, and PatchTST.

\begin{figure*}[th]
     \centering
     \begin{subfigure}[b]{0.32\textwidth}
         \centering
         \includegraphics[width=1\columnwidth]{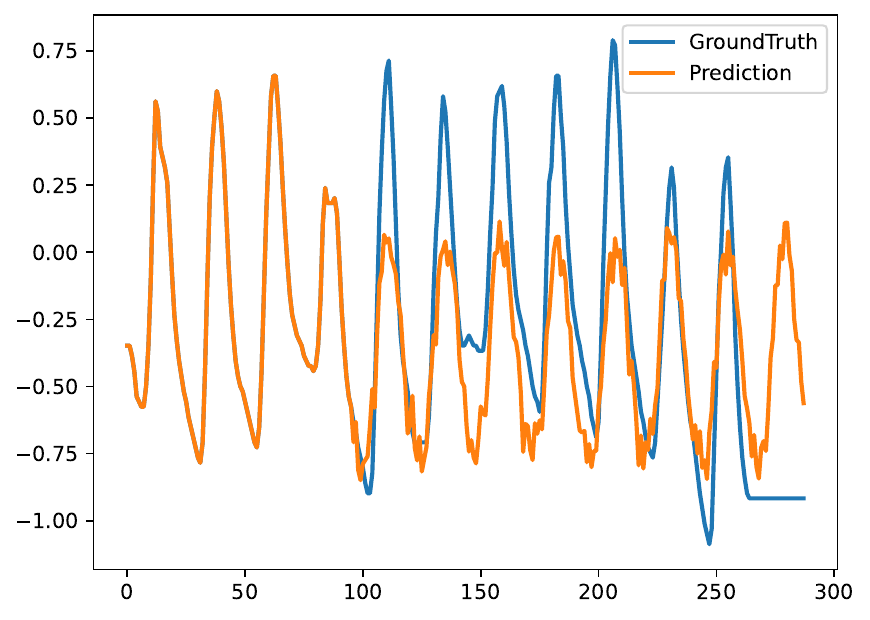}
         \caption{CNN-stream}
     \end{subfigure}
     \begin{subfigure}[b]{0.32\textwidth}
         \centering
         \includegraphics[width=1\columnwidth]{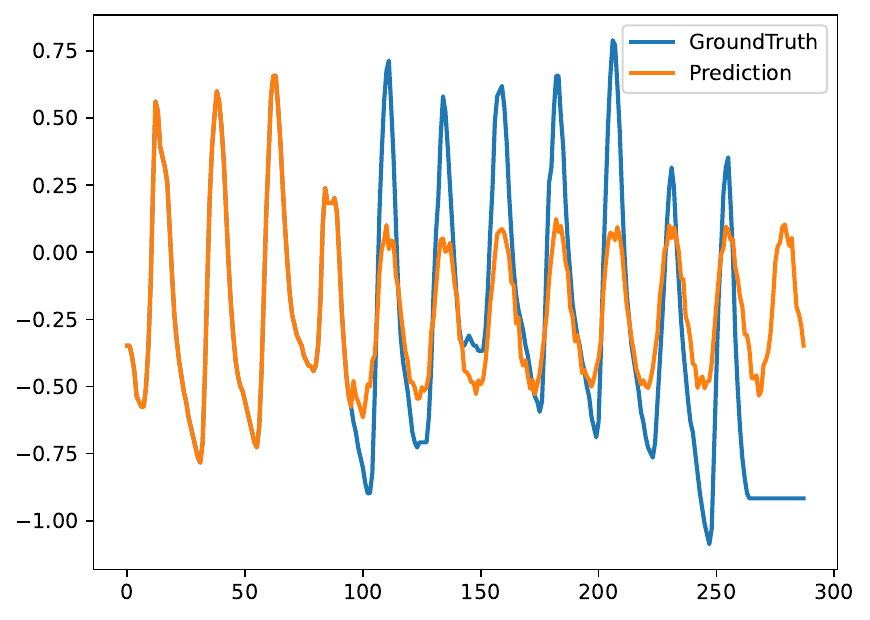}
         \caption{MLP-stream}
     \end{subfigure}
     \begin{subfigure}[b]{0.32\textwidth}
         \centering
         \includegraphics[width=1\columnwidth]{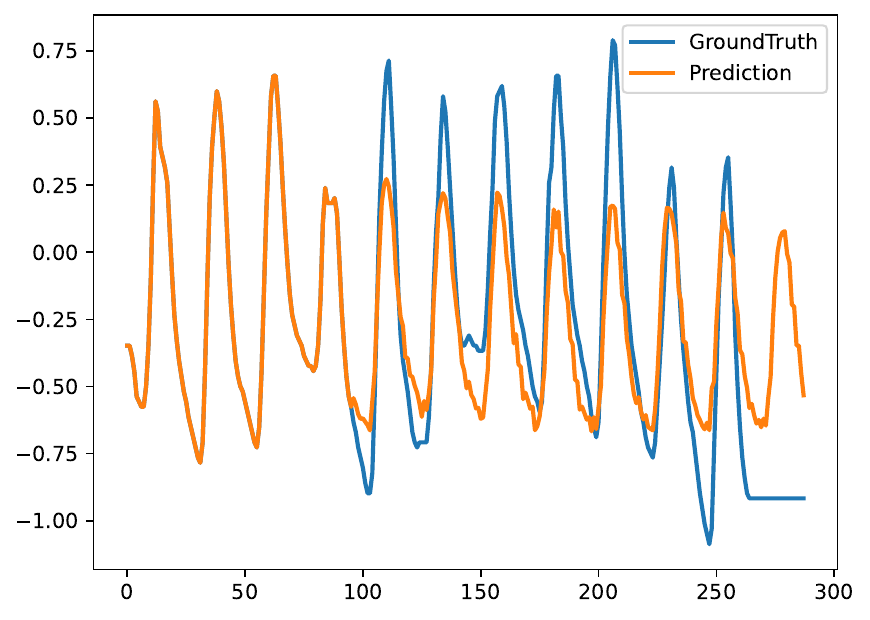}
         \caption{Dual-stream}
     \end{subfigure}
        \caption{Sample prediction graph of the next T = 192 points with lookback window L = 96 from the ETTh2 dataset.}
        \label{fig:qual7}
\end{figure*}

\begin{figure*}[th]
     \centering
     \begin{subfigure}[b]{0.32\textwidth}
         \centering
         \includegraphics[width=1\columnwidth]{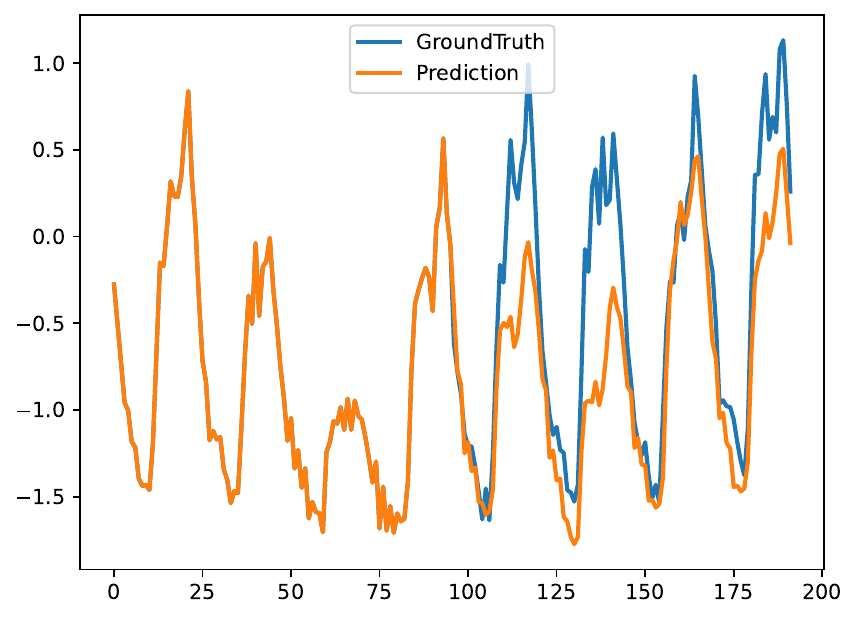}
         \caption{CNN-stream}
     \end{subfigure}
     \begin{subfigure}[b]{0.32\textwidth}
         \centering
         \includegraphics[width=1\columnwidth]{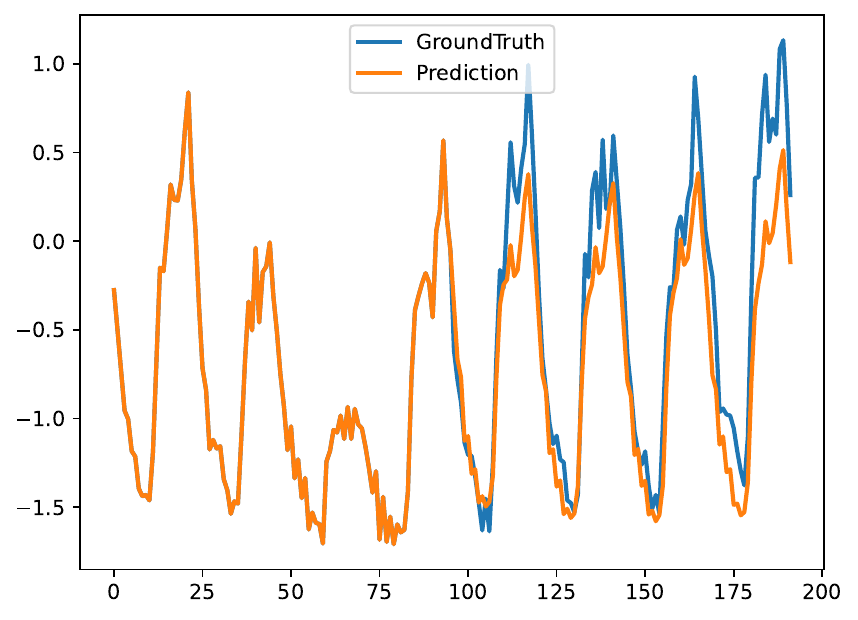}
         \caption{MLP-stream}
     \end{subfigure}
     \begin{subfigure}[b]{0.32\textwidth}
         \centering
         \includegraphics[width=1\columnwidth]{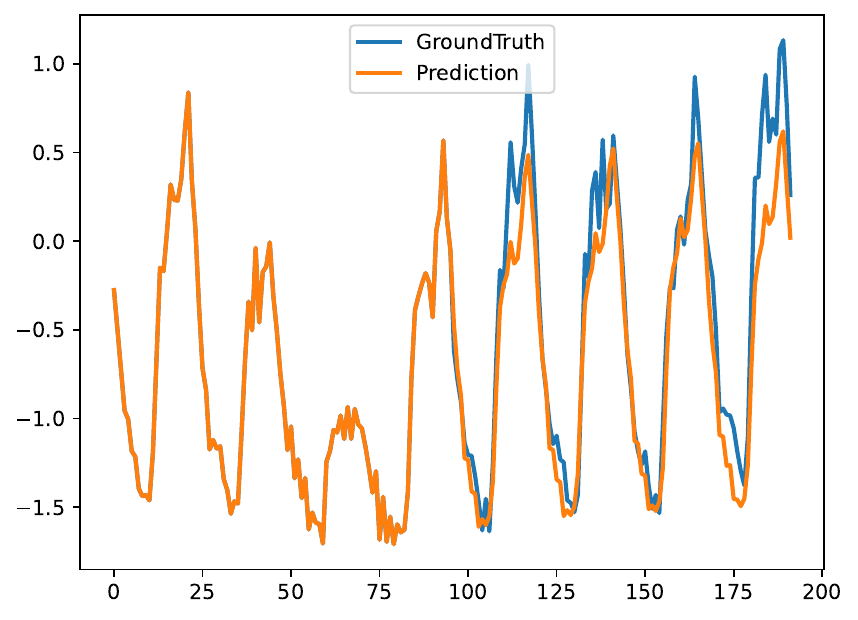}
         \caption{Dual-stream}
     \end{subfigure}
        \caption{Sample prediction graph of the next T = 96 points with lookback window L = 96 from the Electricity dataset.}
        \label{fig:qual8}
\end{figure*}

\begin{figure*}[th]
     \centering
     \begin{subfigure}[b]{0.32\textwidth}
         \centering
         \includegraphics[width=1\columnwidth]{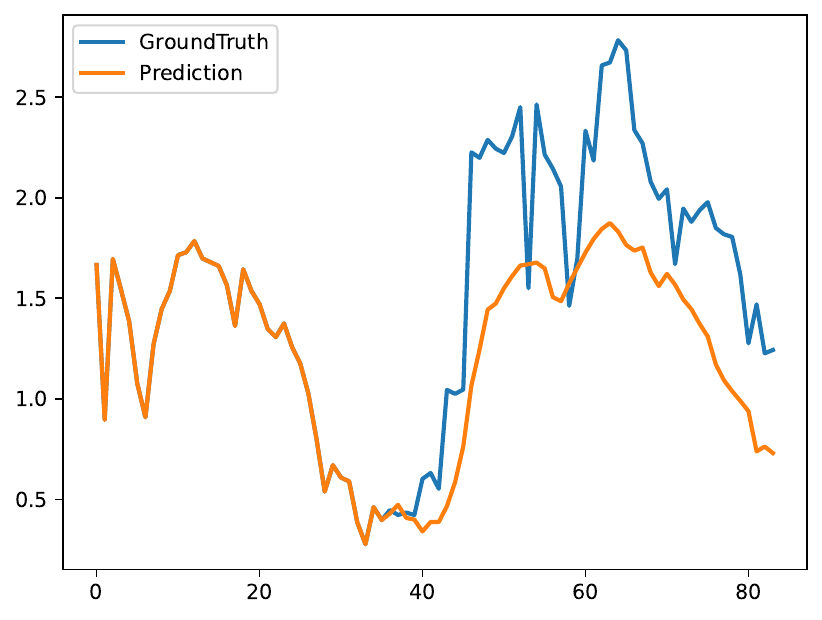}
         \caption{CNN-stream}
     \end{subfigure}
     \begin{subfigure}[b]{0.32\textwidth}
         \centering
         \includegraphics[width=1\columnwidth]{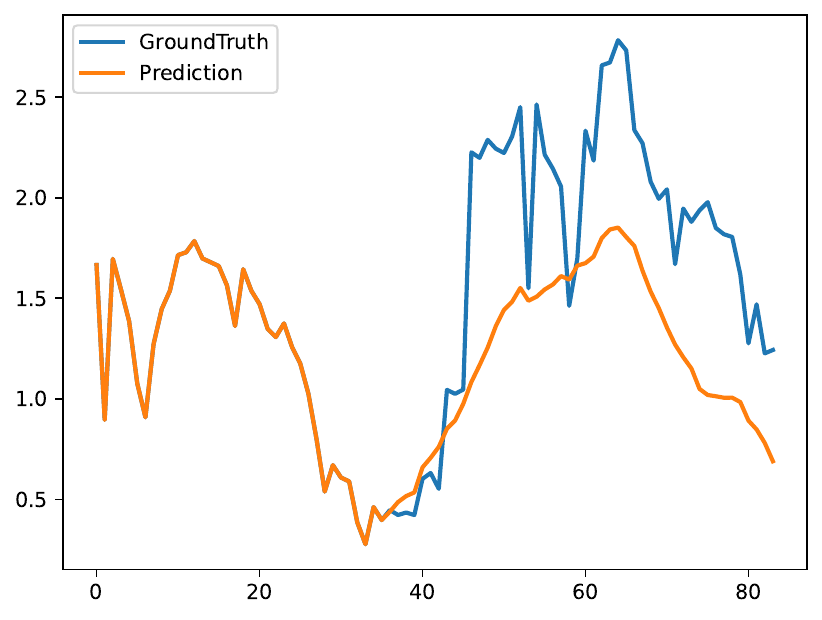}
         \caption{MLP-stream}
     \end{subfigure}
     \begin{subfigure}[b]{0.32\textwidth}
         \centering
         \includegraphics[width=1\columnwidth]{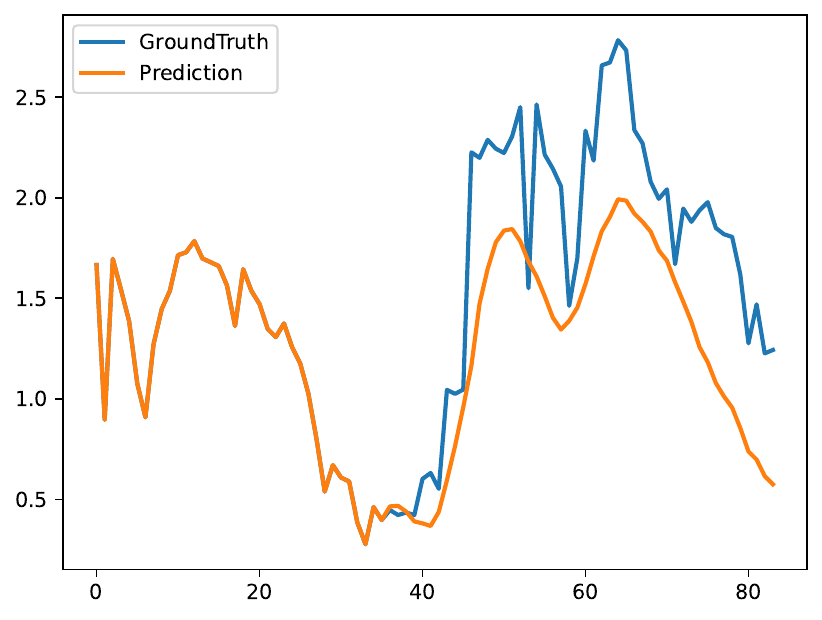}
         \caption{Dual-stream}
     \end{subfigure}
        \caption{Sample prediction graph of the next T = 48 points with lookback window L = 36 from the Illness dataset.}
        \label{fig:qual9}
        \vskip 0.5in
\end{figure*}

\begin{figure*}[th]
     \centering
     \begin{subfigure}[b]{0.32\textwidth}
         \centering
         \includegraphics[width=1\columnwidth]{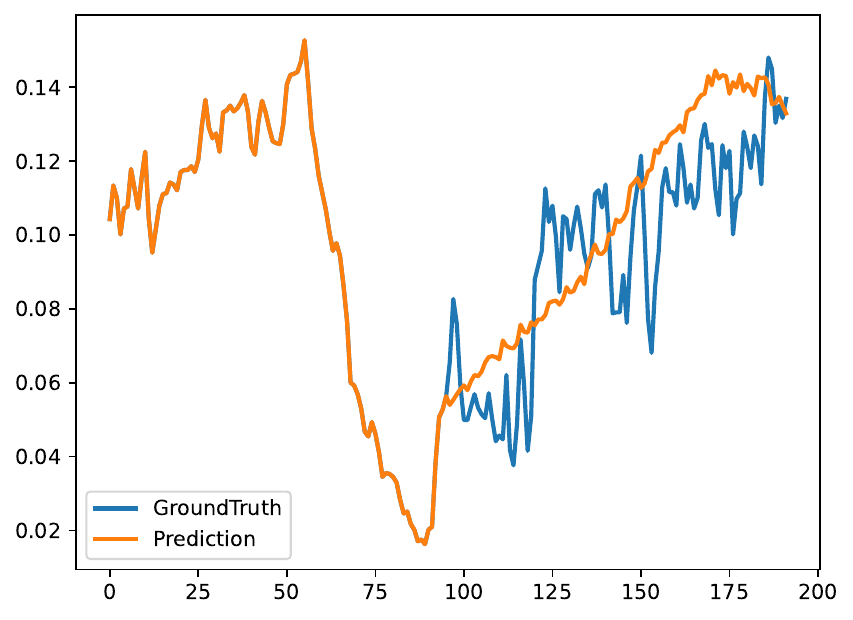}
         \caption{CNN-stream}
     \end{subfigure}
     \begin{subfigure}[b]{0.32\textwidth}
         \centering
         \includegraphics[width=1\columnwidth]{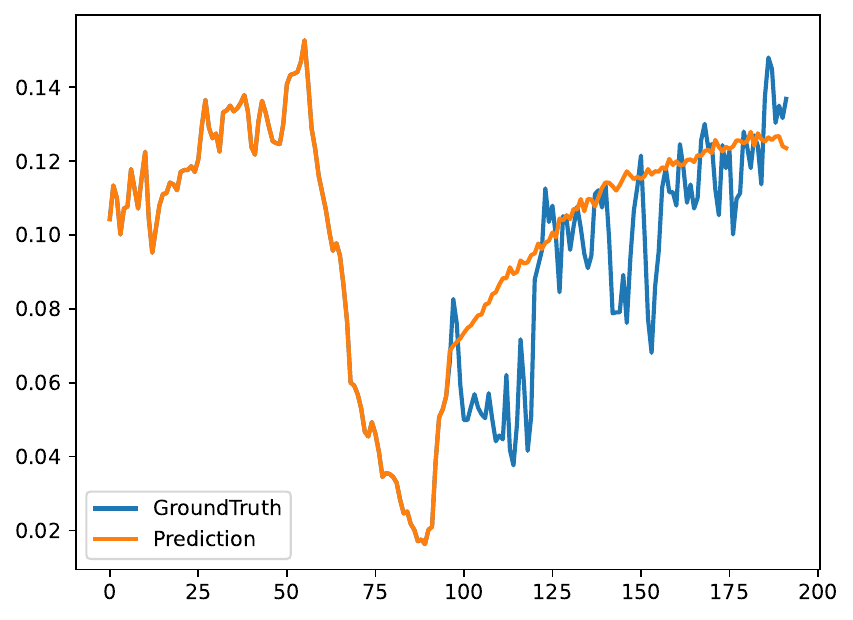}
         \caption{MLP-stream}
     \end{subfigure}
     \begin{subfigure}[b]{0.32\textwidth}
         \centering
         \includegraphics[width=1\columnwidth]{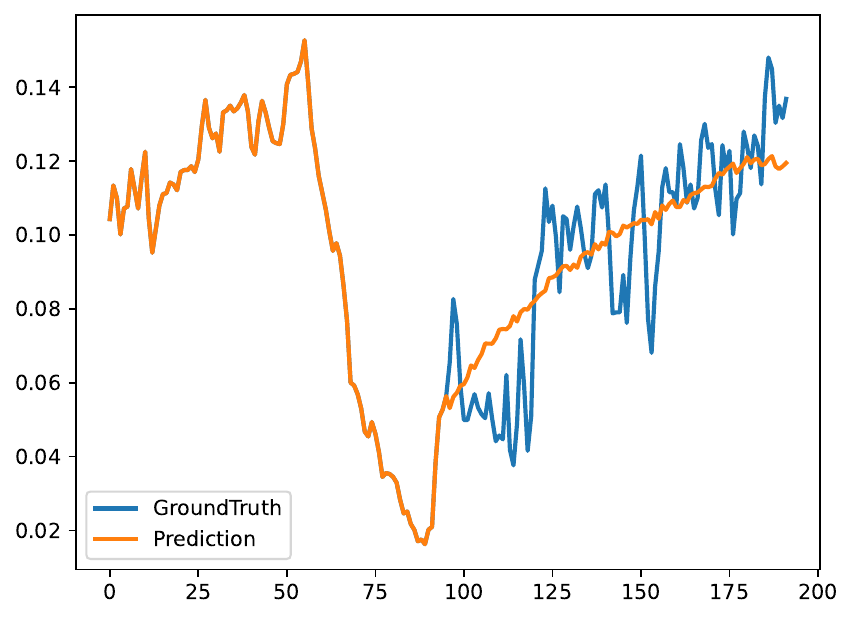}
         \caption{Dual-stream}
     \end{subfigure}
        \caption{Sample prediction graph of the next T = 96 points with lookback window L = 96 from the Weather dataset.}
        \label{fig:qual10}
\end{figure*}

\begin{figure*}[th]
     \centering
     \begin{subfigure}[b]{0.32\textwidth}
         \centering
         \includegraphics[width=1\columnwidth]{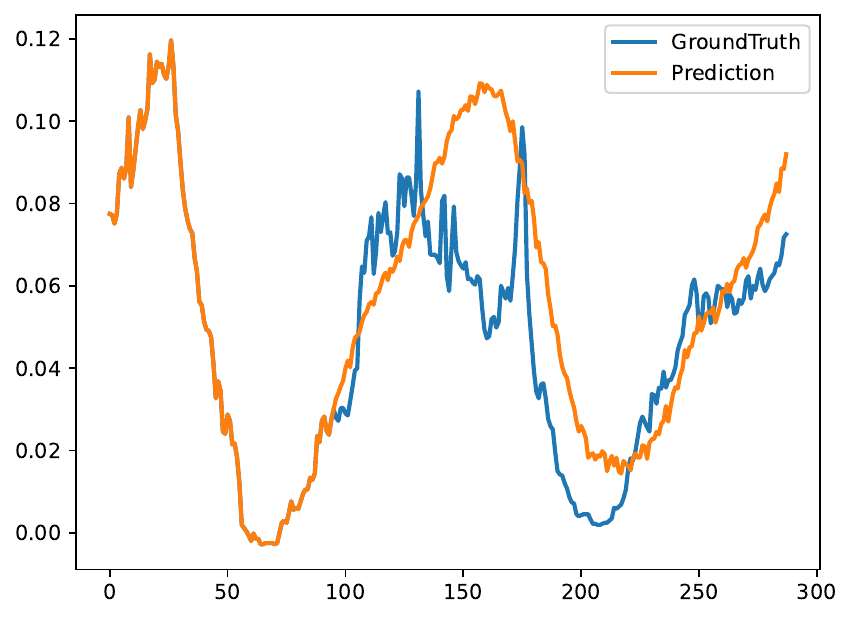}
         \caption{CNN-stream}
     \end{subfigure}
     \begin{subfigure}[b]{0.32\textwidth}
         \centering
         \includegraphics[width=1\columnwidth]{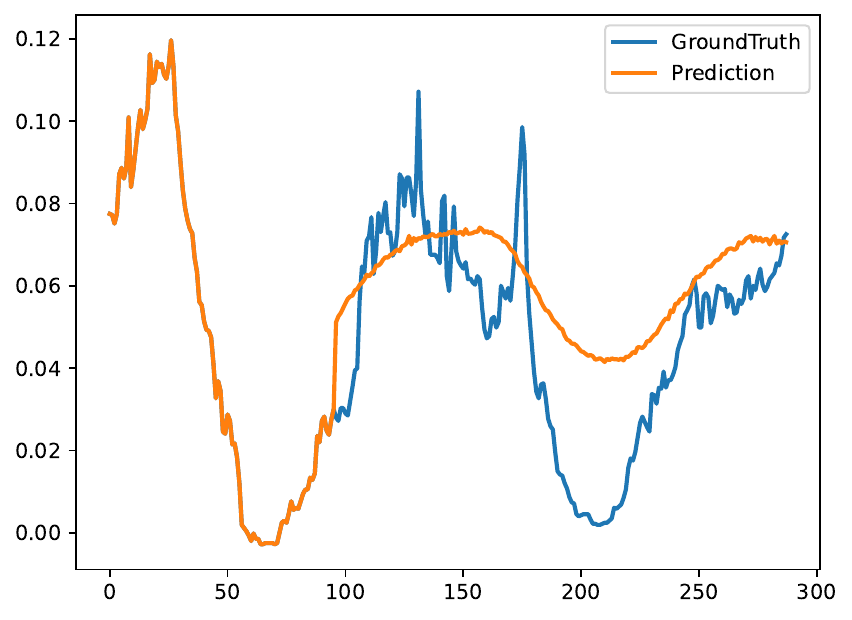}
         \caption{MLP-stream}
     \end{subfigure}
     \begin{subfigure}[b]{0.32\textwidth}
         \centering
         \includegraphics[width=1\columnwidth]{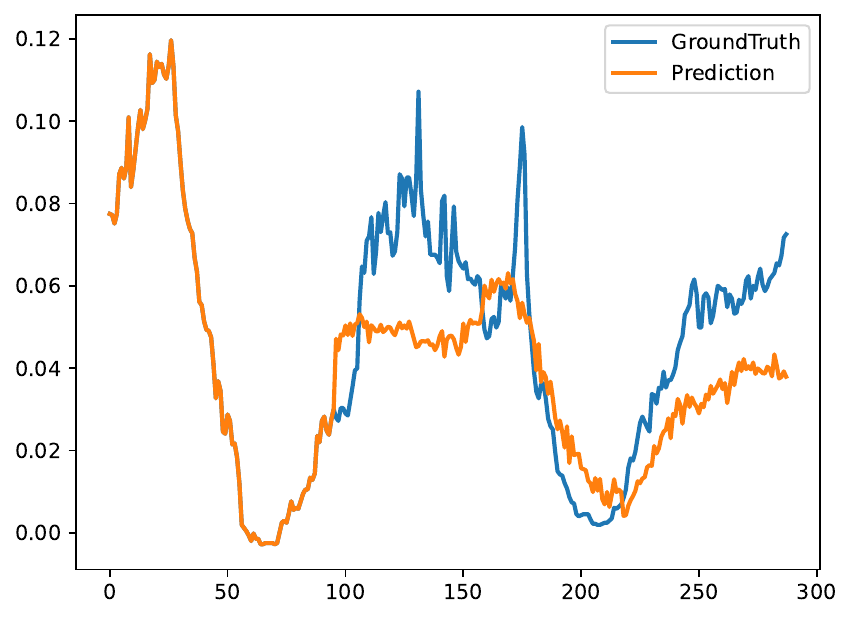}
         \caption{Dual-stream}
     \end{subfigure}
        \caption{Sample prediction graph of the next T = 192 points with lookback window L = 96 from the Weather dataset.}
        \label{fig:qual11}
\end{figure*}

\begin{figure*}[th]
     \centering
     \begin{subfigure}[b]{0.32\textwidth}
         \centering
         \includegraphics[width=1\columnwidth]{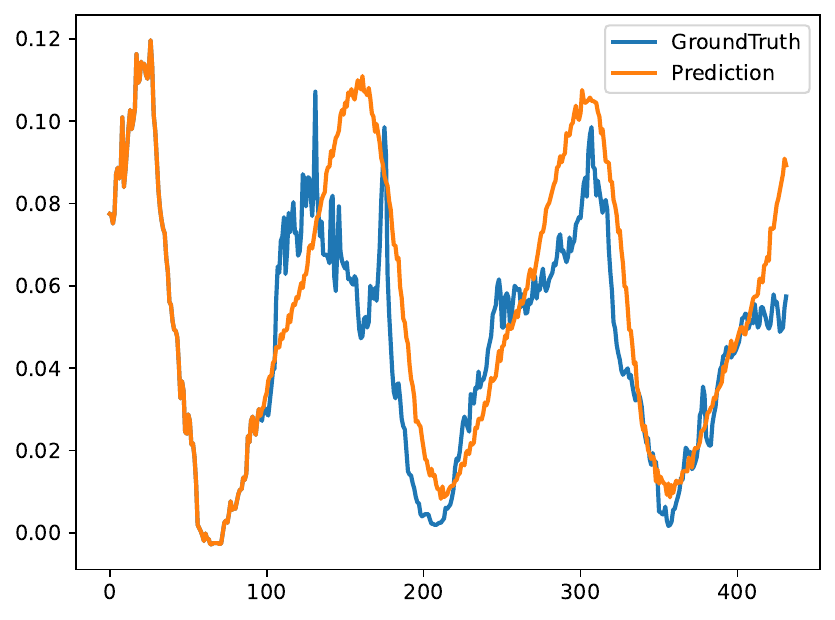}
         \caption{CNN-stream}
     \end{subfigure}
     \begin{subfigure}[b]{0.32\textwidth}
         \centering
         \includegraphics[width=1\columnwidth]{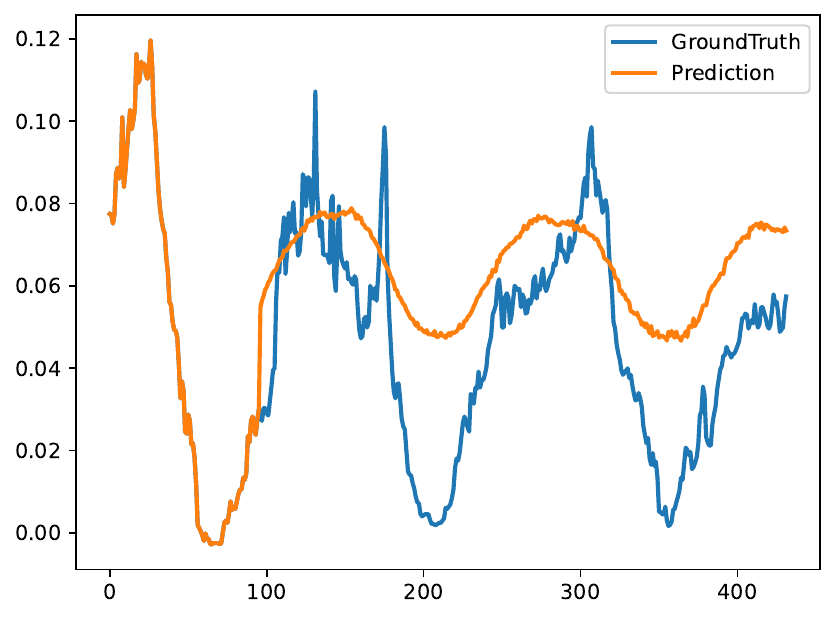}
         \caption{MLP-stream}
     \end{subfigure}
     \begin{subfigure}[b]{0.32\textwidth}
         \centering
         \includegraphics[width=1\columnwidth]{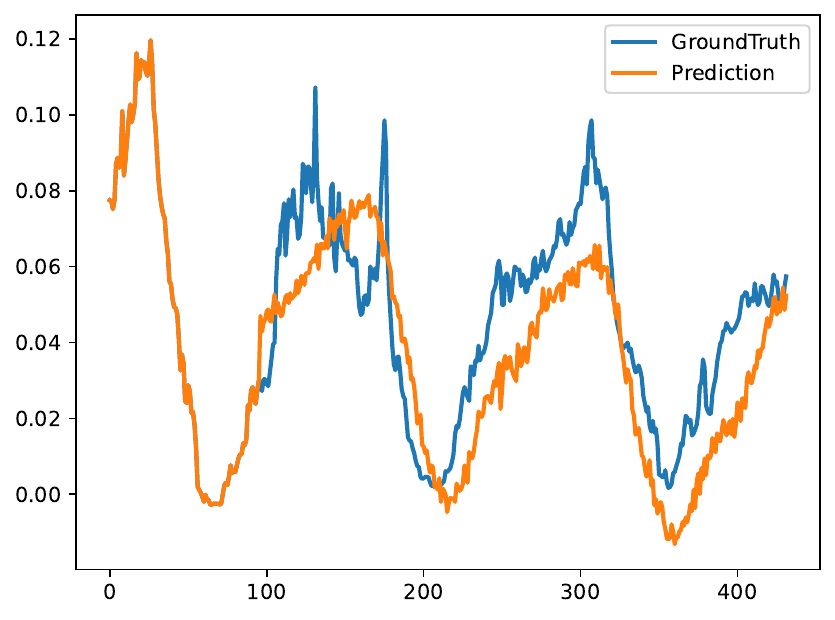}
         \caption{Dual-stream}
     \end{subfigure}
        \caption{Sample prediction graph of the next T = 336 points with lookback window L = 96 from the Weather dataset.}
        \label{fig:qual12}
\end{figure*}

Figures \ref{fig:qual7}, \ref{fig:qual8}, \ref{fig:qual9}, \ref{fig:qual10}, \ref{fig:qual11}, \ref{fig:qual12} illustrate qualitative visualizations of the separate predictions from the CNN-only and MLP-only streams, compared to the original dual-stream forecast.

\end{document}